\documentclass[manuscript,screen]{acmart}

\usepackage{microtype,siunitx,booktabs}
\usepackage{xcolor}
\usepackage{soul}
\usepackage{url}
\usepackage{booktabs}
\usepackage{multirow}
\usepackage{array,booktabs}
\usepackage{lscape}
\usepackage{graphicx}
\graphicspath{ {./images/} }

\usepackage{wrapfig,lipsum,booktabs}
\usepackage[framemethod=tikz]{mdframed}
\usepackage{lipsum}
\AtBeginDocument{%
  \providecommand\BibTeX{{%
    \normalfont B\kern-0.5em{\scshape i\kern-0.25em b}\kern-0.8em\TeX}}}

\newcommand{\eg}{\emph{e.g.}}
\setcopyright{acmcopyright}
\copyrightyear{2018}
\acmYear{2018}
\acmDOI{10.1145/1122445.1122456}

\acmConference[Woodstock '18]{Woodstock '18: ACM Symposium on Neural
  Gaze Detection}{June 03--05, 2018}{Woodstock, NY}
\acmBooktitle{Woodstock '18: ACM Symposium on Neural Gaze Detection,
  June 03--05, 2018, Woodstock, NY}
\acmPrice{15.00}
\acmISBN{978-1-4503-XXXX-X/18/06}

\begin{document}

\title{A Survey of Computer Vision Technologies In Urban and Controlled-environment Agriculture}

\author{Jiayun Luo}
\email{jiayun.luo@ntu.edu.sg}
\orcid{0000-0002-4151-6682}
\affiliation{%
  \institution{Nanyang Technological University}
  \country{Singapore}
  }
\author{Boyang Li}
\authornote{The authors can be reached at the following address: 50 Nanyang Avenue, School of Computer Science and Engineering, Nanyang Technological University, Singapore 639798. Boyang Li is the corresponding author. The research is funded by WeBank-NTU Joint Research Center and China-Singapore International Joint Research Institute.}

\email{boyang.li@ntu.edu.sg}
\orcid{0000-0002-6230-2376}
\affiliation{%
  \institution{Nanyang Technological University}
  \country{Singapore}
}
\author{Cyril Leung}
\email{CLeung@ntu.edu.sg}
\orcid{0000-0001-9911-2069}
\affiliation{%
  \institution{Nanyang Technological University}
  \country{Singapore}
}
\affiliation{\institution{China-Singapore International Joint Research Institute}
\city{Guangzhou}
\country{China}
}

\renewcommand{\shortauthors}{Luo, et al.}

\begin{abstract}

In the evolution of agriculture to its next stage, Agriculture 5.0, artificial intelligence will play a central role. Controlled-environment agriculture, or CEA, is a special form of urban and suburban agricultural practice that offers numerous economic, environmental, and social benefits, including shorter transportation routes to population centers, reduced environmental impact, and increased productivity. Due to its ability to control environmental factors, CEA couples well with computer vision (CV) in the adoption of real-time monitoring of the plant conditions and autonomous cultivation and harvesting. The objective of this paper is to familiarize CV researchers with agricultural applications and agricultural practitioners with the solutions offered by CV. We identify five major CV applications in CEA, analyze their requirements and motivation, and survey the state of the art as reflected in 68 technical papers using deep learning methods. In addition, we discuss five key subareas of computer vision and how they related to these CEA problems, as well as fourteen vision-based CEA datasets. We hope the survey will help researchers quickly gain a bird-eye view of the striving research area and will spark inspiration for new research and development. 


\end{abstract}


\maketitle

\section{INTRODUCTION }
\label{sec:int}



Artificial intelligence (AI), especially computer vision (CV), is finding an ever broadening range of applications in modern agriculture. The next stage of agricultural technological development, Agriculture 5.0 \citep{zambon2019revolution,ahmad2021agriculture,ragazou2022agriculture,fraser2019agriculture}, will constitute AI-driven autonomous decision making as a central component. The term Agriculture 5.0 stems from a chronology \citep{zambon2019revolution} that begins with Agriculture 1.0, which heavily depends on human labor and animal power, and Agriculture 2.0, enabled by synthetic fertilizers, pesticide, and combustion-powered machinery, and develops to Agriculture 3.0 and 4.0, characterized by GPS-enabled precision control, and Internet-of-Thing (IoT) driven data collection \cite{saiz2020smart}. Built upon the rich agricultural data collected, Agriculture 5.0 holds the promise to further increase productivity, satiate the food demand of a growing global population, and mitigate the negative environmental impact of existing agricultural practices.

As an integral component of Agriculture 5.0, controlled-environment agriculture (CEA), a farming practice carried out within urban, indoor, resource-controlled, and sensor-driven factories, is particularly suitable for the application of AI and CV. This is because CEA provides ample infrastructure support for data collection and autonomous execution of algorithmic decisions. 
In terms of productivity, CEA could produce higher yield per unit area of land \citep{saturnbioponics,SpreadVertifarm} and boost the nutritional content of agricultural products \citep{kopsell2015blue,trojak2022effects}. In terms of environmental impact, CEA farms can insulate environmental influences, relieve the need for fertilizer and pesticides, and efficiently utilize recycled resources like water, thereby may be much more environmentally friendly and self-sustainable than traditional farming. 


In the light of current global challenges, such as disruptions to global supply chains and the threat of climate change, CEA appears especially appealing as a food source for urban population centers. Under pressures of deglobalization brought by geopolitical tensions \citep{zhang2021case} and global pandemics \citep{sidor2020dietary,rahimi2021impact}, CEA provides the possibility to build farms close to large cities, which shortens the transportation distance and maintains secure food supplies even when long-distance routes are disrupted. The city-state Singapore, for example, has promised to source 30\% of its food domestically by 2030 \citep{30by30gov,Singapore30in30}, which is only possible through suburban farms such as CEAs. 
Furthermore, CEA, as a form of precision agriculture, is by itself a viable solution to the reduction of the emission of greenhouse gasses \citep{SpreadVertifarm,benis2017development,10.1145/3485128}. CEA can also shield plants from adverse climate conditions exacerbated by climate change as its environments are fully controlled \citep{gomez2020mitigation} and is able to effectively reuse the arable land eroded due to climate change \citep{zhang2011climate}.

We argue that AI and CV are critical to the economic viability and long-term sustainability of CEAs as these technologies could save expenses associated with production and improve productivity. Suburban CEAs have high land costs. An analysis in Victoria, Australia \cite{benke2017future} shows that, due to the higher land cost resulting from proximity to cities, with an estimated 50-fold productivity improvement per land area, it still takes 6 to 7 years for a CEA to reach the break-even point. Thus, further productivity improvement from AI would act as strong drivers for CEA adoption. 
Moreover, vertical or stacked setup of vertical farms impose additional difficulty for farmers to perform daily surveillance and operations. Automated solutions empowered by computer vision could effectively solve this problem. Finally, AI and CV technologies have the potential to fully characterize the complex, individually different, time-varying, and dynamic conditions of living organisms \citep{berckmans2017general}, which will enable precise and individualized management and further elevate yield. Thus, AI and CV technologies appear to be a natural fit to CEAs.


Most of the recent development of AI can be attributed to the newly discovered capability to train deep neural networks \cite{DLReview-2015} that can (1) automatically learn multi-level representations of input data that are transferable to diverse downstream tasks \cite{chen2020simple,hermann2020shapes}, (2) easily scale up to match the growing size of data \cite{Sun_2017_ICCV}, and (3) conveniently utilize massively parallel hardware architectures like GPUs \cite{goyal2018accurate,EfficientNet-Supercomputer-Scale-2021}. As function approximators, deep learning proves to be surprisingly effective in generalizing to previously unseen data \cite{zhang2017understanding}. Deep learning has achieved tremendous success in computer vision \cite{tan2020efficientnet}, natural language processing \cite{brown2020language,devlin2019bert,XuGuo-2021}, multimedia \cite{Anderson_2018_CVPR,PelinDogan-2018}, robotics \cite{Niko2018}, game playing \cite{Silver2017}, and many other areas.  

The AI revolution in agriculture is already underway. State-of-the-art neural network technologies, such as ResNet \cite{he2015deep} and MobileNet \cite{howard2017mobilenets} for image recognition, and Faster R-CNN \cite{ren2016faster}, Mask R-CNN \cite{he2018mask}, and YOLO \cite{redmon2016look} for object detection,  have been applied to the management of crops \cite{10.3389}, livestock \cite{ani9070470,TIAN2019104840}, and plants in indoor and vertical farms \cite{REYESYANES2020105827,8520836}. AI has been used to provide decision support in a myriad of tasks from DNA analysis \citep{10.3389} and growth monitoring \cite{REYESYANES2020105827,8520836} to disease detection \cite{selvaraj2019ai} and profit prediction \cite{8524140}. 


While several surveys have explored the use of computer vision (CV) techniques in agriculture, none of them specifically focus on CEA applications. 
Some surveys summarize studies based on aspects of practical applications in agriculture. 
\citep{sornalakshmi2022technical,domingues2022machine,habib2021machine,iqbal2018automated,cubero2016automated} survey pest and disease detection studies. \citep{bhargava2021fruits, tripathi2020role, gomes2012applications} discuss fruit and vegetable quality grading and disease detection.
\citep{tian2020computer} summarizes studies in six sub-fields, including crop growth monitoring, pest and disease detection, automatic harvesting/fruit detection, fruit quality testing, automated management of modern farms and the monitoring of farmland information with Unmanned Aerial Vehicle (UAV). 
Other survey organize existing works from a technical perspective, namely algorithms used \citep{rehman2019current} or formats of data \citep{chandra2020computer}.
\citep{kakani2020critical}, as an exception, introduces the development history of CV and AI in smart agriculture, without investigating any individual studies. Our work aims to address this gap and provide insights tailored to CEA-specific contexts.

As the volume of research in smart agriculture grows rapidly, we hope the current review article can bridge researchers from AI and agriculture and create a mild learning curve when they wish to familiarize themselves in the other area.  We believe computer vision has the closest connections with, and is the most immediately applicable in, urban agriculture and CEAs. Hence, in this paper, we focus on reviewing deep-learning based computer vision technologies in urban farming and CEAs. We focus on deep learning because it is the predominant approach in AI and CV research. The contributions of this paper are two-fold, with the former targeted at AI researchers and the latter targeted at agriculture researchers: 

\begin{itemize}
  \item We identify five major CV applications in CEA and analyze their requirements and motivation. Further, we survey the state of the art as reflected in 68 technical papers and 14 vision-based CEA datasets.
  \item We discuss five key subareas of computer vision and how they relate to CEA. In addition, we identify four potential future directions for research in CV for CEA. 
\end{itemize}



\begin{figure}[t]

\caption{
An illustration of the end-to-end agriculture process of CEAs, from seed planting to harvest and sales, with five major deep learning based CV in agriculture applications--Growth Monitoring, Fruit and Flower Detection, Fruit Counting, Maturity Level Classification and Pest and Disease Detection -- mapped to the corresponding applicable plant growth stages. Autonomous Seed Sowing and Autonomous Harvest and Sales in gray boxes are relevant steps in the agriculture process of CEAs but are out of the scope of our survey which focus on CV in CEAs.  Orange lines represent arrows originated from pest and disease detection. Green lines represent arrows with stage 4 as destination. 
}
\label{fig:overview}
\centering
\includegraphics[width=0.9\textwidth]{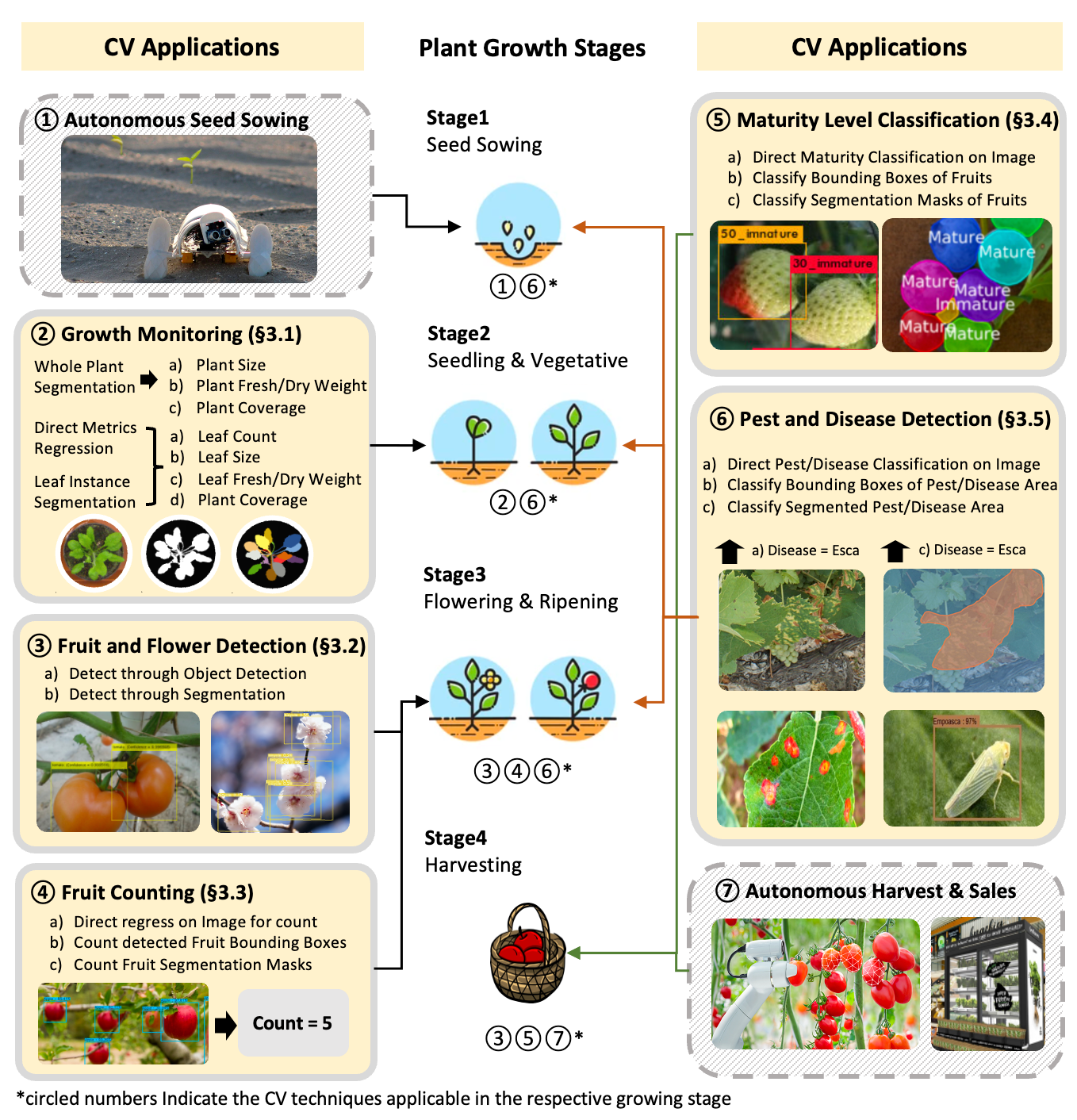}
\end{figure}

In figure \ref{fig:overview} we provide an graphical preview of our content. It illustrates the end-to-end agriculture process of CEAs, from seed planting to harvest and sales, with five major deep learning based CV applications--Growth Monitoring, Fruit and Flower Detection, Fruit Counting, Maturity Level Classification and Pest and Disease Detection--mapped to the corresponding applicable plant growth stages. We do not survey the autonomous seed planting and harvesting step as they are more relevant to robot functioning and robotic control, i.e grasping, carrying and placing of objects rather than computer vision (we do include the localization of fruit in the fruit and flower detection section that facilitate harvesting robot to locate the targeted object and perform action). However, we provide here some literature related to agriculture robot and end-effector design for reference \citep{zhang2020state,duckett2018agricultural,cheein2013agricultural,r2018research,bechar2016agricultural}

We structure the survey following the process in the figure: First, to provide a bird-eye view of CV capabilities available to researchers in smart agriculture, we summarize several major CV problems and influential technical solutions in \S \ref{sec:cv}. Next, we review 68 papers with respect to the application of computer vision in the CEA system in \S \ref{sec:cea}. The discussion is organized into five subsections: Growth Monitoring, Fruit and Flower Detection, Fruit Counting, Maturity Level Classification, and Pest and Disease Detection. In the discussion, we focus on fruits and vegetables that are suitable for CEA, including tomato \citep{hortidaily-cucumber,hao2015response,yuan2020robust,afonso2020tomato}, mango \citep{Horti-mango}, guava \citep{silva2016molecular,xavier2022gas}, strawberry \citep{8863343, 9119372}, capsicum \citep{horticulturae7090284}, banana \citep{hortidaily-banana}, lettuce \citep{zhang2020growth}, cucumber \citep{hortidaily-cucumber,hao1999effects,ma2017segmentation}, citrus \citep{Hortibizdaily-citrus} and blueberry \citep{Thespoon-blurberry}. Next, we provide a summary of fourteen publicly available datasets of plants and fruits in \S \ref{sec:data} to facilitate future studies in Controlled-environment agriculture. Finally, we highlight a few research directions that could generate high-impact research in the near future in \S \ref{sec:future}.


One thing to note here is that, except for the Leaf Instance Segmentation task under the Growth Monitoring section, all the tasks are performed with model trained from different datasets and evaluated on different metrics. Table \ref{perfomancetab:fruitflowerdetect} \ref{perfomancetab:fruitcount}, \ref{perfomancetab:matureclass}, \ref{perfomancetab:pestdiseasedetect} showcase the variety in datasets and evaluation metrics. This variation results in incomparable performance between studies. Such a phenomenon further indicates the necessity of our survey, which summarizes the current progress in literature and encourages the development of general benchmarks to promote consistency and comparability in future research.

\section{Computer Vision Capabilities Relevant to Smart Agriculture}
\label{sec:cv}

\subsection{Image Recognition}
\label{image recog}

The classic problem of image recognition is to classify an image containing a single object to the corresponding object class. 
The success of deep convolutional networks in this area dates (at least) back to LeNet \cite{LeNet-1998} of 1998, which recognizes hand-written digits. The fundamental building block of such networks is the convolution operation. Using the principles of local connections and weight sharing, convolutional networks benefit from an inductive bias of translational invariance. That is, a convolutional network applies (approximately) the same operation to all pixel locations of the image. 

The victory of AlexNet \cite{AlexNet-2012} in the 2012 ImageNet Large Scale Visual Recognition Challenge \cite{ILSVRC15} is often considered as a landmark event that introduced deep neural networks into the AI mainstream. Subsequently, many variants of convolutional networks \cite{Simonyan15:VGG,Inception2015,larsson2017fractalnet,jacobsen2018irevnet} have been proposed. Due to space limits, here we provide a brief review of a few influential works, which is by no means exhaustive. ResNet \cite{he2015-resnet-report} introduces residual connections that allow the training of networks of more than 100 layers. ResNeXT \cite{ResNeXT2016} and MobileNet \cite{MobileNet-2017} employ grouped convolution that reduces interaction between channels and improves the efficiency of the network parameters. ShuffleNet \cite{zhang2017shufflenet} utilizes the shuffling of channels, which complements group convolution. EfficientNet \cite{EfficientNet-2019} shows simultaneous scaling of the network width, height, and image resolution is key to efficient use of parameters. 

Recently, the transformer model has proven to be a highly competitive architecture for image recognition and other computer vision tasks \cite{dosovitskiy2020:ViT}. These models cut the input image into a sequence of small image patches and often apply strong regularization such as RandAugment \cite{RandAugment2019}. Variants such as CaiT \cite{Touvron_2021:ViT-Cait}, CeiT \cite{Yuan_2021_ICCV}, Swin Transformer \cite{liu2021:SwinTransformer}, and others \cite{yan2021contnet,dai2021coatnet,Twins2021,zhou2021deepvit} achieve outstanding performance on ImageNet. 

Despite the maturity of the technology for image classification, the assumption that an image contains only one object may not be easily satisfied in real-world scenarios. Thus, it is often necessary to adopt a problem formulation as object detection or semantic / instance segmentation.

\subsection{Object Detection}
\label{obj detect}
The object detection task is to identify and locate all objects in the image. It can be understood as the task resulted from relaxing the assumption that the input image contains a single object. This is one natural problem formulation for real-world images and has seen wide adoption in agricultural applications. 

In broad strokes, contemporary object detection methods can be categorized into anchor-box-based and point-based / proposal-free approaches. In anchor-box methods \cite{FastRCNN,FasterRCNN2015}, the process starts with a number of predefined anchor boxes that are periodically tiled to cover the entire input image. For each anchor box, the network
makes two types of predictions. First, it determines if the anchor box contains one of the predefined object classes. Second, if the box contains an object, the network attempts to move and reshape the box to become closer to the ground-truth location of the object. One-stage anchor-box detectors \cite{YOLO9000,SSD2016,fu2017dssd,RetinaNet2017,zhao2019m2det,DeformableCN2017} make these predictions all at once. In comparison, two-stage detectors \cite{FastRCNN,FasterRCNN2015,MaskRCNN2017,Lin_2017_CVPR}, in the first stage discard anchor boxes that do not contain any object and classify the remaining boxes into finer object categories in the second stage. The location adjustment, known as bounding box regression, can happen in both stages. It is also possible to employ more than two stages \cite{CascadeRCNN2018}. 
When the objects have diverse shapes and scales, these methods must create a large number of proposal boxes and evaluate them all, which can lead to high computational cost. 

While point-based object detectors \cite{CornerNet2018,CornerNet2019,FCOS2019,Bottom_Up_Zhou_2019,FoveaBox2020} still need to identify rectangular boxes around the objects,  they make predictions at the level of grid locations on the feature maps. The networks predict if a grid location is a corner or the center of an object bounding box. After that, the algorithm assembles the corners and centers into bounding boxes. 
The point-based approaches can reduce the total number of decisions to be made. A careful comparison and analysis of anchor-box methods and point-based methods can be found in \cite{Zhang_2020_CVPR}.

\subsection{Semantic, Instance, and Panoptic Segmentation}
\label{segmentation}
Segmentation is a pixel-level classification task, aiming to classify every pixel in the image into a type of object or an object instance. The variations of the task differ by their definitions of the classes. In \emph{semantic segmentation} \cite{Gould2009,Ladicky2009,Ciresan:NIPS2012,Farabet:TPAMI2013,long2015FCN}, each type of object, such as cat, cow, grass, or sky, is its own class, but different instances of the same object type (\eg, two cats) share the same class. In \emph{instance segmentation} \cite{Hariharan2014,pinheiro2015doll,Hayder2016:boundary-aware,DaiJifeng2016:Instance}, different instances of the same object type become unique classes, so that two cats are no longer the same class. However, object types such as sky or grass, which are not easily divided into instances, are ignored. In the recently proposed \emph{panoptic segmentation} \cite{Kirillov_2019_CVPR,LiuHuanyu2019,LiYanwei2019,Geus2021,ChengBowen2020,zhang2021knet}, objects are first separated into things and stuff. Things are countable and each instance of things is its own class, whereas stuff is uncountable, impossible to separate into instances, appearing as texture or amorphous regions \cite{Adelson2001:Seeing-Stuff}, and remains as one class. We note that the distinction between things and stuff is not rigid and can change depending on the application. For example, grass is typically considered as stuff, but in the leaf instance segmentation task, each leaf of a plant becomes an instance and is a separate class. 

The primary requirement of pixel-level classification is to learn pixel-level representations that consider sufficient context and within reasonable computational budget. A typical solution is to introduce a series of downsampling followed by a series of upsampling operations. Since classic works such as the Fully Convolutional Network (FCN) \cite{long2015FCN} and U-Net \cite{Ronneberger2015}, this has been the mainstream strategy for various segmentation strategies. 

Due to its use in leaf segmentation, a problem in plant phenotyping, instance segmentation may be the most relevant segmentation formulation for urban farming. 
Despite the apparent similarity to semantic segmentation, instance segmentation poses challenges due to the variable number of instance classes and possible permutation of class indices \cite{Brabandere-InstSeg-DiscriminativeLoss2017}. This could be handled by combining proposal-based object detection and segmentation \cite{Hariharan2014,ChenYi-Ting2015,Pinheiro2016,LiYi2017,chen2019hybrid}. Mask-RCNN \cite{MaskRCNN2017} exemplifies this approach. Leveraging its object detection capability, the network associates each object with a bounding box. After that, the network predicts a binary mask for the object within the bounding box. However, such methods may not perform well when there is substantial occlusion among objects or when objects are of irregular shapes \cite{Brabandere-InstSeg-DiscriminativeLoss2017}.

Departing from the detect-then-segment paradigm, recurrent methods \cite{romera2016recurrent,ren2017end,Salvador2017} that outputs one segmentation mask at one time may be considered as implicitly modeling occlusion. Pixel embedding methods \cite{Davy2019,Wolny_2022_CVPR,ying2021embedmask,wu2020improving,ChenLong-Object-aware-Embedding:2019,Brabandere-InstSeg-DiscriminativeLoss2017,Payer-Instance-Segmentation:2018} learn vector representations for every pixel and cluster the vectors. These methods are especially suitable for segmenting plant leaves and we will discuss them in greater detail in \S \ref{sec:growth-monitoring}. Taking a page from the proposal-free object detector YOLO \cite{redmon2016look},  SOLO \cite{WangXinlong-2020-solo} and SOLOv2 \cite{solov2} divide the image into grids. The grid that the center an object falls into is responsible for predicting the segmentation mask of the object.

\subsection{Uncertainty Quantification}
\label{subsec:uncertainty}
Real-world applications often require qualification of the amount of uncertainty in the predictions made by machine learning, especially when the predictions carry serious implications. 
For example, if the system incorrectly determines that fruits are not mature enough, it may delay harvesting and cause overripe fruits with diminished values.
Thus, users of the ML system are justified to ask how certain we are about the decision. In addition, when facing real-world input, it is desirable for the network to answer ``I don't know'' when facing an input that it does not recognize \cite{LiZhizhong2020}. Well-calibrated uncertainty measurements may enable such a capability.

However, research shows that deep neural networks exhibit severe vulnerability to overconfidence, or under-estimation of the uncertainty in its own decisions \cite{Guo-Weinberger-2017:Calibration,Mehrtash2020:Dice-loss-calibration}. That is, the accuracy of the network decision is frequently lower than the probability that the network assigns to the decision. As a result, proper calibration of the networks should be a concern for systems built for real-world applications. 

Calibration of deep neural networks may be performed post-doc (after training) using temperature scaling and histogram binning \cite{Guo-Weinberger-2017:Calibration,Ding_2021_ICCV,WangDengBao2021}. 
Also, regularization during training such as label smoothing \cite{Szegedy2016:Inception} and mixup \cite{zhang2018mixup} have been shown to improve calibration \cite{Muller:NEURIPS2019,pereyra2017regularizing,Thulasidasan:NEURIPS2019}. Researchers propose new loss functions to replace existing ones that are susceptible to overconfidence \cite{Mukhoti:NEURIPS2020,yeung2021calibrating}. Moreover, ensemble methods such as Vertical Voting \cite{xie2013horizontal}, Batch Ensemble \cite{wen2020batchensemble}, and Multi-input Multi-output \cite{havasi2021training} can derive uncertainty estimates.




\subsection{Interpretability}
\label{subsec:interpret}

Modern AI systems are known for its inability to provide faithful and human-understandable explanations for its own decisions. The unique characteristics of deep learning, such as network over-parameterization, large amount of training data, and stochastic optimization, while being beneficial to the predictive accuracy (e.g., \cite{LiYuanzhi:NEURIPS2018,arora2018optimization,smith2021origin,steinerhow}), all create obstacles toward understand how and why a neural network reaches its decisions. 
The lack of human-understandable explanations leads to difficulties in the verification and trust of network decisions \citep{carvalho2019machine,zhang2021survey}. 

We categorize model interpretation techniques into a few major classes, including visualization, feature attribution, instance attribution, inherently explainable models, and approximation by simple models. 
Visualization techniques present holistically what the model has learned from the training data by visualizing the model weights for direct visual inspection \citep{fong2018net2vec,bau2017network,erhan2009visualizing,szegedy2013intriguing,mahendran2015understanding,mordvintsev2015inceptionism}.  
In comparison, feature attribution and instance attribution are often considered as local explanations as they aim to explain model predictions on individual samples. 
Feature attribution methods \citep{chefer2021transformer, yeh2019fidelity,sundararajan2017axiomatic,smilkov2017smoothgrad,selvaraju2017grad,montavon2019layer,qi2021embedding,shitole2021one,ancona2017towards,chen2020generating} generate a saliency map of an image or video frame, which highlights the pixels that contribute the most to its prediction. Instance attribution methods \citep{koh2017understanding,brophy2020trex,chen2021hydra,yeh2018representer,barshan2020relatif,pruthi2020estimating,shitole2021one} attribute a network decision to training instances that, through the training process, exert positive or negative influence on the particular decision. 
Moreover, inherently explainable models \citep{sha2021learning,bastings2019interpretable,lei2016rationalizing,chen2022can,yu2021understanding} incorporate explainable components into the network architecture, which reduces the need to apply post-hoc interpretation techniques. In contrast, researchers also try to post-hoc approximate complex neural networks with simple models such as rule-based models  \citep{dhurandhar2018explanations,wang2018interpret,goyal2019counterfactual,kanamori2020dace,fu1991rule,pedapati2020learning} or linear models \cite{ribeiro2016should,Ahern2019,KOVALEV2020106164,pmlr-v139-garreau21a,pmlr-v108-garreau20a} that are easily understandable. 

The most significant benefit of interpretation in the context of CEA lies in its ability to aid with the auditing and debugging of AI systems and datasets. With feature attribution, users can make sure the system captures the robust features, or semantically meaningful features, that generalize to real-world data. As in the well-known case of husky vs. wolf image classification, due to a spurious correlation, the neural network learns to classify all images with white backgrounds as wolf and those with green backgrounds as husky \citep{molnar2020interpretable}. Such shortcut learning can be identified by feature attribution and subsequently corrected.
Moreover, instance attribution allows researchers to pinpoint outliers or incorrectly labeled training data that may lead to misclassification \citep{chen2021hydra}.

\section{Controlled-environment Agriculture}
\label{sec:cea}

Controlled-environment agriculture (CEA) is the farming practice carried out within urban, indoor, resource-controlled factories, often accompanied by stacked growth levels (\emph{i.e.}, vertical farming), renewable energy and recycling of water and waste. CEA has recently been adopted in nations around the world \citep{despommier2010vertical, benke2017future} such as Singapore \citep{krishnamurthy2014vertical}, North America \citep{Verticrop}, Japan \citep{JapanVertifarm,SpreadVertifarm}, and UK \citep{saturnbioponics}.

CEA has economic and environmental benefits. Compared to traditional farming, CEA farms produce higher yield per unit area of land \citep{saturnbioponics,SpreadVertifarm}. Controlled environments shield the plants from seasonality and extreme weather, so that plants can grow all year round given suitable lighting, temperature and irrigation \citep{benke2017future}. The growing conditions can as well be further optimized to boost growth and nutritional content \citep{kopsell2015blue,trojak2022effects}. Rapid turnover increases farmers' flexibility in plant choice to catch the trend of consumption \citep{beacham2019vertical}. Moreover, farms investment on pesticides, herbicides, and transportation can be cut down due to reduced contamination from the outside environment and proximity to urban consumers.

CEA farms, when designed properly, can become much more environmentally friendly and self-sustainable than traditional farming. With optimized growing conditions and limited external interference, the need for fertilizer and pesticides decreases, so that we can reduce the amount of chemicals that go into the environment as well as the resulting pollution.
Furthermore, CEA farms can save water and energy through the use of renewable energy and aggressive water recycling. For instance, CEA farms from Spread, a Japanese company, recycle 98\% of used water and reduce the energy cost per head of lettuce by 30\% with LED lightning \citep{SpreadVertifarm}. Finally, CEA farm can be situated in urban or suburban areas, thereby reducing transportation and storage cost. A simulation for different farm designs in Lisbon shows vertical tomato farms with appropriate designs emit less greenhouse gas than conventional farms, mainly due to reduced water use and transportation distance \citep{benis2017development}.  


A significant drawback of CEA, however, lies in its high cost, which may be partially addressed by computer vision technologies. According to \cite{benke2017future}, the higher land cost in Victoria, Australia means that the yield of vertical farms has to be at least 50 times more than traditional farming to break even. Computer vision holds the promise of boosting the level of automation and increasing yield, thereby making CEA farms economically viable. As would be discussed in the following sections, CV techniques can reduce a major amount of variable costs such as wastage cost induced by incorrect or delayed decisions on harvesting, and provide long-term benefit.


Carrying the potential to reduce a significant amount of cost, setting up computer vision systems in the field costs significantly less than expected when compared to the expenses of constructing a CEA building. Building a CEA structure involves high upfront costs, including construction, insulation, lighting, and HVAC systems. According to \citep{JapanVertifarm}, a 1,300 square meter CEA building with a production area of 4,536 square meters would require a capital investment of \$7.4 million and incur annual operational costs of approximately \$3.4 million.

On the other hand, setting up hardware systems for CV models is relatively inexpensive. The necessary components include servers (CPU, GPU, memory, storage), sensors, cameras, networking, as well as cooling system. For example, a server with specifications like a 32-Core 2.80 GHz Intel Xeon Platinum 8462Y+, 128G memory, 4 NVIDIA RTX A6000 "Ada" GPUs, and 2TB storage costs around \$60,000. Using this server for training purposes, assuming a standard VGG-16 architecture, training on 5000 images of size 224x224 pixels, with a batch size of 64 and 50 training epochs, and utilizing 4 NVIDIA A6000 GPUs, the estimated training time is less than an hour. Such a server is sufficient for daily training and inference of commonly used CV models.
For a camera system, if we consider 10 surveillance cameras such as the Hikvision DS-2CD2142FWD-I, the total cost would be around \$1400. Additionally, a high-speed network infrastructure is required to transfer data between the computer hardware, storage, and camera systems. Typically it necessitates 4 to 7 routers to cover an area of 1300 square meters, costing approximately \$2000. Finally, a liquid cooling system could cost between \$1,000 and \$2,000. In summary, a hardware system with a total cost of around \$70,000 is sufficient for the daily operation, training, and inference of CV systems.

CEA can take diverse form factors \citep{beacham2019vertical} and the form factors may pose different requirements for computer vision technologies. Typical forms for CEA are glasshouses with transparent shells or completely enclosed facilities. Depending on the cultivars being planted, internal arrangement of the farm can be classified into stacked horizontal systems, vertical growth surfaces, and multi-floor towers. 
Form factors have influence on lighting, which is an important consideration in CV applications. For example, glasshouses with transparent shells utilize natural light to reduce energy consumption but may not provide sufficient lighting for CV around the clock. In comparison, a completely enclosed facility can have greater control of lighting conditions. 
%
Moreover, internal arrangement of the farm also affect camera angle. If cultivars being planted change frequently as a result of the high turnover rate in CEAs, the arrangement of shelves and plants might change. This would affect the camera angles and thus the resulting inference performance. CV systems need adapt to the change of the environment.

Nevertheless, with the autonomous setup of CEAs, which allow easy new data collection, training a new CV model or fine-tuning a previous model to adapt to the above mentioned changeable environment would be a cinch. Besides, there are also few-shot learning \citep{wang2020generalizing, sun2019meta}, weakly-supervised learning\citep{zhou2018brief, oquab2015object, ahn2019weakly} and 
unsupervised learning techniques \citep{caron2020unsupervised, schmarje2021survey}, which require minimal or zero annotations, that can facilitate the adjustment of the models. 

Besides environmental change, there also exist other factors that need to be take into account when applying CV techniques in CEA. Two typical problems to consider would be 1) How to cope with sub-optimal data with label noise and how to address unbalanced class distribution. 
2) How to interpret the prediction from models or measure the uncertainty of prediction so that users can use the models with confidence. Quantitative measure of the confidence or uncertainty would allow farmers to understand the decision generation process and make decisions with more confidence. 
Table \ref{tab:map} map these factors to consider into CV problems, and list corresponding solutions and the respective sections that discuss the solutions.

\begin{table}[t]

\caption{Factors to consider when applying CV techniques in CEA and some corresponding countermeasures.}
    \label{tab:map}
    \centering
    
    \begin{tabular}{@{}p{5.5cm}p{5.5cm}p{4.5cm}@{}}
        \toprule
        Factors & CV Problems & Example Countermeasures \\
        \midrule
        Environmental Change & OOD Generalization & Collect New Data, Few-shot learning, Weakly-supervised learning, Unsupervised-learning (see \S \ref{sec:cea}) \\

        Sub-optimal Data Quality & Unbalanced Class Distribution, Lable Noise & Multiple-Instance Learning, Generate Image of Minority Classes with GANs, Few-shot Meta-learning (see \S \ref{problem:unbalance}, \S \ref{problem:noise} and \S \ref{subsec:data problem})\\
        Human Factor & Interpretability, Uncertainty Estimates & Paired Confidence Scores, Meta-learning (see \S \ref{subsec:uncertainty}, \S \ref{subsec:interpret} and \S \ref{subsec:uncertainty and interpretability}) \\
        \bottomrule
    \end{tabular} 
\end{table}

In the following, we investigate the application of autonomous computer vision techniques on Growth Monitoring, Fruit and Flower Detection, Fruit Counting, Maturity Level Classification and Pest and Disease Detection to increase production efficiency. In addition to existing applications, we also include techniques that can be easily applied to vertical farms even though they have not yet been applied to them. 






\subsection{Growth Monitoring} 
\label{sec:growth-monitoring}

Growth monitoring, a critical component of plant phenotyping, aims to understanding the life cycle of plants and estimating yield \citep{iljazi2017deep} by monitoring various growth indicators such as the plant size, number of leaves, leaf sizes, land area covered by the plant, and so on. Plant growth monitoring facilitates in quantifying the effects of biological / environmental factors on growth and thus is crucial for finding the optimal growing condition and developing high-yield crops \citep{nassar2018compliant, tang2019rapid}. 

As early as 1903, Wilhelm Pfeffer has recognized the potential of image analysis in monitoring plant growth \citep{pfeffer1900physiology, spalding2013image}. Traditional machine vision techniques such as gray-level pixel thresholding \citep{otsu1979threshold}, Bayesian statistics \citep{bouman1994multiscale} and shallow learning techniques \citep{yu2011modified, ireri2019computer}, have been applied to segment the objects of interest, such as leaves and stems, from the background to analyze plant growth. 
Compared to traditional methods, deep-learning techniques provide automatic representation learning and are less sensitive to image quality variations. For this reason, deep learning techniques for growth monitoring have recently gained popularity.


Among various growth indicators, leaf size and number of leaves per plant are the most commonly used \cite{lancashire1991uniform, scharr2016leaf, gustafson1936some,iljazi2017deep}. Therefore, in the section below, we first discuss leaf instance segmentation, which can support both indicators at the same time, followed by a discussion of techniques for only leaf counting or for other growth indicators.


\subsubsection{Leaf Instance Segmentation}
\label{subsubsec: leaf}
\begin{table}[t]

\caption{Performance of various leaf instance segmentation techniques on the CVPPP A1 test set. Higher SBD and lower |DiC| indicate better performance. (GT-FG) indicates model making use of ground-truth foregrounds}
    \label{tab:leafseg}
    \centering
    \begin{tabular}{@{}clcc@{}}
        \toprule
        Category & Technique & SBD ($\uparrow$) & |DiC| ($\downarrow$) \\
        \midrule
        \multirow[c]{2}{*}{Sequential} & End-to-end instance segmentation \citep{ren2017end} & 84.9  & 0.8 \\
        & RNN-SIS \citep{salvador2017recurrent} & 74.7 & 1.1\\
        & RIS \citep{romera2016recurrent} & 66.6 & 1.1 \\
        \midrule
        \multirow[c]{5}{*}{Pixel Embedding} 
        
        &   Semantic Instance Segmentation
        \citep{Brabandere-InstSeg-DiscriminativeLoss2017} & 84.2 & 1.0 \\
        & Object-aware Embedding \citep{ChenLong-Object-aware-Embedding:2019} & 83.1 & \textbf{0.73} \\
        & RHN + Cosine Embeddings \citep{Payer-Instance-Segmentation:2018} & 84.5 & 1.5 \\
        & Crop Leaf and Plant Instance Segmentation \citep{weyler2022field} & 91.1 & 1.8 \\
        & W-Net (GT-FG) \citep{wu2020improving} & 91.9 & - \\
        & SPOCO (GT-FG) \citep{wolny2022sparse} & \textbf{93.2} & 1.7 \\
        \bottomrule
    \end{tabular} 
\end{table}

Due to the popularity of the CVPPP dataset \cite{minervini2016finely}, the segmentation of leaf instance has attracted special attention from the computer vision community and warrants its own section. 
%
%
leaf instance segmentation methods include recurrent network methods \citep{ren2017end, romera2016recurrent} and pixel embedding methods \citep{weyler2022field,wu2020improving,ChenLong-Object-aware-Embedding:2019,Payer-Instance-Segmentation:2018,Brabandere-InstSeg-DiscriminativeLoss2017}. Parallel proposal methods are popular for general-purpose segmentation (see \S {segmentation}), but are ill-suited for leaf segmentation. As most leaves have irregular shapes, the rectangle proposal boxes used in these methods do not fit the leaves well, resulting in many poorly positioned boxes. In addition, the density of leaves causes many proposal boxes to overlap and compounds the fitting problem. As a result, it is difficult to pick out the best proposal box from the large number of parallel proposals. 
Therefore, we focus on recurrent network based methods and pixel embedding based methods in this section. Quality metrics for leaf segmentation include Symmetric Best Dice (SBD) and Absolute Difference in Count (|DiC|). SBD calculates the average overlap between the predicted mask and the ground truth for all leaves. DiC calculates the average number of miscalculated leaves over the entire test set.



Recurrent network based methods output a mask for a single leaf sequentially. Their decisions are usually informed by the already segmented parts of the image, which are summarized by the recurrent network.
\citep{ren2017end} applies LSTM and DeconvNet to segment one leaf at a time. The network first locates a bounding box for the next leaf, and performs segmentation within that box. After that, leaves segmented in all previous iterations are aggregated by the recurrent network and passed to the next iteration as contextual information.
\citep{romera2016recurrent} employs convolution-based LSTMs (ConvLSTM) with FCN feature maps as input. At each time step, the network outputs a single-leaf mask and a confidence score. During inference, the segmentation stops when the confidence score drops below 0.5. 
\citep{salvador2017recurrent} proposes another similar method that combines feature maps with different abstraction levels for prediction.


Pixel embedding methods learn vector representations for the pixels so that pixels in irregularly shaped leaves can become regularly shaped clusters in the representation space. With that, we can directly cluster the pixels.
\citep{weyler2022field} performs simultaneous instance segmentation of leaves and plants. 
The authors propose an encoder-decoder framework, based on ERFNet \citep{romera2017erfnet}, with two decoders. One decoder predicts the centroids of plants and leaves. The other decoder predicts the offset of each leaf pixels to the leaf centroid. The pixel location plus the offset vector hence should be very close to the leaf centroid. The dispersion among all pixels of the same leaf can be modeled as a Gaussian distribution, whose covariance matrix is also predicted by the second decoder and whose mean is from the first decoder. The training maximizes the Gaussian likelihood for all pixels of the same leaf. The same process is applied to pixels of the same plant. 

\citep{wu2020improving, ChenLong-Object-aware-Embedding:2019, Payer-Instance-Segmentation:2018} are three similar pixel embedding methods. They encourage pixels from the same leaf to have similar embeddings and pixels from different neighboring leaves to have different embeddings to enable clustering in the embedding space.
Their network consists of two modules, the distance regression module and pixel embedding module. \citep{wu2020improving, Payer-Instance-Segmentation:2018} arrange the two modules in sequence, while \citep{ChenLong-Object-aware-Embedding:2019} places them in parallel. The distance regression module predicts the distance between the pixel and the closest object boundary. The pixel embedding module generates an embedding vector for each pixel, so that pixels from the same leaves have similar embeddings and pixels from different neighboring leaves have different embeddings. During inference, pixels are clustered around leaf centers, which are identified as local maxima in the distance map from the distance regression module. 
%

Lastly, \cite{Brabandere-InstSeg-DiscriminativeLoss2017, wolny2022sparse} take a large-margin approach. They ensure that embeddings of pixels from the same leaf are within a circular margin of the leaf center, and the embedding of leaf centers are far away from each other. This removes the need to determine the leaf centroids during inference because the embeddings are already well separated. 
\citep{wolny2022sparse} built upon the method in \cite{Brabandere-InstSeg-DiscriminativeLoss2017} to perform pixel embedding and clustering of leaves under weak supervision, with annotation on only a subset of instances in the images. In addition, a differentiable instance-level loss for a single leaf is formed to overcome the non-differentiability of assigning pixels to instances by comparing a Gaussian shape soft mask with the corresponding ground truth mask. Finally, consistency regularization, which encourages accordance of two embedding frameworks, is applied to improve embedding for unlabeled pixels.

Comparing different approaches, proposal-free pixel embedding techniques seem to be the best choice for the leaf segmentation problem. As can be seen from Table \ref{tab:leafseg}, pixel embedding methods obtain both the highest SBD and lowest |DiC|. One thing to note here, however, is that superior result of W-Net \citep{wu2020improving} and  SPOCO \citep{wolny2022sparse} could be attributed to the inclusion of ground-truth foreground masks during inference. 
Even though the recurrent approach does not generate a large number of proposal boxes at once, it still uses rectangular proposals, which means that it still suffers from the fitting problem to irregular leaf shapes. Moreover, the recurrent methods are usually slower than pixel embeddings, due to the temporal dependence between the leaves.

\subsubsection{Leaf Count and Other Growth Metrics}
\label{subsubsec: othermetrics}

Leaf counts may be estimated without leaf segmentation. 
\citep{ubbens2018use} utilizes synthetic data in the leaf counting task. The authors employ the L-system-based plant simulator \emph{lpfg} \citep{prusinkiewicz2002art,AlgorithmicBotany} to generate  Arabidopsis rosette images. The authors test a CNN, trained with only synthetic data, on real data from CVPPP and obtain superior result than a model trained with CVPPP data only. In addition, CNN trained with the combination of synthetic and real data obtained approximately 27\% reduction in the mean absolute count error compared to CNN using only real data. These results demonstrate the potential of synthetic data in plant phenotyping.

Besides leaf size and leaf count, leaf fresh weight, leaf dry weight, and plant coverage (the area of land covered by the plant) are also used as metrics of growth.
\citep{zhang2020growth} applies CNN to regress leaf fresh weight, leaf dry weight, and leaf area of lettuce on RGB images. \citep{reyes2020real} makes use of Mask R-CNN, a parallel proposal method, for lettuce instance segmentation. The authors derive plant attributes such as contour, side view area, height, and width from the segmentation masks and bounding boxes, using preset formulas. They also estimate growth rate from the changes in area of the plant at each time step; they estimate fresh weight by linearly regressing from the attributes. 
\citep{lu2019monitoring} leverages COCO dataset pretrained Mask R-CNN with ResNet-50 as backbone to segment lettuce leaves. The daily change of mean leaf area is used for growth rate calculation.

\subsection{Fruit and Flower Detection}
\label{subsec:detect}

\begin{table}[t]

\caption{Performance of various fruit and flower detection techniques. Datasets without reference are unpublished datasets. }
    \label{perfomancetab:fruitflowerdetect}
    \centering
    \begin{tabular}{@{}p{2cm}cp{4.5cm}p{2cm}p{5.5cm}@{}}
        \toprule
        Category & Technique &   Evaluation Metric  & Performance & Dataset\\
        \midrule
        \multirow{5}{*}[-1em]{\shortstack[l]{\textbf{Fruit}\\\textbf{Object Detection}}}&  \citep{yuan2020robust} &  Precision (IoU > 0.5) & 94\% & 1730 images of cherry tomatoes \\
        &  \citep{hu2019automatic}   & Accuracy (IoU unspecified) & 95.50\% &  800 images of tomatoes \\
        &  \citep{sa2016deepfruits} & F1 scores (IoU unspecified) & 83.80\% & 122 images of 7 fruits \\
        &  \citep{zhang2019multi}  & True positive rate and False positive rate (IoU unspecified) & 98\%, 17\% &  2116 self-acquired images of fruits and 511 images of fruits from ImageNet \\
        &  \citep{9119372} &  Precision and Recall (IoU > 0.9) & 94.4\%, 93.5\% &  2000 images of strawberries \\
        &  \citep{shi2020attribution}  & F1 scores (IoU unspecified) & 93.5\%-95.1\% & Mango Image Dataset \cite{koirala2019deep} \\
        \midrule
        \multirow{5}{*}[-1em]{\shortstack[l] {\textbf{Fruit}\\\textbf{Segmentation}}} &  \citep{lin2019guava}& Precision and Recall (IoU unspecified) & 98.3\%, 94.8\% &  437 RGB-D images of guavas\\
        &  \citep{afonso2020tomato} & Precision, Recall and F1 scores (IoU > 0.5) & 96\%, 91\%, 93\% &  123 images RGB-D images of tomatoes \\
        &  \citep{8863343} & Precision, Recall, F1 score and Average Precision (IoU > 0.9) & 97\%, 92\%, 94\%, 90\% &  120 images RGB-D images of strawberries\\
        &  \citep{yu2019fruit} & Mean IoU & 89.85\% &  1900 images of strawberries\\
        &  \citep{huang2020using} & Accuracy (IoU unspecified) & 98\% &  900 images of strawberries\\
        \midrule
        \multirow{3}{*}[-1em]{\shortstack[l] {\textbf{Flower}\\\textbf{Object Detection}}}&  \citep{lyu2022embedded}  & Average Precision and F1 scores (IoU > 0.5)  & 96.2\%, 89.0\%  &  1078 images of citrus buds and flowers\\
        &  \citep{sun2018detection} & Average Precision (IoU > 0.5) & 90.50\% &  5624 images of tomato flower and fruit\\
        &  \citep{sun2021apple}  & IoU, F1 scores, Recall and Precision (IoU unspecified) & 81.1\%, 89.6\%, 91.9\%, 87.3\%  & Multi-species fruit flower detection \cite{dias2018multispecies}\\
        \bottomrule
    \end{tabular} 
\end{table}


Algorithms for fruit and flower detection find the location and spatial distribution of fruits and fruit flowers. This task supports various downstream applications such as fruit count estimation, size estimation, weight estimation, robotic pruning, robotic harvesting, and disease detection \citep{gene2020fruit, bargoti2017image,yeshitela2005effects,lyu2022embedded}.
In addition, fruit or flower detection may help devise plantation management strategies \citep{hao2015response, gene2020fruit} because fruit or flower statistics such as positions, facing directions (the directions the flowers face), and spatial scatter can reveal the status of the plant and the suitability of environmental conditions. For example, the knowledge of flower distribution may allow pruning strategies that focus on regions of excessive density and achieve even distribution of fruits which optimize the delivery of nutrient to the fruits. 


Traditional approaches for fruit detection rely on manual feature engineering and feature fusion. As fruits tend to have unique colors and shapes, one natural thought is to apply thresholding on color \citep{wei2014automatic,ostovar2018adaptive} and shape information \citep{liu2018detection,nyarko2018nearest}. Additionally, \citep{chaivivatrakul2010towards, moonrinta2010fruit,lin2020color} employ a combination of color, shape, and texture features. However, manual feature extraction suffers from brittleness when the image distribution changes with different camera resolutions, camera angles, illumination, and species \citep{bargoti2017deep}.

Deep learning methods for fruit detection include object detection and segmentation. 
\citep{yuan2020robust} applies SSD for cherry tomato detection.
\citep{hu2019automatic} leverages Faster R-CNN to detect tomatoes. Inside the generated bounding boxes, color thresholding and fuzzy-rule-based morphological processing methods are applied to remove image background and obtain the contours of individual tomatoes. 
\citep{sa2016deepfruits} leverages Faster R-CNN with VGG-16 as the backbone for sweet pepper detection. RGB and near-infrared (NIR) images are used together for detection. Two fusion approaches, early and late fusion, are proposed. Early fusion alters the first pretrained layer to allow 4 input channels (RGB and NIR), whereas late fusion aggregates the two modalities by training independent proposal models for each modality and then combining the proposed boxes by averaging the predicted class probabilities.
\citep{zhang2019multi} trains three multi-task cascaded convolutional networks (MTCNN) \citep{zhang2016joint} for detecting apples, strawberries and oranges. MTCNN contains a proposal network, a bounding box refinement network, and an output network in a feature pyramid architecture with gradually increased input sizes for each network.  The model is trained on synthetic images, which are random combinations of cropped negative patches and fruits patches, in addition to real-world images. 
\citep{9119372} proposed R-YOLO with MobileNet-V1 as the backbone to detect ripe strawberries. Different from regular horizontal bounding boxes in object detection, the model generates rotated bounding boxes by adding a rotation-angle parameter to the anchors.

Delicate fruits, such as strawberries and tomatoes, are particularly vulnerable to damage during harvesting. Therefore, much research has been devoted to segmenting such fruits from backgrounds in order to determine the precise picking point. Precise fruit masks are expected to enable robotic fruit picking while avoiding damages on the neighboring fruits. 
\citep{lin2019guava} performs semantic segmentation for guava fruits and determines their poses using FCN with RGB-D images as input. The FCN outputs a binary mask for fruits and another binary mask for branches. With the fruit binary mask, the authors employ Euclidean clustering \citep{rusu2010semantic} to cluster single guava fruit. From the clustering result and the branch binary mask, fruit centroids and the closest branch are located. Finally, the system predicts the vertical axis of the fruit as the direction perpendicular to the closest branch to facilitate robotic harvesting. Similarly, \citep{afonso2020tomato} leverages Mask R-CNN with ResNet as backbone for semantic segmentation of tomatoes. In addition, the authors filter the false positive detection of tomatoes from the non-targeted rows by setting a depth threshold. \citep{8863343} utilizes Mask R-CNN with a ResNet101 backbone to perform instance segmentation of ripe strawberries, raw strawberries, straps and tables. Depth images are aligned with the segmentation mask to project the shape of strawberries into 3D space to facilitate automatic harvesting. \citep{yu2019fruit} also applies Mask R-CNN with a ResNet101 + FPN backbone to perform instance segmentation and ripeness classification on strawberries. \citep{huang2020using} leverages a similar network for instance segmentation of tomatoes. With the segmentation mask, the systems determine the cut points of the fruits.


Besides accuracy, the processing speed of neural networks is also important for their deployment on mobile devices or agricultural robots.
\citep{shi2020attribution} performs network pruning on YOLOv3-tiny to form a lightweight mango detection network. A YOLOv3-tiny pretrained on the COCO dataset has learned to extract fruit-relevant features because the COCO dataset contains apple and orange images, but it also has learned irrelevant features. The authors thus use a generalized attribution method \citep{shrikumar2016not} to determine the contribution of each layer to fruit features extraction and remove convolution kernels responsible for detecting non-fruit classes. They find that the lower level features are shared across all classes detection and pruning in the higher layers does not harm fruit detection performance. After pruning, the network achieves significantly lowers float-point operations (FLOPs) at the same level of accuracy. 


Object detection is also applied for flower detection. 
\citep{lyu2022embedded} proposes a modified YOLOv4-Tiny with cascade fusion (CFNet) to detect citrus buds, citrus flowers, and gray mold, which is a disease commonly found on citrus plants. The authors propose additionally a block module with channel shuffle and depth separable convolution for YOLOv4-Tiny.
\citep{sun2018detection} shrinks the anchor boxes of Faster-RCNN to fit small fruits and applies soft non-maximum suppression to retain boxes that may contain occluded objects.
%
As flowers usually have similar morphological characteristics, flowers from other non-targeted species could possibly be used as training data in a transfer learning scenario. In \citep{sun2021apple}, the authors fine-tune a DeepLab-ResNet model \citep{chen2017deeplab} for fruit flower detection. The model is trained on apple flower dataset but achieves high F1 scores on pear and peach flower images (0.777 and 0.854 respectively).

\subsection{Fruit Counting}
\label{plant count}

\begin{table}[t]

\caption{Performance of various fruit counting techniques. Datasets without reference are unpublished datasets. \citep{rahnemoonfar2017deep} uses direct regression method thus does not need IoU threshold}
    \label{perfomancetab:fruitcount}
    \centering
    \begin{tabular}{@{}p{2.5cm}cp{5cm}p{2cm}p{4cm}@{}}
        \toprule
        Category & Technique &   Evaluation Metric  & Performance & Dataset\\
        \midrule
        \multirow{1}{*}[0em]{\shortstack[l]{\textbf{Count Regression}}}&  \citep{rahnemoonfar2017deep} &  Accuracy & 91.0\% - 93\% & 4,800 synthetic tomato images \\
        \midrule
        \multirow{2}{*}[0em]{\shortstack[l] {\textbf{Count Fruit}\\\textbf{Bounding Boxes}}} &  \citep{koirala2019deep}& F1 scores, Average Precision(IoU > 0.24) & 96.8\%, 98.3\% &  MangoYolo Dataset \cite{koirala2019deep}\\
        &  \citep{wang2019mango} & $R^{2}$, RMSE & 0.66, 2.1 & MangoYolo Dataset \cite{koirala2019deep}\\
        \midrule
        \multirow{2}{*}[0em]{\shortstack[l] {\textbf{Count Fruit}\\\textbf{Segmentation Masks}}}&  \citep{kestur2019mangonet}  & Accuracy, F1 score (IoU > 0.6)  & 73.6\%, 84.4\%  & 12,590 images of mangoes\\
        &  \citep{ni2020deep} & Average Precision (IoU > 0.5), RMSE & 71.6\%, 1.484 &  724 images of blueberries\\

        \bottomrule
    \end{tabular} 
\end{table}

Pre-harvest estimation of yields plays an important role in the planning of harvesting resources and marketing strategies  \citep{yang2021applications,he2022fruit}. As fruits are usually sold to consumers as a pack of uniformly sized fruits or individual fruits, the fruit count also provides an effective yield metric \citep{koirala2019deep}, besides the distribution of fruit sizes.
Traditional yield estimation is obtained through manual counting of samples from a few randomly selected areas \citep{he2022fruit}. Nonetheless, when the production is large-scale, to counteract the effect of plant variability, accurate estimation would require a large quantity of samples from different areas of the field, resulting in high cost. Thus, researchers resort to CV-based counting methods. 


A direct counting method is to regress on the image and output the fruit count. In \citep{rahnemoonfar2017deep}, the authors apply a modified version of Inception-ResNet for direct tomato counting. The authors train the model on simulated images and test on real images, which suggest, once again, the viability of using simulated images to circumvent the cost for formulating a large dataset.

Besides direct regression, object detection \citep{koirala2019deep,wang2019mango}, semantic segmentation \citep{kestur2019mangonet}, and instance segmentation \citep{ni2020deep} have also been used for fruit counting. These methods provide an intermediate level of results from which the count can be easily gathered. \citep{koirala2019deep} proposes MangoYOLO based on YOLOv2-tiny and YOLOv3 for mango detection and counting. 
The authors increase the resolution of the feature map to facilitate detection of small fruits. 
\citep{halstead2018fruit} proposes pre-trained Faster R-CNN network, building upon DeepFruits \cite{sa2016deepfruits}, to estimate the quantity of sweet pepper. The authors design a tracking sub-system for sweet pepper counting. The sub-system identifies new fruits by measuring the IoU between and comparing the boundary of detected and new fruits.
\citep{kestur2019mangonet} performs semantic segmentation for mango counting using a modification of FCN. The coordinates of blob-like regions in the semantic segmentation mask is used to generate bounding boxes corresponding to mango fruits. Finally, \citep{ni2020deep} applies Mask R-CNN to for instance segmentation of blueberries. The model also classifies the maturity of individual blueberries and counts the number of berries according to the masks.

Occlusion poses a difficult challenge for counting. Due to this issue, automatic count from detection or segmentation results is almost always lower than the actual number of fruits. To solve this, \citep{koirala2019deep} calculates and applies the ratio between the actual hand harvest count and the automatic fruit count; it also uses both front and back views of mango trees to mitigate occlusion from one angle. Taking this idea one step further, \citep{wang2019mango} uses dual-view videos to detect and track mangoes when the camera moves. Utilizing different views of the same tree in a video, 
\citep{wang2019mango} recognizes around 20\% more fruits. However, the detected count is still significantly lower than the actual number, underscoring the research challenge of exhaustive and accurate counting.

\subsection{Maturity Level Classification}
\label{maturity}

\begin{table}[t]

\caption{Performance of various maturity level classification techniques. Datasets without reference are unpublished datasets. Performance "-" are papers with unsummarizable metric results. \citep{8520836} uses direct classification method thus does not need IoU threshold}
    \label{perfomancetab:matureclass}
    \centering
    \begin{tabular}{@{}p{2.5cm}cp{5.3cm}p{2cm}p{4cm}@{}}
        \toprule
        Category & Technique &   Evaluation Metric  & Performance & Dataset\\
        \midrule
        \multirow{1}{*}[0em]{\shortstack[l]{\textbf{Classification}}}&  \citep{8520836} &  Accuracy & 91.9\% & 200 images of tomatoes \\
        \midrule
        \multirow{2}{*}[0em]{\shortstack[l] {\textbf{Classification on }\\\textbf{Bounding Boxes}}} &  \citep{9119372}& Precision, Recall (IoU > 0.9) & 94.4\%, 93.5\% &  2000 images of strawberries\\
        &  \citep{halstead2018fruit} & F1 score (IoU > 0.4) & 77.30\% & 285 images of capsicums\\
        \midrule
        \multirow{4}{*}[-1em]{\shortstack[l] {\textbf{Classification on} \\\textbf{Segmentation Masks}}}&  \citep{afonso2020tomato}  & Precision, Recall and F1 scores (IoU > 0.5)  & - & 123 images RGB-D images of tomatoes\\
        &  \citep{8863343} & Precision, Recall, F1 score and Average Precision (IoU > 0.9) & - &  120 images RGB-D images of strawberries\\
        &  \citep{yu2019fruit} & Precision, Recall (IoU > 0.9) & 95.78\%, 95.41\% &  1900 images of strawberries\\
        &  \citep{huang2020using} & Class frequency weighted precision and recall (IoU Unspecified) & 96.1\%, 96.0\% & 900 images of strawberries\\

        \bottomrule
    \end{tabular} 
\end{table}

Maturity level classification aims to determine the ripeness of fruits or vegetables to aid in proper harvesting and food quality assurance. Premature harvesting results in plants that are unpalatable or incapable of ripening, while delayed harvesting can result in overripe plants or food decay \citep{huang2020using}. 

The optimal maturity level differs for different targeted products and destinations. Fruits and vegetables can be consumed at different growing stages. For example, lettuce can be consumed either as baby lettuce or fully grown lettuce. The same situation happens with baby corn and normal corn. Products are to be transported to different destinations, so we must consider the length of transportation and ripening speed when deciding the correct maturity level at harvest \citep{zhang2018deep}. 

Manually distinguishing the subtle differences in maturity levels is time-consuming, prone to inconsistency, and costly. The labor cost of harvesting accounts for a large percentage of operation cost in farms, with 42\% of variable production expenses in U.S. fruit and vegetable farms being spent on labor for harvesting \citep{huffman2012status}. Automatic maturity level classification with computer vision, in contrast, can 
assist automatic harvesting \citep{zhang2018deep,8863343,altaheri2019date} and reduce cost.

Similar to fruit detection, we can apply thresholding methods on color to detect ripeness. For example, \citep{ arefi2011recognition} applies color thresholding on HSI and YIQ color spaces. \citep{ teixido2012definition} applies linear color models. \citep{li2016immature} utilizes the combination of color and texture features. \citep{wu2019automatic,fernandez2014multisensory,seng2009new,senthilnath2016detection,kurtulmus2014immature} apply shallow learning methods based on a multitude of features. 

More recently, researchers evaluate the performance of deep learning based computer vision methods on maturity level classification and attain satisfactory results.
For example, \citep{8520836} applies CNN to classify tomato maturity into five levels. However, to further facilitate automatic harvesting, object detection and instance segmentation are more commonly used for getting the exact shape, location and maturity level of fruits, and position of peduncles for robotic end-effectors to cut on. 

With object detection, \citep{9119372} applies the R-YOLO network described in the fruit detection section (\S \ref{subsec:detect}) to detect ripe strawberries.  \citep{halstead2018fruit}, as mentioned in the fruit counting section \S \ref{plant count}, proposes pre-trained Faster R-CNN network to estimate both the ripeness and quantity of sweet pepper. Two formulations of the model are tested. One treats ripe/unripe as additional classes on top of foreground/background, and the other performs foreground/background classification first and then performs ripeness classification on foreground regions. The second approach generates better ripeness classification results as the ripe/unripe classes are more balanced when only the foreground regions are considered. 

Using the segmentation methods discussed in \S \ref{subsec:detect}, \citep{afonso2020tomato} classifies semantic segmentation masks of tomatoes into raw and ripe tomatoes.  \citep{8863343,yu2019fruit} performs instance segmentation and classifies instance masks into ripe and raw strawberries. \citep{huang2020using} performs instance segmentation on tomatoes first. After transforming the mask region into HSV color space, the authors employ a fuzzy system to classify tomatoes into four classes: immature (completely green), breaker (green to tannish), preharvest (light red), and harvest (fully colored).

\subsection{Pest and Disease Detection}
\label{subsec:ill}

\begin{table}[t]

\caption{Performance of various pest and disease detection techniques. Datasets without reference are unpublished datasets. Performance "-" are papers with unsummarizable metric results. *Studies perform direct classification on image thus do not need IoU threshold. \citep{gozzovelli2021tip} uses patch level segmentation which does not need IoU threshold as well.}
    \label{perfomancetab:pestdiseasedetect}
    \centering
    \begin{tabular}{@{}p{1.5cm}cp{3.5cm}p{2.5cm}p{6.5cm}@{}}
        \toprule
        Category & Technique &   Evaluation Metric  & Performance & Dataset\\
        \midrule
        \multirow{8}{*}[0em]{\shortstack[l]{\textbf{Single- and}\\ \textbf{Multi-label} \\\textbf{Classification}}}&  \citep{zhang2019cucumber} &  Accuracy* & 94.65\% & 700 diseased and normal leaf images \\
        &  \citep{singh2019multilayer} &  Accuracy* & 97.13\% & 1070 self acquired leaf images, and 1130 images from the Plant Village dataset \cite{hughes2015open} \\
        &  \citep{aravind2019disease} &  Accuracy* & 93.33\% & Images of 643 leaf samples \\
        &  \citep{selvaraj2019ai} &  mAP* & 72.8\% - 97.9\% & 12,600 images of bananas\\
        &  \citep{ferentinos2018deep} &  Accuracy* & 99.50\% & 87,848 images of leaves \\
        &  \citep{ma2018recognition} &  Accuracy* & 93.40\% & Plant Village dataset \cite{hughes2015open} \\
        &  \citep{fuentes2017robust} &  mAP (IoU > 0.5) & 86\% & 5000 images of diseases and pests of tomatoes  \\
        &  \citep{yuan2022improved} &  mIOU, recall, and F1-score (IoU unspecified) & 84.8\%, 88.1\%, 91.8\% & Plant Village dataset \cite{hughes2015open} \\
        \midrule
        \multirow{4}{*}[0em]{\shortstack[l] {\textbf{Handling}\\\textbf{Unbalanced}\\\textbf{Class}\\\textbf{Distribution}}} &  \citep{bollis2020weakly}& Accuracy* & 60.7\% - 91.8\% &  IP102 \cite{wu2019ip102}, Citrus Pest Benchmark \cite{bollis2020weakly} \\
        &  \citep{gozzovelli2021tip} & Average Precision & 67\% - 85\% & Plant Village dataset \cite{hughes2015open} and Plant Leaves \cite{siddharth2019database}\\
        &  \citep{nuthalapati2021multi} & Accuracy* & 88.5\% - 95.5\% & Plant Village dataset \cite{hughes2015open} and Plant and Pest \cite{li2021meta}\\\
        &  \citep{li2021meta} & Accuracy* & 43.9\% - 81\% &  Plant Village \citep{hughes2015open}, Crop Pests Recognition \citep{li2020crop}\\
        \midrule
        \multirow{2}{*}[0em]{\shortstack[l] {\textbf{Noise and }\\\textbf{Uncertainty}\\\textbf{Estimate}}} &  \citep{shi2020rectified}  & Accuracy* & 94.58\% & Plant Village \citep{hughes2015open}\\
        &  \citep{frank2021confidence} & Accuracy* & - &  15,892 images of tomatoes from Plant Village \citep{hughes2015open}, extra 8911 images of corns, 6,635 images of soybeans\\

        \bottomrule
    \end{tabular} 
\end{table}
Plants are susceptible to environmental disorders caused by temperature, humidity, nutritional excess/deficiency, light changes and biotic disorders due to fungi, bacteria, virus or other pests \citep{fuentes2017robust, singh2019multilayer}. Infectious diseases or pest pandemic induce inferior plant quality or plant death, resulting in at least 10\% of global food production losses \citep{strange2005plant}. 

Although controlled vertical farming restricts the entry of pests and diseases, it cannot eliminate them. Pests and diseases can enter the farm from accidental contamination from employees, seeds, irrigation water and nutrient solution, poorly maintained environment or phytosanitation protocols, unsealed entrance and ventilation systems \citep{roberts2020vertical}. For this reason, pest and disease detection is still worth studying in the context of CEA. 

Manual diagnosis of plant is complex due to the large quantity of vertically arranged plants in the field and numerous possible symptoms of diseases on different species. In addition, plants show different patterns along infection cycles and their symptoms can vary in different part of the plant \citep{bock2008visual}. Consequently, autonomous computer vision systems that recognize diseases according to the species and plant organs are gaining traction. 
From a technological perspective, we sort existing techniques into three parts, single- and multi-label classification, handling unbalanced class distributions, as well as label noise and uncertainty estimates.


\subsubsection{Single- and Multi-label Classification}

Studies perform single-label, or one-label-per-image, classification of diseases of either one single species \citep{zhang2019cucumber,singh2019multilayer,aravind2019disease,selvaraj2019ai} or multiple species \citep{ferentinos2018deep}. 
\citep{zhang2019cucumber} creates a lightweight version of AlexNet, replacing the fully connected network with a global pooling layer, to classify six types of cucumber diseases.  \citep{singh2019multilayer} leverages CNNs for classifying leaves into mango leaves, diseased mango leaves and other plant leaves. \citep{aravind2019disease} utilizes AlexNet and VGG16 to recognize five types of pests and diseases of tomatoes. 
\citep{ferentinos2018deep} applies AlexNet, AlexNetOWTBn \citep{krizhevsky2014one}, GoogLeNet, Overfeat \citep{sermanet2013overfeat}, and VGG for classifying 25 different healthy or diseased plants.
%


Having a single label per image can be inaccurate. In the real world, one plant or one leaf can carry multiple diseases or contain multiple diseased regions.  By detecting multiple targeted areas or disease classes, the multi-label setting can lead to improved efficiency and accuracy.


To deal with the possibility of having multiple diseases or multiple areas of diseases on one plant simultaneously, two types of methods are proposed. \citep{ma2018recognition} first segments out different infection areas on cucumber leaves using color thresholding following \cite{ma2017segmentation}, then applies DCNN on segmented areas to classify four types of cucumber diseases. Nevertheless, the color thresholding technique may not generalize to other plant species and environment. 
Another type of method leverages object detection or segmentation for locating and classifying infection areas. \cite{selvaraj2019ai} locates multiple diseased regions of banana plants simultaneously using object detection but assigns only one disease label to each image. \citep{fuentes2017robust} compared Faster R-CNN, R-FCN and SSD for detecting nine classes of diseases and pests that affect tomato plants. Multiple diseases and pests in one plant are detected simultaneously.
\citep{yuan2022improved} applies improved DeepLab v3+ for segmentation of multiple black rot spots on grape leaves. The efficient channel attention mechanism \citep{9156697} is added to the backbone of DeepLab v3+ for capturing local cross-channel interaction. Feature pyramid network and Atrous Spatial Pyramid Pooling \citep{chen2018encoder} are utilized for fusing feature maps from the backbone network at different scales to improve segmentation.


\subsubsection{Handling Unbalanced Class Distributions}
\label{problem:unbalance}
A common obstacle encountered in disease detection is unbalanced disease class distributions. There are typically much fewer diseased plants than healthy plants; the unequal frequencies introduce difficulties in finding images of rare diseases; the data unbalance leads to difficulty for model training. To remedy such problem, researchers propose weakly supervised learning \citep{bollis2020weakly}, generative adversarial network (GAN) \citep{gozzovelli2021tip}, and few-shot learning \citep{nuthalapati2021multi, li2021meta}.

Specifically, \citep{bollis2020weakly} applies multiple instance learning (MIL), a type of weakly supervised learning method, for multi-class classification of six mite species of citrus. In MIL, the learner receives a set of labeled bags, containing multiple image instances. We know that at least one instance is associated with the class label, but do not know the exact instance. The MIL algorithm tries to identify the common characteristic shared by images in the positively labeled bags. In this work, a CNN is first trained with labeled bags. Next, by calculating saliency maps of images in bags, the model identifies salient patches that have a high probability of containing mites. These patches inherit labels from their bags and are used to refine the CNN trained above. 

\citep{gozzovelli2021tip} leverages generative adversarial network (GAN) to generate realistic image patches of tip-burn lettuce and trains U-net for tip-burn segmentation. For the generation stage, lettuce canopy image patches are inputted into Wasserstein GANs \citep{arjovsky2017wasserstein} to generate stressed (tip-burned) patches so that there are an equal number of stressed and healthy patches. Then, in the segmentation stage, the authors generate a binary label map for the images using a classifier and an edge map. The binary label map labels each mini-patches (super-pixels) as stressed or healthy. The authors then feed the label map, alongside the original images, as input to U-net for mask segmentation. 

In few-shot meta-learning, we are given a meta-train set and a meta-test set, with the two sets containing mutually exclusive image classes (i.e. classes in the training set do not appear in the testing set). Meta-train or meta-test sets contain a number of episodes, each of which consists of some training (supporting) images and some test (query) images. The rationale of meta-learning is to equip the model with the ability to quickly learn to classify the test images from a small number of training images within each episode. The model acquires this meta-learning capability on the meta-train set and is evaluated on the meta-test set. 


As an example ,
\citep{nuthalapati2021multi} performs pests and diseases classification with few-shot meta-learning. The model framework consists of an embedding module and a distance module. The embedding module first projects supporting images into an embedding space using ResNet-18, then feeds embedding vectors into a transformer to incorporate information of other support samples in the same episode. After that, the distance module calculates the Mahalanobis distance\citep{galeano2015mahalanobis} of the query and support samples to classify the query. Similarly, \citep{li2021meta} uses a shallow CNN for embedding and the Euclidean distance for calculating the similarity between the embeddings of the query and support samples.

\subsubsection{Label Noise and Uncertainty Estimates}
\label{problem:noise}

\citep{shi2020rectified} is another example of meta-learning, but it is used to improve the network's robustness against label noise. The model consists of two phrases. The first phrase is the conventional training of a CNN for classification. In the second phrase, the authors generate ten synthetic mini batches of images, containing real images with the labels taken from similar images. As a result, these mini-batches could contain noisy labels. After one step update on the synthetic instances, the network is trained to output similar predictions with the CNN from the first phrase. The result is a model that is not easily affected by noisy training data. 

Finally, having a confidence score associated with the model prediction allows farmers to make decisions selectively under different confidence levels and boost the acceptance of deep learning models in agriculture. As an example,
\citep{frank2021confidence} performs classification of tomato diseases and pair the prediction with a confidence score following \citep{davis2019hierarchical}. The confidence score, calculated using Bayes' rule, is defined as the probability of the true class label conditioned on the class probability predicted by the CNN. In addition, the authors build an ontology of disease classification. For example, the parent node ``stressed plant'' has as children ``bacteria infection'' and ``virus infection'', which in turn has ``mosaic virus'' as a child. If the confidence score of a specific terminal disease label is below a certain threshold, the model switches to its more general parent label in the tree for higher confidence. By the axiom of probability, the predicted probability of the parent label is the summation of all the predicted probability of its direct descendants. For a general discussion of machine learning techniques that create well-calibrated uncertainty estimates, we refer readers to \S \ref{subsec:uncertainty}.

\section{Datasets}
\label{sec:data}

High-quality datasets with human annotations are one of the most important factors in the success of a machine learning project \cite{Whang2021:Data-Centric-AI,AndrewNg2021:Data-Centric,Miranda2021:Data-Centric}. In this section, we review established datasets that enable training of CV models. 
We exclude datasets for plants that we have not found literature regarding their suitability in CEA, such as apples\citep{bhusal2019apple, hani2020minneapple},  broccoli\citep{kusumam20173d}, and dates\citep{x46j-sk98-19}. 
We have manually checked every dataset listed and assure that they are available for downloading at the time of writing. 
By summarizing the dataset related to CEA, we aim to facilitate interested researchers on their future studies. In the meantime, we would like to encourage scholars to publish more datasets dedicated to CEA. 



As listed in Table \ref{tab:vertical1} and Table \ref{tab:vertical2}, we discover fourteen datasets in CEA, with three for Growth Monitoring, five for Fruit Detection, and six for Pest and Disease Detection. Each targeted task contains at least one dataset that covers multiple species to facilitate training of generalizable and transferable models. The largest dataset is CVPPP with 6,287 and 165,120 RGB images for Arabidopsis and Tobacco respectively, aiming for growth monitoring related tasks. 
All the available datasets are composed of real images. While real images provide realistic data, we also want to encourage publication of synthetic datasets, which usually feature balanced class distribution and accurate labeling.
Another point noteworthy is that many real images are collected under simplified laboratory environments, which may bias the data toward specific lighting conditions, backgrounds, plant orientation, or camera positions. For real world application, practitioners may need to further finetune the trained models on more realistic data. 



\begin{table*}[t]
    \centering
    \caption{Dataset for CV tasks in CEA}
    \label{tab:vertical1}
    \renewcommand{\arraystretch}{1.5}
    \begin{tabular}{@{}p{1.5cm}p{1.5cm}p{1cm}p{7.5cm}p{4cm}@{}}
    \toprule
     \textbf{Target Task} & \textbf{Dataset} &\textbf{Release Year} & \textbf{Data Description} & \textbf{URL} \\
    \hline
    \toprule
    \multirow{2}{*}[-5em]{\shortstack[l]{\textbf{Growth}\\\textbf{Monitoring}}} & CVPPP dataset  \citep{minervini2016finely} & 2014 &  6,287 and 165,120 RGB images (resolution 72x72) of Arabidopsis and Tobacco respectively . Annotations include bounding boxes and segmentation masks for every plant and every leaf, and the leaf centers.  & \url{https://www.plant-phenotyping.org/datasets-download}\\ 
    & Oil Radish dataset \citep{Mortensen_2019_CVPR_Workshops} & 2019 & 129 RGB images (resolution 1x1) of oil radish with binary semantic segmentation mask and respective plant fresh and dry weight, as well as nutrient content. & \url{https://competitions.codalab.org/competitions/20981#learn_the_details}\\ 
    & Leaf Counting dataset \citep{s18051580} & 2018 &  9,372 RGB images (resolution 72x72) of weeds with the number of leaves counted. & \url{https://vision.eng.au.dk/leaf-counting-dataset/}\\ 
    \midrule
    \multirow{4}{*}[-7em]{\shortstack[l]{\textbf{Fruit and} \\\textbf{Flower}\\\textbf{Detection}}} & DeepFruits \citep{s16081222} & 2016 & RGB images (resolution 72x72 to 400x400) of sweet pepper, rock melon, apple, mango, orange and strawberry images annotated with rectangular bounding boxes. Each fruit has 42-170 images. & \url{https://drive.google.com/drive/folders/1CmsZb1caggLRN7ANfika8WuPiywo4mBb}\\
    & Orchard Fruit\citep{bargoti2017deep} & 2016 & 1,120, 1,964 and 620 RGB images (resolution 72x72) of apple, mango and almond, respectively. Apples are annotated with bounding circles; mango and almond are annotated with rectangular bounding boxes & \url{http://data.acfr.usyd.edu.au/ag/treecrops/2016-multifruit/}\\
    & MangoYOLO \citep{koirala2019deep} & 2019 & 1,730 RGB images of mango (resolution 72x72 and 300x300), annotated with rectangular bounding boxes; photos are under artificial lighting & \url{https://figshare.com/articles/dataset/MangoYOLO_data_set/13450661}\\
    & MangoNet Semantic Dataset\citep{kestur2019mangonet} & 2019 & 45 training images and 4 test images (resolution 180x180) of mango. Each image is annotated with semantic segmentation mask which is colored green in regions of mangoes and black in non mango regions. & \url{https://github.com/avadesh02/MangoNet-Semantic-Dataset}\\
    & Fruit Flower detection \citep{8392727} & 2018 & 162, 20, and 15 images (resolution 72x72) of apple, peach, and pear flowers annotated with binary semantic segmentation mask with white represents flower pixels. & \url{https://data.nal.usda.gov/dataset/data-multi-species-fruit-flower-detection-using-refined-semantic-segmentation-network}\\
    \bottomrule
\end{tabular}
\end{table*}

\begin{table*}[t]
    \centering
    \caption{Dataset for CV tasks in CEA}
    \label{tab:vertical2}
    \renewcommand{\arraystretch}{1.5}
    \begin{tabular}{@{}p{2cm}p{2.5cm}p{1cm}p{6.5cm}p{4cm}@{}}
    \toprule
     \textbf{Target Task} & \textbf{Dataset} &\textbf{Release Year} & \textbf{Data Description} & \textbf{URL} \\
    \hline
    \toprule
    \multirow[c]{4}{*}[-3em]{\textbf{\shortstack[l]{Pest and \\Disease\\Detection}}} & Plant Village \citep{hughes2015open} & 2019 & 61,486 RGB images (resolution 72x72) of plant leaves, with 39 different classes of diseased and healthy plant leaves & \url{https://data.mendeley.com/datasets/tywbtsjrjv/1}\\ 

    & Crop Pests Recognition \citep{li2020crop} & 2020 & 5,629 RGB images (resolution 72x72) of 10 pest classes, each class containing over 400 images & \url{https://bit.ly/2DdUFza}\\ 
    & Plant and Pest \citep{li2021meta} & 2021 & 6,000 RGB images (resolution 72x72) of 20 different classes of plant leaves and pests from Plant Village \citep{hughes2015open} and Crop Pests Recognition \citep{li2020crop} & \url{https://zenodo.org/record/4529076#.YupE_-xBzlw}\\ 
    &  Citrus Pest Benchmark \citep{bollis2020weakly} & 2022 & 10,816 multi-class RGB images (resolution 1200x1200) categorized into seven classes of pests  & \url{https://github.com/edsonbollis/Citrus-Pest-Benchmark}\\ 
    & IP102 \citep{wu2019ip102} & 2019 & 75,000 images (resolution 400x300) of 102 insect classes and among these 19,000 are annotated with bounding boxes.  & \url{https://github.com/xpwu95/IP102}\\ 
    & Plant Leaves \citep{siddharth2019database} & 2022 & 4503 images (resolution 6000x4000) of which contains 2278 images of healthy leaves and 2225 images of the diseased leaves  & \url{https://data.mendeley.com/datasets/hb74ynkjcn/1}\\ 
    \bottomrule
\end{tabular}

\end{table*}



\section{Future Research Directions}
\label{sec:future}

So far we have discussed the objectives, benefits, and realizations of Growth Monitoring, Fruit and Flower Detection, Fruit Counting, Maturity Level Classification, and Pest and Disease Detection in CEA precision farming. Based on the current research status and existing technical capabilities of computer vision, we would like to point out several areas where computer vision technologies could provide short- to mid-term benefits to urban and suburban CEA. We identify three such areas, including realistic datasets that are unbalanced and noisy, uncertainty quantification, and multi-task learning / system integration.

\subsection{Handling Realistic Data}
\label{subsec:data problem}
The ability to handle realistic data is a critical competence that has not received sufficient research attention (with a few notable exceptions \citep{bollis2020weakly,gozzovelli2021tip,nuthalapati2021multi,li2021meta,shi2020rectified}). Unlike well-curated datasets that have accurate and abundant labels and relatively balanced label distributions, real-world data exhibit skewed label distribution as well as substantial noise in the labels. For effective real-world application, it is important that the CV algorithms can maintain good predictive performance under these conditions. In addition, the algorithmic tolerance of data imperfection can lower annotation cost and enable wider applications of CV.
There has been substantial research on these topics in the computer vision community, such as long-tail recognition \citep{shen2016relay,deng2019arcface,wang2018cosface,zhong2019unequal,liu2020deep,boyan2020bbn}, few-shot and zero-shot learning \citep{xian2016latent,li2019rethinking,snell2017prototypical,song2018selective,song2018transductive}, as well as noise-resistant classification \citep{cheng2022instance,algan2020meta,zheng2021mlc,jiang2020beyond,wei2020combating} and metric learning \citep{ibrahimi2022learning,wang2017robust,ChangLiu-BoyangLi-PRISM-2021}. We believe that research on smart agriculture could benefit from the existing body of literature.

\subsection{Quantifying Uncertainty and Interpretability}
\label{subsec:uncertainty and interpretability}
Real-world applications call for reliable estimation of the quality of automated decisions. An incorrect prediction made by an AI system may have profound implications. For example, if the system incorrectly determines that fruits are not mature enough, it may delay harvesting and cause overripe fruits with diminished values. However, it is impossible to eliminate incorrect or uncertain predictions, as they originate from factors difficult to control and precisely measure, including model assumptions, test data shift, incomplete training data and 
so on \citep{abdar2021review,hullermeier2021aleatoric}. Thus, we argue that uncertainty quantification is another crucial factor for real-world deployment. Such quantification would allow farmers to make informed decisions on whether to follow the machine recommendation or not. For the convenience of readers, we provide a brief review of such deep learning techniques in \S \ref{subsec:uncertainty}. 

Besides uncertainty quantification, pair the model with explanation on its decisions could enhance user confidence and assist auditing and debugging of the AI system. Specifically, instance attribution methods, as discussed in \S \ref{subsec:interpret}, enable detection of the biased or low quality data points with extreme influence on prediction \citep{chen2021hydra}. For example, if the model is trained with an image of dry leaves with dust that resemble a certain disease of the plant, in the inference process, the model might misclassify diseased leaves as normal dry leaves or vice versa and induce plant death or unnecessary treatments. With instance attribution interpretation, researchers can identify misleading data points and perform adversarial training to improve model accuracy.

\subsection{Multi-task Learning and System Integration}
Real-world deployment usually requires the coordination of multiple CV capabilities provided by different networks. When the system is designed well, these networks could facilitate each other and achieve synergistic effects. For example, instance segmentation can be used for fruit and flower localization (\S \ref{subsec:detect}), growth monitoring (\S \ref{sec:growth-monitoring}), and fruit maturity level detection (\S \ref{maturity}). However, academic research tends to study these problems in isolation, thereby unable to reap benefits of multi-task learning. 

Multi-task learning  \cite{caruana1997multitask,LiuQiuhua2007,bakker2003task} focuses on leveraging mutually beneficial supervisory signals from multiple correlated tasks. Recently, CV researchers have built large-scale networks \cite{lu2022unified,gupta2022towards,cho2021unifying,kamath2022webly,wang2022unifying,chen2021pix2seq,jaegle2021perceiver,zhu2022uni} that perform a wide range of tasks and achieve state-of-the-art results on most tasks. This demonstrates the benefits of multi-task learning and could inspire similar work dedicated to smart farming in CEAs. 

Another motivation for considering multi-task learning and system integration is that errors can propagate in a pipeline architecture. For example, a network could first incorrectly detect a leaf occluding a mature fruit as the fruit and then classify it as an immature fruit. As a result, simply concatenating multiple techniques will result in inferior overall performance than what practitioners may expect. Thus, we encourage system designers to consider end-to-end training, or other innovative techniques \cite{XuGuo-BoyangLi-NAACL-2021,yang2022empirical,wu2019logan}
for aligning and interfacing different components within a system.

Finally, multi-task learning handles multiple tasks simultaneously, which saves computation power, enhances data efficiency, and alleviates the necessity to maintain and iterate multiple models. Such benefits are crucial for popularizing CEAs, as they facilitate the efficient use of energy, computation power, and human resources.  Consequently, both the initial setup and ongoing maintenance investments for CEA farms can be reduced, expediting the emergence of economically viable CEAs. Furthermore,  mindful selection and combination of targeted tasks have the potential to further improve overall efficiency \citep{standley2020tasks}.

\subsection{Effective Use of Multimodality}
\label{multimodal}

Fusion of multi-modal data enhances inference ability of models by incorporating complementary view of data \cite{lahat2015multimodal}.
In the context of CEA, thermal or depth images capture the depth or temperature differences between foreground and background and enable filtering of non-target objects (e.g., fruits or leaves). Abnormal temperature changes during growth cycle can also indicate disease infection before visual symptoms appear \citep{chaerle2007monitoring, chaerle2004thermal}. Furthermore, as different materials absorb, reflect, and transmit light in different ways and at different wavelengths, multi-spectral imaging (MSI) and hyper-spectral imaging (HSI), which capturing images at multiple wavelengths of light, can be used to perform more specific internal inspection of leaves, fruits and plants as compared to thermal and depth images. Finally, LiDAR and RGB-D systems allow the generation of high density 3D point clouds of plants, fruits \citep{8863343,lin2019guava} or environment \citep{vulpi2022rgb}, which facilitate 3D volume measurement or cut-point detection during harvesting. 




Existing works have demonstrated the efficacy of multi-spectral imaging (MSI) and hyper-spectral imaging (HSI) \citep{afonso2020tomato,BLOK2021213,wang2019early}. MSI have been utilized for yield prediction \citep{torres2022optimizing} and early disease detection \citep{peng2022early, veys2019multispectral}. However, current literature explored majorly the power of MSI with shallow machine learning. We found only one work that leverages deep learning on MSI input \citep{torres2022optimizing}, which applies a pruned VGG-16 for wheat yield estimation. 
%
%
HSI provides finer-grained resolution and divides the range of wavelength into many more spectral bands than MSI, typically ranging from tens to hundreds of bands, though at a higher cost. Hyper-spectral images have been used as the sole modality in early disease detection with both shallow machine learning methods \citep{alsuwaidi2018feature, alsuwaidi2018combining,susivc2018discrimination} and deep learning methods \citep{wang2019early, s21030742,forster2019hyperspectral,gutierrez2019ground}. Due to relevancy and space limit, we will only talk about the deep learning methods here. 
Specifically, with a GAN-based data augmentation method, \citep{wang2019early} performs early detection of tomato spotted wilt virus before visible symptoms appear using hyper-spectral images.
\citep{s21030742} performs early detection of grapevine vein-clearing virus and shows the discriminative power of HSI in combination with CNN and shallow machine learning algorithms. \citep{forster2019hyperspectral} attains early barley disease detection through generating future prediction of hyper-spectral barley leaf images using GAN.
Moreover, HSI has also been utilized for yield prediction through fruit counting. \citep{gutierrez2019ground} leverages CNN and HSI to segment semantic mango masks and count the number of fruits.

However, systematic exploration of fusion techniques for multimodal inputs remains relatively rare in CEA applications. Many existing approaches adopt pipeline-based multimodal integration techniques that do not exhaust the potential of deep learning due to the lack of end-to-end training. For example, in \citep{afonso2020tomato}, the authors set a depth threshold to filter false positive tomato detection from the background.
\citep{BLOK2021213} first performs broccoli segmentation on the RGB image. Within the segmentation mask, the authors find the mode of the depth value distribution, which is used to calculate the diameter of the broccoli head. \citep{lin2019guava} conducts semantic segmentation for guava fruits using RGB images and reconstructs their 3D positions from the depth input. \citep{8863343} utilizes Mask R-CNN to perform instance segmentation of strawberries and align depth image with the segmentation mask to obtain 3D shape of strawberries.
These methods use the two modalities separately and do not apply end-to-end training of the pipeline. As exceptions, \citep{sa2016deepfruits} proposes late fusion of RGB and near-infrared images in sweet pepper detection. 
\citep{vit2018comparing} incorporates depth information by replacing the blue channel with depth channel and applies masked R-CNN to locate tomatoes.



%


In computer vision research, numerous techniques for fusing and joint utilization of multimodal information have been proposed over the years, which we believe could contribute to CV applications in CEA. Due to space limits, we list only a few examples here. \cite{sharma2020yolors} proposes two different ways to combine multiple modalities in object detection, Concatenation and Element-wise Cross Product. The former combines feature maps from different modalities along the channel dimension and let the network discover the best way to combine them from data. The latter technique, Element-wise Cross Product, applies element-wise multiplication to every possible pair of feature maps from the two modalities. \citep{carreira2017quo} experiments with a variety of fusion techniques for RGB and optical flow and discovers a high-performing late-fusion strategy in action recognition. In self-supervised learning, 
\citep{han2020self} identifies similar data points using one modality and treats them as positive pairs in another modality. This technique provides another paradigm to leverage the complementary nature of multimodality.

\section{Conclusions}
\label{sec:end}
Smart agriculture, and particularly computer vision for controlled-environment agriculture (CV4CEA), are rapidly emerging as an interdisciplinary area of research that could potentially lead to enormous economic, environmental and social benefits. In this survey, we first provide brief overviews of existing CV technologies that range from image recognition to structured understanding such as segmentation; from uncertain quantification to interpretable machine learning. Next, we systematically review existing applications of CV4CEA, including growth monitoring, fruit and flower detection, fruit counting, maturity level classification, and pest / disease detection. Finally, we highlight a few research directions that could generate high-impact research in the near future.

Like any interdisciplinary area, research progress in CV4CEA requires expertise in both computer vision and agriculture. However, it could take a substantial amount of time for any researcher to acquire in-depth understanding of both subjects. 
By reviewing existing applications, available CV technologies, and identifying possible future research directions, we aim to provide a quick introduction of CV4CEA to researchers with expertise in agriculture or computer vision alone. It is our hope that the current survey will serve as a bridge between researchers from diverse backgrounds and contribute to accelerated innovation in the next decade. 


\section{Acknowledgments}

The authors gratefully acknowledge the support from the WeBank-NTU Joint Research Center (Grant number NWJ-2020-008) and from the China-Singapore International Joint Research Institute (Grant number 206-A021002). 


\begin{spacing}{1.08}
\bibliographystyle{ACM-Reference-Format}
\bibliography{main-July-2022}


\begin{thebibliography}{383}


\ifx \showCODEN    \undefined \def \showCODEN     #1{\unskip}     \fi
\ifx \showDOI      \undefined \def \showDOI       #1{#1}\fi
\ifx \showISBNx    \undefined \def \showISBNx     #1{\unskip}     \fi
\ifx \showISBNxiii \undefined \def \showISBNxiii  #1{\unskip}     \fi
\ifx \showISSN     \undefined \def \showISSN      #1{\unskip}     \fi
\ifx \showLCCN     \undefined \def \showLCCN      #1{\unskip}     \fi
\ifx \shownote     \undefined \def \shownote      #1{#1}          \fi
\ifx \showarticletitle \undefined \def \showarticletitle #1{#1}   \fi
\ifx \showURL      \undefined \def \showURL       {\relax}        \fi
\providecommand\bibfield[2]{#2}
\providecommand\bibinfo[2]{#2}
\providecommand\natexlab[1]{#1}
\providecommand\showeprint[2][]{arXiv:#2}

\bibitem[\protect\citeauthoryear{??}{30b}{[n.\,d.]}]%
        {30by30gov}
 \bibinfo{year}{[n.\,d.]}\natexlab{}.
\newblock \bibinfo{title}{30 by 30: Strengthening our food security}.
\newblock \bibinfo{howpublished}{\url{https://www.ourfoodfuture.gov.sg/30by30}}.
\newblock
\newblock
\shownote{Accessed: 2022-8-15}.


\bibitem[\protect\citeauthoryear{??}{The}{[n.\,d.]}]%
        {Thespoon-blurberry}
 \bibinfo{year}{[n.\,d.]}\natexlab{}.
\newblock \bibinfo{title}{AeroFarms Partners With Hortifrut to Grow Blueberries, Caneberries Via Vertical Farming}.
\newblock \bibinfo{howpublished}{\url{https://thespoon.tech/aerofarms-partners-with-hortifrut-to-grow-blueberries-caneberries-via-vertical-farming/}}.
\newblock
\newblock
\shownote{Accessed: 2022-7-28}.


\bibitem[\protect\citeauthoryear{??}{Alg}{[n.\,d.]}]%
        {AlgorithmicBotany}
 \bibinfo{year}{[n.\,d.]}\natexlab{}.
\newblock \bibinfo{title}{Algorithmic Botany}.
\newblock \bibinfo{howpublished}{\url{http://www.algorithmicbotany.org/virtual_laboratory/}}.
\newblock
\newblock
\shownote{Accessed: 2022-6-20}.


\bibitem[\protect\citeauthoryear{??}{Hor}{[n.\,d.]}]%
        {Hortibizdaily-citrus}
 \bibinfo{year}{[n.\,d.]}\natexlab{}.
\newblock \bibinfo{title}{All in(doors) on citrus production}.
\newblock \bibinfo{howpublished}{\url{https://www.hortibiz.com/newsitem/news/all-indoors-on-citrus-production/}}.
\newblock
\newblock
\shownote{Accessed: 2022-7-28}.


\bibitem[\protect\citeauthoryear{??}{hor}{[n.\,d.]}]%
        {hortidaily-banana}
 \bibinfo{year}{[n.\,d.]}\natexlab{}.
\newblock \bibinfo{title}{Greenhouse in Shanghai successfully plants bananas on water}.
\newblock \bibinfo{howpublished}{\url{https://www.hortidaily.com/article/9369964/greenhouse-in-shanghai-successfully-plants-bananas-on-water/ }}.
\newblock
\newblock
\shownote{Accessed: 2022-7-28}.


\bibitem[\protect\citeauthoryear{??}{Ver}{[n.\,d.]}]%
        {Verticrop}
 \bibinfo{year}{[n.\,d.]}\natexlab{}.
\newblock \bibinfo{title}{Introducing VertiCrop™}.
\newblock \bibinfo{howpublished}{\url{https://verticrop.com/}}.
\newblock
\newblock
\shownote{Accessed: 2022-5-24}.


\bibitem[\protect\citeauthoryear{??}{Hor}{[n.\,d.]}]%
        {Horti-mango}
 \bibinfo{year}{[n.\,d.]}\natexlab{}.
\newblock \bibinfo{title}{Mango trees cultivation under greenhouse conditions}.
\newblock \bibinfo{howpublished}{\url{https://horti-generation.com/mango-trees-cultivation-under-greenhouse-conditions/ }}.
\newblock
\newblock
\shownote{Accessed: 2022-7-28}.


\bibitem[\protect\citeauthoryear{??}{sat}{[n.\,d.]}]%
        {saturnbioponics}
 \bibinfo{year}{[n.\,d.]}\natexlab{}.
\newblock \bibinfo{title}{Saturn Bioponics}.
\newblock \bibinfo{howpublished}{\url{http://www.saturnbioponics.com/}}.
\newblock
\newblock
\shownote{Accessed: 2022-05-25}.


\bibitem[\protect\citeauthoryear{??}{Spr}{[n.\,d.]}]%
        {SpreadVertifarm}
 \bibinfo{year}{[n.\,d.]}\natexlab{}.
\newblock \bibinfo{title}{Spread-A new way to grow vegetable}.
\newblock \bibinfo{howpublished}{\url{https://spread.co.jp/en/environment/}}.
\newblock
\newblock
\shownote{Accessed: 2022-05-24}.


\bibitem[\protect\citeauthoryear{??}{hor}{[n.\,d.]}]%
        {hortidaily-cucumber}
 \bibinfo{year}{[n.\,d.]}\natexlab{}.
\newblock \bibinfo{title}{Tomatoes and cucumbers in a vertical farm without daylight}.
\newblock \bibinfo{howpublished}{\url{https://www.hortidaily.com/article/9212847/tomatoes-and-cucumbers-in-a-vertical-farm-without-daylight/}}.
\newblock
\newblock
\shownote{Accessed: 2022-7-28}.


\bibitem[\protect\citeauthoryear{Abdar, Pourpanah, Hussain, Rezazadegan, Liu, Ghavamzadeh, Fieguth, Cao, Khosravi, Acharya, et~al\mbox{.}}{Abdar et~al\mbox{.}}{2021}]%
        {abdar2021review}
\bibfield{author}{\bibinfo{person}{Moloud Abdar}, \bibinfo{person}{Farhad Pourpanah}, \bibinfo{person}{Sadiq Hussain}, \bibinfo{person}{Dana Rezazadegan}, \bibinfo{person}{Li Liu}, \bibinfo{person}{Mohammad Ghavamzadeh}, \bibinfo{person}{Paul Fieguth}, \bibinfo{person}{Xiaochun Cao}, \bibinfo{person}{Abbas Khosravi}, \bibinfo{person}{U~Rajendra Acharya}, {et~al\mbox{.}}} \bibinfo{year}{2021}\natexlab{}.
\newblock \showarticletitle{A review of uncertainty quantification in deep learning: Techniques, applications and challenges}.
\newblock \bibinfo{journal}{\emph{Information Fusion}}  \bibinfo{volume}{76} (\bibinfo{year}{2021}), \bibinfo{pages}{243--297}.
\newblock


\bibitem[\protect\citeauthoryear{Adelson}{Adelson}{2001}]%
        {Adelson2001:Seeing-Stuff}
\bibfield{author}{\bibinfo{person}{Edward~H. Adelson}.} \bibinfo{year}{2001}\natexlab{}.
\newblock \showarticletitle{{On seeing stuff: the perception of materials by humans and machines}}. In \bibinfo{booktitle}{\emph{Human Vision and Electronic Imaging VI}}, \bibfield{editor}{\bibinfo{person}{Bernice~E. Rogowitz} {and} \bibinfo{person}{Thrasyvoulos~N. Pappas}} (Eds.), Vol.~\bibinfo{volume}{4299}. International Society for Optics and Photonics, \bibinfo{publisher}{SPIE}, \bibinfo{pages}{1 -- 12}.
\newblock
\urldef\tempurl%
\url{https://doi.org/10.1117/12.429489}
\showDOI{\tempurl}


\bibitem[\protect\citeauthoryear{Afonso, Fonteijn, Fiorentin, Lensink, Mooij, Faber, Polder, and Wehrens}{Afonso et~al\mbox{.}}{2020}]%
        {afonso2020tomato}
\bibfield{author}{\bibinfo{person}{Manya Afonso}, \bibinfo{person}{Hubert Fonteijn}, \bibinfo{person}{Felipe~Schadeck Fiorentin}, \bibinfo{person}{Dick Lensink}, \bibinfo{person}{Marcel Mooij}, \bibinfo{person}{Nanne Faber}, \bibinfo{person}{Gerrit Polder}, {and} \bibinfo{person}{Ron Wehrens}.} \bibinfo{year}{2020}\natexlab{}.
\newblock \showarticletitle{Tomato fruit detection and counting in greenhouses using deep learning}.
\newblock \bibinfo{journal}{\emph{Frontiers in plant science}}  \bibinfo{volume}{11} (\bibinfo{year}{2020}), \bibinfo{pages}{571299}.
\newblock


\bibitem[\protect\citeauthoryear{Ahern, Noack, Guzman-Nateras, Dou, Li, and Huan}{Ahern et~al\mbox{.}}{[n.\,d.]}]%
        {Ahern2019}
\bibfield{author}{\bibinfo{person}{Isaac Ahern}, \bibinfo{person}{Adam Noack}, \bibinfo{person}{Luis Guzman-Nateras}, \bibinfo{person}{Dejing Dou}, \bibinfo{person}{Boyang Li}, {and} \bibinfo{person}{Jun Huan}.} \bibinfo{year}{[n.\,d.]}\natexlab{}.
\newblock \showarticletitle{NormLime: A New Feature Importance Metric for Explaining Deep Neural Networks}.
\newblock \bibinfo{journal}{\emph{arXiv Preprint 1909.04200}} (\bibinfo{year}{[n.\,d.]}).
\newblock
\urldef\tempurl%
\url{https://doi.org/10.48550/ARXIV.1909.04200}
\showDOI{\tempurl}


\bibitem[\protect\citeauthoryear{Ahmad and Nabi}{Ahmad and Nabi}{2021}]%
        {ahmad2021agriculture}
\bibfield{author}{\bibinfo{person}{Latief Ahmad} {and} \bibinfo{person}{Firasath Nabi}.} \bibinfo{year}{2021}\natexlab{}.
\newblock \bibinfo{booktitle}{\emph{Agriculture 5.0: Artificial Intelligence, IoT and Machine Learning}}.
\newblock \bibinfo{publisher}{CRC Press}.
\newblock


\bibitem[\protect\citeauthoryear{Ahn, Cho, and Kwak}{Ahn et~al\mbox{.}}{2019}]%
        {ahn2019weakly}
\bibfield{author}{\bibinfo{person}{Jiwoon Ahn}, \bibinfo{person}{Sunghyun Cho}, {and} \bibinfo{person}{Suha Kwak}.} \bibinfo{year}{2019}\natexlab{}.
\newblock \showarticletitle{Weakly supervised learning of instance segmentation with inter-pixel relations}. In \bibinfo{booktitle}{\emph{Proceedings of the IEEE/CVF conference on computer vision and pattern recognition}}. \bibinfo{pages}{2209--2218}.
\newblock


\bibitem[\protect\citeauthoryear{Algan and Ulusoy}{Algan and Ulusoy}{2020}]%
        {algan2020meta}
\bibfield{author}{\bibinfo{person}{G{\"o}rkem Algan} {and} \bibinfo{person}{Ilkay Ulusoy}.} \bibinfo{year}{2020}\natexlab{}.
\newblock \showarticletitle{Meta soft label generation for noisy labels}. In \bibinfo{booktitle}{\emph{ICPR}}.
\newblock


\bibitem[\protect\citeauthoryear{AlSuwaidi, Grieve, and Yin}{AlSuwaidi et~al\mbox{.}}{2018a}]%
        {alsuwaidi2018combining}
\bibfield{author}{\bibinfo{person}{Ali AlSuwaidi}, \bibinfo{person}{Bruce Grieve}, {and} \bibinfo{person}{Hujun Yin}.} \bibinfo{year}{2018}\natexlab{a}.
\newblock \showarticletitle{Combining spectral and texture features in hyperspectral image analysis for plant monitoring}.
\newblock \bibinfo{journal}{\emph{Measurement Science and Technology}} \bibinfo{volume}{29}, \bibinfo{number}{10} (\bibinfo{year}{2018}), \bibinfo{pages}{104001}.
\newblock


\bibitem[\protect\citeauthoryear{AlSuwaidi, Grieve, and Yin}{AlSuwaidi et~al\mbox{.}}{2018b}]%
        {alsuwaidi2018feature}
\bibfield{author}{\bibinfo{person}{Ali AlSuwaidi}, \bibinfo{person}{Bruce Grieve}, {and} \bibinfo{person}{Hujun Yin}.} \bibinfo{year}{2018}\natexlab{b}.
\newblock \showarticletitle{Feature-ensemble-based novelty detection for analyzing plant hyperspectral datasets}.
\newblock \bibinfo{journal}{\emph{IEEE Journal of Selected Topics in Applied Earth Observations and Remote Sensing}} \bibinfo{volume}{11}, \bibinfo{number}{4} (\bibinfo{year}{2018}), \bibinfo{pages}{1041--1055}.
\newblock


\bibitem[\protect\citeauthoryear{Altaheri, Alsulaiman, Faisal, and Muhammed}{Altaheri et~al\mbox{.}}{2019a}]%
        {altaheri2019date}
\bibfield{author}{\bibinfo{person}{H Altaheri}, \bibinfo{person}{M Alsulaiman}, \bibinfo{person}{M Faisal}, {and} \bibinfo{person}{G Muhammed}.} \bibinfo{year}{2019}\natexlab{a}.
\newblock \showarticletitle{Date fruit dataset for automated harvesting and visual yield estimation}. In \bibinfo{booktitle}{\emph{Proc. IEEE DataPort}}.
\newblock


\bibitem[\protect\citeauthoryear{Altaheri, Alsulaiman, Faisal, and Muhammed}{Altaheri et~al\mbox{.}}{2019b}]%
        {x46j-sk98-19}
\bibfield{author}{\bibinfo{person}{Hamdi Altaheri}, \bibinfo{person}{Mansour Alsulaiman}, \bibinfo{person}{Mohammed Faisal}, {and} \bibinfo{person}{Ghulam Muhammed}.} \bibinfo{year}{2019}\natexlab{b}.
\newblock \bibinfo{title}{Date Fruit Dataset for Automated Harvesting and Visual Yield Estimation}.
\newblock
\newblock
\urldef\tempurl%
\url{https://doi.org/10.21227/x46j-sk98}
\showDOI{\tempurl}


\bibitem[\protect\citeauthoryear{Ancona, Ceolini, {\"O}ztireli, and Gross}{Ancona et~al\mbox{.}}{2017}]%
        {ancona2017towards}
\bibfield{author}{\bibinfo{person}{Marco Ancona}, \bibinfo{person}{Enea Ceolini}, \bibinfo{person}{Cengiz {\"O}ztireli}, {and} \bibinfo{person}{Markus Gross}.} \bibinfo{year}{2017}\natexlab{}.
\newblock \showarticletitle{Towards better understanding of gradient-based attribution methods for deep neural networks}.
\newblock \bibinfo{journal}{\emph{arXiv preprint arXiv:1711.06104}} (\bibinfo{year}{2017}).
\newblock


\bibitem[\protect\citeauthoryear{Anderson, He, Buehler, Teney, Johnson, Gould, and Zhang}{Anderson et~al\mbox{.}}{2018}]%
        {Anderson_2018_CVPR}
\bibfield{author}{\bibinfo{person}{Peter Anderson}, \bibinfo{person}{Xiaodong He}, \bibinfo{person}{Chris Buehler}, \bibinfo{person}{Damien Teney}, \bibinfo{person}{Mark Johnson}, \bibinfo{person}{Stephen Gould}, {and} \bibinfo{person}{Lei Zhang}.} \bibinfo{year}{2018}\natexlab{}.
\newblock \showarticletitle{Bottom-Up and Top-Down Attention for Image Captioning and Visual Question Answering}. In \bibinfo{booktitle}{\emph{Proceedings of the IEEE Conference on Computer Vision and Pattern Recognition (CVPR)}}.
\newblock


\bibitem[\protect\citeauthoryear{Aravind, Raja, Ashiwin, and Mukesh}{Aravind et~al\mbox{.}}{2019}]%
        {aravind2019disease}
\bibfield{author}{\bibinfo{person}{Krishnaswamy~R Aravind}, \bibinfo{person}{Purushothaman Raja}, \bibinfo{person}{Rajendran Ashiwin}, {and} \bibinfo{person}{Konnaiyar~V Mukesh}.} \bibinfo{year}{2019}\natexlab{}.
\newblock \showarticletitle{Disease classification in Solanum melongena using deep learning}.
\newblock \bibinfo{journal}{\emph{Spanish Journal of Agricultural Research}} \bibinfo{volume}{17}, \bibinfo{number}{3} (\bibinfo{year}{2019}), \bibinfo{pages}{e0204--e0204}.
\newblock


\bibitem[\protect\citeauthoryear{Arefi, Motlagh, Mollazade, and Teimourlou}{Arefi et~al\mbox{.}}{2011}]%
        {arefi2011recognition}
\bibfield{author}{\bibinfo{person}{Arman Arefi}, \bibinfo{person}{Asad~Modarres Motlagh}, \bibinfo{person}{Kaveh Mollazade}, {and} \bibinfo{person}{Rahman~Farrokhi Teimourlou}.} \bibinfo{year}{2011}\natexlab{}.
\newblock \showarticletitle{Recognition and localization of ripen tomato based on machine vision}.
\newblock \bibinfo{journal}{\emph{Australian Journal of Crop Science}} \bibinfo{volume}{5}, \bibinfo{number}{10} (\bibinfo{year}{2011}), \bibinfo{pages}{1144--1149}.
\newblock


\bibitem[\protect\citeauthoryear{Arjovsky, Chintala, and Bottou}{Arjovsky et~al\mbox{.}}{2017}]%
        {arjovsky2017wasserstein}
\bibfield{author}{\bibinfo{person}{Martin Arjovsky}, \bibinfo{person}{Soumith Chintala}, {and} \bibinfo{person}{L{\'e}on Bottou}.} \bibinfo{year}{2017}\natexlab{}.
\newblock \showarticletitle{Wasserstein generative adversarial networks}. In \bibinfo{booktitle}{\emph{International conference on machine learning}}. PMLR, \bibinfo{pages}{214--223}.
\newblock


\bibitem[\protect\citeauthoryear{Arora, Cohen, and Hazan}{Arora et~al\mbox{.}}{2018}]%
        {arora2018optimization}
\bibfield{author}{\bibinfo{person}{Sanjeev Arora}, \bibinfo{person}{Nadav Cohen}, {and} \bibinfo{person}{Elad Hazan}.} \bibinfo{year}{2018}\natexlab{}.
\newblock \showarticletitle{On the optimization of deep networks: Implicit acceleration by overparameterization}. In \bibinfo{booktitle}{\emph{International Conference on Machine Learning}}. PMLR, \bibinfo{pages}{244--253}.
\newblock


\bibitem[\protect\citeauthoryear{Aruul Mozhi~Varman, Baskaran, Aravindh, and Prabhu}{Aruul Mozhi~Varman et~al\mbox{.}}{2017}]%
        {8524140}
\bibfield{author}{\bibinfo{person}{S Aruul Mozhi~Varman}, \bibinfo{person}{Arvind~Ram Baskaran}, \bibinfo{person}{S Aravindh}, {and} \bibinfo{person}{E Prabhu}.} \bibinfo{year}{2017}\natexlab{}.
\newblock \showarticletitle{Deep Learning and IoT for Smart Agriculture Using WSN}. In \bibinfo{booktitle}{\emph{2017 IEEE International Conference on Computational Intelligence and Computing Research (ICCIC)}}. \bibinfo{pages}{1--6}.
\newblock
\urldef\tempurl%
\url{https://doi.org/10.1109/ICCIC.2017.8524140}
\showDOI{\tempurl}


\bibitem[\protect\citeauthoryear{Bakker and Heskes}{Bakker and Heskes}{2003}]%
        {bakker2003task}
\bibfield{author}{\bibinfo{person}{BJ Bakker} {and} \bibinfo{person}{TM Heskes}.} \bibinfo{year}{2003}\natexlab{}.
\newblock \showarticletitle{Task clustering and gating for bayesian multitask learning}.
\newblock  (\bibinfo{year}{2003}).
\newblock


\bibitem[\protect\citeauthoryear{Bargoti and Underwood}{Bargoti and Underwood}{2017a}]%
        {bargoti2017deep}
\bibfield{author}{\bibinfo{person}{Suchet Bargoti} {and} \bibinfo{person}{James Underwood}.} \bibinfo{year}{2017}\natexlab{a}.
\newblock \showarticletitle{Deep fruit detection in orchards}. In \bibinfo{booktitle}{\emph{2017 IEEE International Conference on Robotics and Automation (ICRA)}}. IEEE, \bibinfo{pages}{3626--3633}.
\newblock


\bibitem[\protect\citeauthoryear{Bargoti and Underwood}{Bargoti and Underwood}{2017b}]%
        {bargoti2017image}
\bibfield{author}{\bibinfo{person}{Suchet Bargoti} {and} \bibinfo{person}{James~P Underwood}.} \bibinfo{year}{2017}\natexlab{b}.
\newblock \showarticletitle{Image segmentation for fruit detection and yield estimation in apple orchards}.
\newblock \bibinfo{journal}{\emph{Journal of Field Robotics}} \bibinfo{volume}{34}, \bibinfo{number}{6} (\bibinfo{year}{2017}), \bibinfo{pages}{1039--1060}.
\newblock


\bibitem[\protect\citeauthoryear{Barshan, Brunet, and Dziugaite}{Barshan et~al\mbox{.}}{2020}]%
        {barshan2020relatif}
\bibfield{author}{\bibinfo{person}{Elnaz Barshan}, \bibinfo{person}{Marc-Etienne Brunet}, {and} \bibinfo{person}{Gintare~Karolina Dziugaite}.} \bibinfo{year}{2020}\natexlab{}.
\newblock \showarticletitle{Relatif: Identifying explanatory training samples via relative influence}. In \bibinfo{booktitle}{\emph{International Conference on Artificial Intelligence and Statistics}}. PMLR, \bibinfo{pages}{1899--1909}.
\newblock


\bibitem[\protect\citeauthoryear{Bastings, Aziz, and Titov}{Bastings et~al\mbox{.}}{2019}]%
        {bastings2019interpretable}
\bibfield{author}{\bibinfo{person}{Jasmijn Bastings}, \bibinfo{person}{Wilker Aziz}, {and} \bibinfo{person}{Ivan Titov}.} \bibinfo{year}{2019}\natexlab{}.
\newblock \showarticletitle{Interpretable neural predictions with differentiable binary variables}.
\newblock \bibinfo{journal}{\emph{arXiv preprint arXiv:1905.08160}} (\bibinfo{year}{2019}).
\newblock


\bibitem[\protect\citeauthoryear{Bau, Zhou, Khosla, Oliva, and Torralba}{Bau et~al\mbox{.}}{2017}]%
        {bau2017network}
\bibfield{author}{\bibinfo{person}{David Bau}, \bibinfo{person}{Bolei Zhou}, \bibinfo{person}{Aditya Khosla}, \bibinfo{person}{Aude Oliva}, {and} \bibinfo{person}{Antonio Torralba}.} \bibinfo{year}{2017}\natexlab{}.
\newblock \showarticletitle{Network dissection: Quantifying interpretability of deep visual representations}. In \bibinfo{booktitle}{\emph{Proceedings of the IEEE conference on computer vision and pattern recognition}}. \bibinfo{pages}{6541--6549}.
\newblock


\bibitem[\protect\citeauthoryear{Beacham, Vickers, and Monaghan}{Beacham et~al\mbox{.}}{2019}]%
        {beacham2019vertical}
\bibfield{author}{\bibinfo{person}{Andrew~M Beacham}, \bibinfo{person}{Laura~H Vickers}, {and} \bibinfo{person}{James~M Monaghan}.} \bibinfo{year}{2019}\natexlab{}.
\newblock \showarticletitle{Vertical farming: a summary of approaches to growing skywards}.
\newblock \bibinfo{journal}{\emph{The Journal of Horticultural Science and Biotechnology}} \bibinfo{volume}{94}, \bibinfo{number}{3} (\bibinfo{year}{2019}), \bibinfo{pages}{277--283}.
\newblock


\bibitem[\protect\citeauthoryear{Bechar and Vigneault}{Bechar and Vigneault}{2016}]%
        {bechar2016agricultural}
\bibfield{author}{\bibinfo{person}{Avital Bechar} {and} \bibinfo{person}{Cl{\'e}ment Vigneault}.} \bibinfo{year}{2016}\natexlab{}.
\newblock \showarticletitle{Agricultural robots for field operations: Concepts and components}.
\newblock \bibinfo{journal}{\emph{Biosystems Engineering}}  \bibinfo{volume}{149} (\bibinfo{year}{2016}), \bibinfo{pages}{94--111}.
\newblock


\bibitem[\protect\citeauthoryear{Benis, Reinhart, and Ferr{\~a}o}{Benis et~al\mbox{.}}{2017}]%
        {benis2017development}
\bibfield{author}{\bibinfo{person}{Khadija Benis}, \bibinfo{person}{Christoph Reinhart}, {and} \bibinfo{person}{Paulo Ferr{\~a}o}.} \bibinfo{year}{2017}\natexlab{}.
\newblock \showarticletitle{Development of a simulation-based decision support workflow for the implementation of Building-Integrated Agriculture (BIA) in urban contexts}.
\newblock \bibinfo{journal}{\emph{Journal of cleaner production}}  \bibinfo{volume}{147} (\bibinfo{year}{2017}), \bibinfo{pages}{589--602}.
\newblock


\bibitem[\protect\citeauthoryear{Benke and Tomkins}{Benke and Tomkins}{2017}]%
        {benke2017future}
\bibfield{author}{\bibinfo{person}{Kurt Benke} {and} \bibinfo{person}{Bruce Tomkins}.} \bibinfo{year}{2017}\natexlab{}.
\newblock \showarticletitle{Future food-production systems: vertical farming and controlled-environment agriculture}.
\newblock \bibinfo{journal}{\emph{Sustainability: Science, Practice and Policy}} \bibinfo{volume}{13}, \bibinfo{number}{1} (\bibinfo{year}{2017}), \bibinfo{pages}{13--26}.
\newblock


\bibitem[\protect\citeauthoryear{Berckmans}{Berckmans}{2017}]%
        {berckmans2017general}
\bibfield{author}{\bibinfo{person}{Daniel Berckmans}.} \bibinfo{year}{2017}\natexlab{}.
\newblock \showarticletitle{General introduction to precision livestock farming}.
\newblock \bibinfo{journal}{\emph{Animal Frontiers}} \bibinfo{volume}{7}, \bibinfo{number}{1} (\bibinfo{year}{2017}), \bibinfo{pages}{6--11}.
\newblock


\bibitem[\protect\citeauthoryear{Bhargava and Bansal}{Bhargava and Bansal}{2021}]%
        {bhargava2021fruits}
\bibfield{author}{\bibinfo{person}{Anuja Bhargava} {and} \bibinfo{person}{Atul Bansal}.} \bibinfo{year}{2021}\natexlab{}.
\newblock \showarticletitle{Fruits and vegetables quality evaluation using computer vision: A review}.
\newblock \bibinfo{journal}{\emph{Journal of King Saud University-Computer and Information Sciences}} \bibinfo{volume}{33}, \bibinfo{number}{3} (\bibinfo{year}{2021}), \bibinfo{pages}{243--257}.
\newblock


\bibitem[\protect\citeauthoryear{Bhusal, Karkee, and Zhang}{Bhusal et~al\mbox{.}}{2019}]%
        {bhusal2019apple}
\bibfield{author}{\bibinfo{person}{Santosh Bhusal}, \bibinfo{person}{Manoj Karkee}, {and} \bibinfo{person}{Qin Zhang}.} \bibinfo{year}{2019}\natexlab{}.
\newblock \showarticletitle{Apple Dataset Benchmark from Orchard Environment in Modern Fruiting Wall}.
\newblock  (\bibinfo{year}{2019}).
\newblock


\bibitem[\protect\citeauthoryear{Blok, {van Henten}, {van Evert}, and Kootstra}{Blok et~al\mbox{.}}{2021}]%
        {BLOK2021213}
\bibfield{author}{\bibinfo{person}{Pieter~M. Blok}, \bibinfo{person}{Eldert~J. {van Henten}}, \bibinfo{person}{Frits~K. {van Evert}}, {and} \bibinfo{person}{Gert Kootstra}.} \bibinfo{year}{2021}\natexlab{}.
\newblock \showarticletitle{Image-based size estimation of broccoli heads under varying degrees of occlusion}.
\newblock \bibinfo{journal}{\emph{Biosystems Engineering}}  \bibinfo{volume}{208} (\bibinfo{year}{2021}), \bibinfo{pages}{213--233}.
\newblock
\showISSN{1537-5110}
\urldef\tempurl%
\url{https://doi.org/10.1016/j.biosystemseng.2021.06.001}
\showDOI{\tempurl}


\bibitem[\protect\citeauthoryear{Bock, Parker, Cook, and Gottwald}{Bock et~al\mbox{.}}{2008}]%
        {bock2008visual}
\bibfield{author}{\bibinfo{person}{CH Bock}, \bibinfo{person}{PE Parker}, \bibinfo{person}{AZ Cook}, {and} \bibinfo{person}{TR Gottwald}.} \bibinfo{year}{2008}\natexlab{}.
\newblock \showarticletitle{Visual rating and the use of image analysis for assessing different symptoms of citrus canker on grapefruit leaves}.
\newblock \bibinfo{journal}{\emph{Plant Disease}} \bibinfo{volume}{92}, \bibinfo{number}{4} (\bibinfo{year}{2008}), \bibinfo{pages}{530--541}.
\newblock


\bibitem[\protect\citeauthoryear{Bollis, Pedrini, and Avila}{Bollis et~al\mbox{.}}{2020}]%
        {bollis2020weakly}
\bibfield{author}{\bibinfo{person}{Edson Bollis}, \bibinfo{person}{Helio Pedrini}, {and} \bibinfo{person}{Sandra Avila}.} \bibinfo{year}{2020}\natexlab{}.
\newblock \showarticletitle{Weakly supervised learning guided by activation mapping applied to a novel citrus pest benchmark}. In \bibinfo{booktitle}{\emph{Proceedings of the IEEE/CVF Conference on Computer Vision and Pattern Recognition Workshops}}. \bibinfo{pages}{70--71}.
\newblock


\bibitem[\protect\citeauthoryear{Bouman and Shapiro}{Bouman and Shapiro}{1994}]%
        {bouman1994multiscale}
\bibfield{author}{\bibinfo{person}{Charles~A Bouman} {and} \bibinfo{person}{Michael Shapiro}.} \bibinfo{year}{1994}\natexlab{}.
\newblock \showarticletitle{A multiscale random field model for Bayesian image segmentation}.
\newblock \bibinfo{journal}{\emph{IEEE Transactions on image processing}} \bibinfo{volume}{3}, \bibinfo{number}{2} (\bibinfo{year}{1994}), \bibinfo{pages}{162--177}.
\newblock


\bibitem[\protect\citeauthoryear{Brophy and Lowd}{Brophy and Lowd}{2020}]%
        {brophy2020trex}
\bibfield{author}{\bibinfo{person}{Jonathan Brophy} {and} \bibinfo{person}{Daniel Lowd}.} \bibinfo{year}{2020}\natexlab{}.
\newblock \showarticletitle{TREX: Tree-Ensemble Representer-Point Explanations}.
\newblock \bibinfo{journal}{\emph{arXiv preprint arXiv:2009.05530}} (\bibinfo{year}{2020}).
\newblock


\bibitem[\protect\citeauthoryear{Brown, Mann, Ryder, Subbiah, Kaplan, Dhariwal, Neelakantan, Shyam, Sastry, Askell, Agarwal, Herbert-Voss, Krueger, Henighan, Child, Ramesh, Ziegler, Wu, Winter, Hesse, Chen, Sigler, Litwin, Gray, Chess, Clark, Berner, McCandlish, Radford, Sutskever, and Amodei}{Brown et~al\mbox{.}}{2020}]%
        {brown2020language}
\bibfield{author}{\bibinfo{person}{Tom~B. Brown}, \bibinfo{person}{Benjamin Mann}, \bibinfo{person}{Nick Ryder}, \bibinfo{person}{Melanie Subbiah}, \bibinfo{person}{Jared Kaplan}, \bibinfo{person}{Prafulla Dhariwal}, \bibinfo{person}{Arvind Neelakantan}, \bibinfo{person}{Pranav Shyam}, \bibinfo{person}{Girish Sastry}, \bibinfo{person}{Amanda Askell}, \bibinfo{person}{Sandhini Agarwal}, \bibinfo{person}{Ariel Herbert-Voss}, \bibinfo{person}{Gretchen Krueger}, \bibinfo{person}{Tom Henighan}, \bibinfo{person}{Rewon Child}, \bibinfo{person}{Aditya Ramesh}, \bibinfo{person}{Daniel~M. Ziegler}, \bibinfo{person}{Jeffrey Wu}, \bibinfo{person}{Clemens Winter}, \bibinfo{person}{Christopher Hesse}, \bibinfo{person}{Mark Chen}, \bibinfo{person}{Eric Sigler}, \bibinfo{person}{Mateusz Litwin}, \bibinfo{person}{Scott Gray}, \bibinfo{person}{Benjamin Chess}, \bibinfo{person}{Jack Clark}, \bibinfo{person}{Christopher Berner}, \bibinfo{person}{Sam McCandlish}, \bibinfo{person}{Alec Radford}, \bibinfo{person}{Ilya Sutskever},
  {and} \bibinfo{person}{Dario Amodei}.} \bibinfo{year}{2020}\natexlab{}.
\newblock \showarticletitle{Language Models are Few-Shot Learners}.
\newblock \bibinfo{journal}{\emph{arXiv 2005.14165}} (\bibinfo{year}{2020}).
\newblock


\bibitem[\protect\citeauthoryear{Cai and Vasconcelos}{Cai and Vasconcelos}{2018}]%
        {CascadeRCNN2018}
\bibfield{author}{\bibinfo{person}{Zhaowei Cai} {and} \bibinfo{person}{Nuno Vasconcelos}.} \bibinfo{year}{2018}\natexlab{}.
\newblock \showarticletitle{Cascade R-CNN: Delving Into High Quality Object Detection}. In \bibinfo{booktitle}{\emph{Proceedings of the IEEE Conference on Computer Vision and Pattern Recognition (CVPR)}}.
\newblock


\bibitem[\protect\citeauthoryear{Caron, Misra, Mairal, Goyal, Bojanowski, and Joulin}{Caron et~al\mbox{.}}{2020}]%
        {caron2020unsupervised}
\bibfield{author}{\bibinfo{person}{Mathilde Caron}, \bibinfo{person}{Ishan Misra}, \bibinfo{person}{Julien Mairal}, \bibinfo{person}{Priya Goyal}, \bibinfo{person}{Piotr Bojanowski}, {and} \bibinfo{person}{Armand Joulin}.} \bibinfo{year}{2020}\natexlab{}.
\newblock \showarticletitle{Unsupervised learning of visual features by contrasting cluster assignments}.
\newblock \bibinfo{journal}{\emph{Advances in neural information processing systems}}  \bibinfo{volume}{33} (\bibinfo{year}{2020}), \bibinfo{pages}{9912--9924}.
\newblock


\bibitem[\protect\citeauthoryear{Carreira and Zisserman}{Carreira and Zisserman}{2017}]%
        {carreira2017quo}
\bibfield{author}{\bibinfo{person}{Joao Carreira} {and} \bibinfo{person}{Andrew Zisserman}.} \bibinfo{year}{2017}\natexlab{}.
\newblock \showarticletitle{Quo vadis, action recognition? a new model and the kinetics dataset}. In \bibinfo{booktitle}{\emph{proceedings of the IEEE Conference on Computer Vision and Pattern Recognition}}. \bibinfo{pages}{6299--6308}.
\newblock


\bibitem[\protect\citeauthoryear{Caruana}{Caruana}{1997}]%
        {caruana1997multitask}
\bibfield{author}{\bibinfo{person}{Rich Caruana}.} \bibinfo{year}{1997}\natexlab{}.
\newblock \showarticletitle{Multitask learning}.
\newblock \bibinfo{journal}{\emph{Machine learning}} \bibinfo{volume}{28}, \bibinfo{number}{1} (\bibinfo{year}{1997}), \bibinfo{pages}{41--75}.
\newblock


\bibitem[\protect\citeauthoryear{Carvalho, Pereira, and Cardoso}{Carvalho et~al\mbox{.}}{2019}]%
        {carvalho2019machine}
\bibfield{author}{\bibinfo{person}{Diogo~V Carvalho}, \bibinfo{person}{Eduardo~M Pereira}, {and} \bibinfo{person}{Jaime~S Cardoso}.} \bibinfo{year}{2019}\natexlab{}.
\newblock \showarticletitle{Machine learning interpretability: A survey on methods and metrics}.
\newblock \bibinfo{journal}{\emph{Electronics}} \bibinfo{volume}{8}, \bibinfo{number}{8} (\bibinfo{year}{2019}), \bibinfo{pages}{832}.
\newblock


\bibitem[\protect\citeauthoryear{Chaerle, Hagenbeek, De~Bruyne, Valcke, and Van Der~Straeten}{Chaerle et~al\mbox{.}}{2004}]%
        {chaerle2004thermal}
\bibfield{author}{\bibinfo{person}{Laury Chaerle}, \bibinfo{person}{Dik Hagenbeek}, \bibinfo{person}{Erik De~Bruyne}, \bibinfo{person}{Roland Valcke}, {and} \bibinfo{person}{Dominique Van Der~Straeten}.} \bibinfo{year}{2004}\natexlab{}.
\newblock \showarticletitle{Thermal and chlorophyll-fluorescence imaging distinguish plant-pathogen interactions at an early stage}.
\newblock \bibinfo{journal}{\emph{Plant and Cell Physiology}} \bibinfo{volume}{45}, \bibinfo{number}{7} (\bibinfo{year}{2004}), \bibinfo{pages}{887--896}.
\newblock


\bibitem[\protect\citeauthoryear{Chaerle, Leinonen, Jones, and Van Der~Straeten}{Chaerle et~al\mbox{.}}{2007}]%
        {chaerle2007monitoring}
\bibfield{author}{\bibinfo{person}{Laury Chaerle}, \bibinfo{person}{Ilkka Leinonen}, \bibinfo{person}{Hamlyn~G Jones}, {and} \bibinfo{person}{Dominique Van Der~Straeten}.} \bibinfo{year}{2007}\natexlab{}.
\newblock \showarticletitle{Monitoring and screening plant populations with combined thermal and chlorophyll fluorescence imaging}.
\newblock \bibinfo{journal}{\emph{Journal of experimental botany}} \bibinfo{volume}{58}, \bibinfo{number}{4} (\bibinfo{year}{2007}), \bibinfo{pages}{773--784}.
\newblock


\bibitem[\protect\citeauthoryear{Chaivivatrakul, Moonrinta, and Dailey}{Chaivivatrakul et~al\mbox{.}}{2010}]%
        {chaivivatrakul2010towards}
\bibfield{author}{\bibinfo{person}{Supawadee Chaivivatrakul}, \bibinfo{person}{Jednipat Moonrinta}, {and} \bibinfo{person}{Matthew~N Dailey}.} \bibinfo{year}{2010}\natexlab{}.
\newblock \showarticletitle{Towards Automated Crop Yield Estimation-Detection and 3D Reconstruction of Pineapples in Video Sequences.}. In \bibinfo{booktitle}{\emph{VISAPP (1)}}. \bibinfo{pages}{180--183}.
\newblock


\bibitem[\protect\citeauthoryear{Chandra, Desai, Guo, and Balasubramanian}{Chandra et~al\mbox{.}}{2020}]%
        {chandra2020computer}
\bibfield{author}{\bibinfo{person}{Akshay~L Chandra}, \bibinfo{person}{Sai~Vikas Desai}, \bibinfo{person}{Wei Guo}, {and} \bibinfo{person}{Vineeth~N Balasubramanian}.} \bibinfo{year}{2020}\natexlab{}.
\newblock \showarticletitle{Computer vision with deep learning for plant phenotyping in agriculture: A survey}.
\newblock \bibinfo{journal}{\emph{arXiv preprint arXiv:2006.11391}} (\bibinfo{year}{2020}).
\newblock


\bibitem[\protect\citeauthoryear{Cheein and Carelli}{Cheein and Carelli}{2013}]%
        {cheein2013agricultural}
\bibfield{author}{\bibinfo{person}{Fernando Alfredo~Auat Cheein} {and} \bibinfo{person}{Ricardo Carelli}.} \bibinfo{year}{2013}\natexlab{}.
\newblock \showarticletitle{Agricultural robotics: Unmanned robotic service units in agricultural tasks}.
\newblock \bibinfo{journal}{\emph{IEEE industrial electronics magazine}} \bibinfo{volume}{7}, \bibinfo{number}{3} (\bibinfo{year}{2013}), \bibinfo{pages}{48--58}.
\newblock


\bibitem[\protect\citeauthoryear{Chefer, Gur, and Wolf}{Chefer et~al\mbox{.}}{2021}]%
        {chefer2021transformer}
\bibfield{author}{\bibinfo{person}{Hila Chefer}, \bibinfo{person}{Shir Gur}, {and} \bibinfo{person}{Lior Wolf}.} \bibinfo{year}{2021}\natexlab{}.
\newblock \showarticletitle{Transformer interpretability beyond attention visualization}. In \bibinfo{booktitle}{\emph{Proceedings of the IEEE/CVF Conference on Computer Vision and Pattern Recognition}}. \bibinfo{pages}{782--791}.
\newblock


\bibitem[\protect\citeauthoryear{Chen, He, Narasimhan, and Chen}{Chen et~al\mbox{.}}{2022}]%
        {chen2022can}
\bibfield{author}{\bibinfo{person}{Howard Chen}, \bibinfo{person}{Jacqueline He}, \bibinfo{person}{Karthik Narasimhan}, {and} \bibinfo{person}{Danqi Chen}.} \bibinfo{year}{2022}\natexlab{}.
\newblock \showarticletitle{Can Rationalization Improve Robustness?}
\newblock \bibinfo{journal}{\emph{arXiv preprint arXiv:2204.11790}} (\bibinfo{year}{2022}).
\newblock


\bibitem[\protect\citeauthoryear{Chen, Zheng, and Ji}{Chen et~al\mbox{.}}{2020b}]%
        {chen2020generating}
\bibfield{author}{\bibinfo{person}{Hanjie Chen}, \bibinfo{person}{Guangtao Zheng}, {and} \bibinfo{person}{Yangfeng Ji}.} \bibinfo{year}{2020}\natexlab{b}.
\newblock \showarticletitle{Generating hierarchical explanations on text classification via feature interaction detection}.
\newblock \bibinfo{journal}{\emph{arXiv preprint arXiv:2004.02015}} (\bibinfo{year}{2020}).
\newblock


\bibitem[\protect\citeauthoryear{Chen, Pang, Wang, Xiong, Li, Sun, Feng, Liu, Shi, Ouyang, et~al\mbox{.}}{Chen et~al\mbox{.}}{2019a}]%
        {chen2019hybrid}
\bibfield{author}{\bibinfo{person}{Kai Chen}, \bibinfo{person}{Jiangmiao Pang}, \bibinfo{person}{Jiaqi Wang}, \bibinfo{person}{Yu Xiong}, \bibinfo{person}{Xiaoxiao Li}, \bibinfo{person}{Shuyang Sun}, \bibinfo{person}{Wansen Feng}, \bibinfo{person}{Ziwei Liu}, \bibinfo{person}{Jianping Shi}, \bibinfo{person}{Wanli Ouyang}, {et~al\mbox{.}}} \bibinfo{year}{2019}\natexlab{a}.
\newblock \showarticletitle{Hybrid task cascade for instance segmentation}. In \bibinfo{booktitle}{\emph{Proceedings of the IEEE/CVF Conference on Computer Vision and Pattern Recognition}}. \bibinfo{pages}{4974--4983}.
\newblock


\bibitem[\protect\citeauthoryear{Chen, Strauch, and Merhof}{Chen et~al\mbox{.}}{2019b}]%
        {ChenLong-Object-aware-Embedding:2019}
\bibfield{author}{\bibinfo{person}{Long Chen}, \bibinfo{person}{Martin Strauch}, {and} \bibinfo{person}{Dorit Merhof}.} \bibinfo{year}{2019}\natexlab{b}.
\newblock \showarticletitle{Instance Segmentation of Biomedical Images with an Object-aware Embedding Learned with Local Constraints}. In \bibinfo{booktitle}{\emph{International Conference on Medical Image Computing and Computer-Assisted Intervention}}.
\newblock
\urldef\tempurl%
\url{https://doi.org/10.48550/ARXIV.2004.09821}
\showDOI{\tempurl}


\bibitem[\protect\citeauthoryear{Chen, Papandreou, Kokkinos, Murphy, and Yuille}{Chen et~al\mbox{.}}{2017}]%
        {chen2017deeplab}
\bibfield{author}{\bibinfo{person}{Liang-Chieh Chen}, \bibinfo{person}{George Papandreou}, \bibinfo{person}{Iasonas Kokkinos}, \bibinfo{person}{Kevin Murphy}, {and} \bibinfo{person}{Alan~L Yuille}.} \bibinfo{year}{2017}\natexlab{}.
\newblock \showarticletitle{Deeplab: Semantic image segmentation with deep convolutional nets, atrous convolution, and fully connected crfs}.
\newblock \bibinfo{journal}{\emph{IEEE transactions on pattern analysis and machine intelligence}} \bibinfo{volume}{40}, \bibinfo{number}{4} (\bibinfo{year}{2017}), \bibinfo{pages}{834--848}.
\newblock


\bibitem[\protect\citeauthoryear{Chen, Zhu, Papandreou, Schroff, and Adam}{Chen et~al\mbox{.}}{2018}]%
        {chen2018encoder}
\bibfield{author}{\bibinfo{person}{Liang-Chieh Chen}, \bibinfo{person}{Yukun Zhu}, \bibinfo{person}{George Papandreou}, \bibinfo{person}{Florian Schroff}, {and} \bibinfo{person}{Hartwig Adam}.} \bibinfo{year}{2018}\natexlab{}.
\newblock \showarticletitle{Encoder-decoder with atrous separable convolution for semantic image segmentation}. In \bibinfo{booktitle}{\emph{Proceedings of the European conference on computer vision (ECCV)}}. \bibinfo{pages}{801--818}.
\newblock


\bibitem[\protect\citeauthoryear{Chen, Kornblith, Norouzi, and Hinton}{Chen et~al\mbox{.}}{2020a}]%
        {chen2020simple}
\bibfield{author}{\bibinfo{person}{Ting Chen}, \bibinfo{person}{Simon Kornblith}, \bibinfo{person}{Mohammad Norouzi}, {and} \bibinfo{person}{Geoffrey Hinton}.} \bibinfo{year}{2020}\natexlab{a}.
\newblock \showarticletitle{A Simple Framework for Contrastive Learning of Visual Representations}.
\newblock \bibinfo{journal}{\emph{arXiv 2002.05709}} (\bibinfo{year}{2020}).
\newblock


\bibitem[\protect\citeauthoryear{Chen, Saxena, Li, Fleet, and Hinton}{Chen et~al\mbox{.}}{2021b}]%
        {chen2021pix2seq}
\bibfield{author}{\bibinfo{person}{Ting Chen}, \bibinfo{person}{Saurabh Saxena}, \bibinfo{person}{Lala Li}, \bibinfo{person}{David~J Fleet}, {and} \bibinfo{person}{Geoffrey Hinton}.} \bibinfo{year}{2021}\natexlab{b}.
\newblock \showarticletitle{Pix2seq: A language modeling framework for object detection}.
\newblock \bibinfo{journal}{\emph{arXiv preprint arXiv:2109.10852}} (\bibinfo{year}{2021}).
\newblock


\bibitem[\protect\citeauthoryear{Chen, Li, Yu, Wu, and Miao}{Chen et~al\mbox{.}}{2021a}]%
        {chen2021hydra}
\bibfield{author}{\bibinfo{person}{Yuanyuan Chen}, \bibinfo{person}{Boyang Li}, \bibinfo{person}{Han Yu}, \bibinfo{person}{Pengcheng Wu}, {and} \bibinfo{person}{Chunyan Miao}.} \bibinfo{year}{2021}\natexlab{a}.
\newblock \showarticletitle{Hydra: Hypergradient data relevance analysis for interpreting deep neural networks}. In \bibinfo{booktitle}{\emph{Proceedings of the AAAI Conference on Artificial Intelligence}}, Vol.~\bibinfo{volume}{35}. \bibinfo{pages}{7081--7089}.
\newblock


\bibitem[\protect\citeauthoryear{Chen, Liu, and Yang}{Chen et~al\mbox{.}}{2015}]%
        {ChenYi-Ting2015}
\bibfield{author}{\bibinfo{person}{Yi-Ting Chen}, \bibinfo{person}{Xiaokai Liu}, {and} \bibinfo{person}{Ming-Hsuan Yang}.} \bibinfo{year}{2015}\natexlab{}.
\newblock \showarticletitle{Multi-instance object segmentation with occlusion handling}. In \bibinfo{booktitle}{\emph{2015 IEEE Conference on Computer Vision and Pattern Recognition (CVPR)}}. \bibinfo{pages}{3470--3478}.
\newblock
\urldef\tempurl%
\url{https://doi.org/10.1109/CVPR.2015.7298969}
\showDOI{\tempurl}


\bibitem[\protect\citeauthoryear{Cheng, Collins, Zhu, Liu, Huang, Adam, and Chen}{Cheng et~al\mbox{.}}{2020}]%
        {ChengBowen2020}
\bibfield{author}{\bibinfo{person}{Bowen Cheng}, \bibinfo{person}{Maxwell~D. Collins}, \bibinfo{person}{Yukun Zhu}, \bibinfo{person}{Ting Liu}, \bibinfo{person}{Thomas~S. Huang}, \bibinfo{person}{Hartwig Adam}, {and} \bibinfo{person}{Liang-Chieh Chen}.} \bibinfo{year}{2020}\natexlab{}.
\newblock \showarticletitle{Panoptic-DeepLab: A Simple, Strong, and Fast Baseline for Bottom-Up Panoptic Segmentation}. In \bibinfo{booktitle}{\emph{2020 IEEE/CVF Conference on Computer Vision and Pattern Recognition (CVPR)}}. \bibinfo{pages}{12472--12482}.
\newblock
\urldef\tempurl%
\url{https://doi.org/10.1109/CVPR42600.2020.01249}
\showDOI{\tempurl}


\bibitem[\protect\citeauthoryear{Cheng, Liu, Ning, Wang, Han, Niu, Gao, and Sugiyama}{Cheng et~al\mbox{.}}{2022}]%
        {cheng2022instance}
\bibfield{author}{\bibinfo{person}{De Cheng}, \bibinfo{person}{Tongliang Liu}, \bibinfo{person}{Yixiong Ning}, \bibinfo{person}{Nannan Wang}, \bibinfo{person}{Bo Han}, \bibinfo{person}{Gang Niu}, \bibinfo{person}{Xinbo Gao}, {and} \bibinfo{person}{Masashi Sugiyama}.} \bibinfo{year}{2022}\natexlab{}.
\newblock \showarticletitle{Instance-Dependent Label-Noise Learning with Manifold-Regularized Transition Matrix Estimation}. In \bibinfo{booktitle}{\emph{Proceedings of the IEEE/CVF Conference on Computer Vision and Pattern Recognition}}. \bibinfo{pages}{16630--16639}.
\newblock


\bibitem[\protect\citeauthoryear{Cho, Lei, Tan, and Bansal}{Cho et~al\mbox{.}}{2021}]%
        {cho2021unifying}
\bibfield{author}{\bibinfo{person}{Jaemin Cho}, \bibinfo{person}{Jie Lei}, \bibinfo{person}{Hao Tan}, {and} \bibinfo{person}{Mohit Bansal}.} \bibinfo{year}{2021}\natexlab{}.
\newblock \showarticletitle{Unifying vision-and-language tasks via text generation}. In \bibinfo{booktitle}{\emph{International Conference on Machine Learning}}. PMLR, \bibinfo{pages}{1931--1942}.
\newblock


\bibitem[\protect\citeauthoryear{Chu, Tian, Wang, Zhang, Ren, Wei, Xia, and Shen}{Chu et~al\mbox{.}}{2021}]%
        {Twins2021}
\bibfield{author}{\bibinfo{person}{Xiangxiang Chu}, \bibinfo{person}{Zhi Tian}, \bibinfo{person}{Yuqing Wang}, \bibinfo{person}{Bo Zhang}, \bibinfo{person}{Haibing Ren}, \bibinfo{person}{Xiaolin Wei}, \bibinfo{person}{Huaxia Xia}, {and} \bibinfo{person}{Chunhua Shen}.} \bibinfo{year}{2021}\natexlab{}.
\newblock \showarticletitle{Twins: Revisiting the Design of Spatial Attention in Vision Transformers}.
\newblock \bibinfo{journal}{\emph{arXiv 2104.13840}} (\bibinfo{year}{2021}).
\newblock


\bibitem[\protect\citeauthoryear{Ciresan, Giusti, Gambardella, and Schmidhuber}{Ciresan et~al\mbox{.}}{2012}]%
        {Ciresan:NIPS2012}
\bibfield{author}{\bibinfo{person}{Dan Ciresan}, \bibinfo{person}{Alessandro Giusti}, \bibinfo{person}{Luca Gambardella}, {and} \bibinfo{person}{J\"{u}rgen Schmidhuber}.} \bibinfo{year}{2012}\natexlab{}.
\newblock \showarticletitle{Deep Neural Networks Segment Neuronal Membranes in Electron Microscopy Images}. In \bibinfo{booktitle}{\emph{Advances in Neural Information Processing Systems}}, \bibfield{editor}{\bibinfo{person}{F.~Pereira}, \bibinfo{person}{C.J. Burges}, \bibinfo{person}{L.~Bottou}, {and} \bibinfo{person}{K.Q. Weinberger}} (Eds.), Vol.~\bibinfo{volume}{25}. \bibinfo{publisher}{Curran Associates, Inc.}
\newblock
\urldef\tempurl%
\url{https://proceedings.neurips.cc/paper/2012/file/459a4ddcb586f24efd9395aa7662bc7c-Paper.pdf}
\showURL{%
\tempurl}


\bibitem[\protect\citeauthoryear{Cubero, Lee, Aleixos, Albert, and Blasco}{Cubero et~al\mbox{.}}{2016}]%
        {cubero2016automated}
\bibfield{author}{\bibinfo{person}{Sergio Cubero}, \bibinfo{person}{Won~Suk Lee}, \bibinfo{person}{Nuria Aleixos}, \bibinfo{person}{Francisco Albert}, {and} \bibinfo{person}{Jose Blasco}.} \bibinfo{year}{2016}\natexlab{}.
\newblock \showarticletitle{Automated systems based on machine vision for inspecting citrus fruits from the field to postharvest—a review}.
\newblock \bibinfo{journal}{\emph{Food and Bioprocess Technology}}  \bibinfo{volume}{9} (\bibinfo{year}{2016}), \bibinfo{pages}{1623--1639}.
\newblock


\bibitem[\protect\citeauthoryear{Cubuk, Zoph, Shlens, and Le}{Cubuk et~al\mbox{.}}{2019}]%
        {RandAugment2019}
\bibfield{author}{\bibinfo{person}{Ekin~D. Cubuk}, \bibinfo{person}{Barret Zoph}, \bibinfo{person}{Jonathon Shlens}, {and} \bibinfo{person}{Quoc~V. Le}.} \bibinfo{year}{2019}\natexlab{}.
\newblock \showarticletitle{RandAugment: Practical automated data augmentation with a reduced search space}.
\newblock \bibinfo{journal}{\emph{arXiv 1909.13719}} (\bibinfo{year}{2019}).
\newblock


\bibitem[\protect\citeauthoryear{Dai, He, Li, Ren, and Sun}{Dai et~al\mbox{.}}{2016}]%
        {DaiJifeng2016:Instance}
\bibfield{author}{\bibinfo{person}{Jifeng Dai}, \bibinfo{person}{Kaiming He}, \bibinfo{person}{Yi Li}, \bibinfo{person}{Shaoqing Ren}, {and} \bibinfo{person}{Jian Sun}.} \bibinfo{year}{2016}\natexlab{}.
\newblock \showarticletitle{Instance-Sensitive Fully Convolutional Networks}. In \bibinfo{booktitle}{\emph{Computer Vision -- ECCV 2016}}, \bibfield{editor}{\bibinfo{person}{Bastian Leibe}, \bibinfo{person}{Jiri Matas}, \bibinfo{person}{Nicu Sebe}, {and} \bibinfo{person}{Max Welling}} (Eds.). \bibinfo{publisher}{Springer International Publishing}, \bibinfo{address}{Cham}, \bibinfo{pages}{534--549}.
\newblock
\showISBNx{978-3-319-46466-4}


\bibitem[\protect\citeauthoryear{Dai, Qi, Xiong, Li, Zhang, Hu, and Wei}{Dai et~al\mbox{.}}{2017}]%
        {DeformableCN2017}
\bibfield{author}{\bibinfo{person}{Jifeng Dai}, \bibinfo{person}{Haozhi Qi}, \bibinfo{person}{Yuwen Xiong}, \bibinfo{person}{Yi Li}, \bibinfo{person}{Guodong Zhang}, \bibinfo{person}{Han Hu}, {and} \bibinfo{person}{Yichen Wei}.} \bibinfo{year}{2017}\natexlab{}.
\newblock \showarticletitle{Deformable Convolutional Networks}. In \bibinfo{booktitle}{\emph{2017 IEEE International Conference on Computer Vision (ICCV)}}. \bibinfo{pages}{764--773}.
\newblock


\bibitem[\protect\citeauthoryear{Dai, Liu, Le, and Tan}{Dai et~al\mbox{.}}{2021}]%
        {dai2021coatnet}
\bibfield{author}{\bibinfo{person}{Zihang Dai}, \bibinfo{person}{Hanxiao Liu}, \bibinfo{person}{Quoc~V Le}, {and} \bibinfo{person}{Mingxing Tan}.} \bibinfo{year}{2021}\natexlab{}.
\newblock \showarticletitle{CoAtNet: Marrying Convolution and Attention for All Data Sizes}.
\newblock \bibinfo{journal}{\emph{arXiv preprint arXiv:2106.04803}} (\bibinfo{year}{2021}).
\newblock


\bibitem[\protect\citeauthoryear{Davis, Liang, Enouen, and Ilin}{Davis et~al\mbox{.}}{2019}]%
        {davis2019hierarchical}
\bibfield{author}{\bibinfo{person}{Jim Davis}, \bibinfo{person}{Tong Liang}, \bibinfo{person}{James Enouen}, {and} \bibinfo{person}{Roman Ilin}.} \bibinfo{year}{2019}\natexlab{}.
\newblock \showarticletitle{Hierarchical semantic labeling with adaptive confidence}. In \bibinfo{booktitle}{\emph{International Symposium on Visual Computing}}. Springer, \bibinfo{pages}{169--183}.
\newblock


\bibitem[\protect\citeauthoryear{De~Brabandere, Neven, and Van~Gool}{De~Brabandere et~al\mbox{.}}{2017}]%
        {Brabandere-InstSeg-DiscriminativeLoss2017}
\bibfield{author}{\bibinfo{person}{Bert De~Brabandere}, \bibinfo{person}{Davy Neven}, {and} \bibinfo{person}{Luc Van~Gool}.} \bibinfo{year}{2017}\natexlab{}.
\newblock \showarticletitle{Semantic Instance Segmentation with a Discriminative Loss Function}. \bibinfo{publisher}{CVPR 2017 Workshop on Deep Learning for Robotic Vision}.
\newblock
\urldef\tempurl%
\url{https://arxiv.org/abs/1708.02551}
\showURL{%
\tempurl}


\bibitem[\protect\citeauthoryear{Deng, Guo, Xue, and Zafeiriou}{Deng et~al\mbox{.}}{2019}]%
        {deng2019arcface}
\bibfield{author}{\bibinfo{person}{Jiankang Deng}, \bibinfo{person}{Jia Guo}, \bibinfo{person}{Niannan Xue}, {and} \bibinfo{person}{Stefanos Zafeiriou}.} \bibinfo{year}{2019}\natexlab{}.
\newblock \showarticletitle{Arcface: Additive angular margin loss for deep face recognition}. In \bibinfo{booktitle}{\emph{CVPR}}. \bibinfo{pages}{4690--4699}.
\newblock


\bibitem[\protect\citeauthoryear{Despommier}{Despommier}{2010}]%
        {despommier2010vertical}
\bibfield{author}{\bibinfo{person}{Dickson Despommier}.} \bibinfo{year}{2010}\natexlab{}.
\newblock \bibinfo{booktitle}{\emph{The vertical farm: feeding the world in the 21st century}}.
\newblock \bibinfo{publisher}{Macmillan}.
\newblock


\bibitem[\protect\citeauthoryear{Devlin, Chang, Lee, and Toutanova}{Devlin et~al\mbox{.}}{2019}]%
        {devlin2019bert}
\bibfield{author}{\bibinfo{person}{Jacob Devlin}, \bibinfo{person}{Ming-Wei Chang}, \bibinfo{person}{Kenton Lee}, {and} \bibinfo{person}{Kristina Toutanova}.} \bibinfo{year}{2019}\natexlab{}.
\newblock \showarticletitle{BERT: Pre-training of Deep Bidirectional Transformers for Language Understanding}.
\newblock \bibinfo{journal}{\emph{arXiv 1810.04805}} (\bibinfo{year}{2019}).
\newblock


\bibitem[\protect\citeauthoryear{Dhurandhar, Chen, Luss, Tu, Ting, Shanmugam, and Das}{Dhurandhar et~al\mbox{.}}{2018}]%
        {dhurandhar2018explanations}
\bibfield{author}{\bibinfo{person}{Amit Dhurandhar}, \bibinfo{person}{Pin-Yu Chen}, \bibinfo{person}{Ronny Luss}, \bibinfo{person}{Chun-Chen Tu}, \bibinfo{person}{Paishun Ting}, \bibinfo{person}{Karthikeyan Shanmugam}, {and} \bibinfo{person}{Payel Das}.} \bibinfo{year}{2018}\natexlab{}.
\newblock \showarticletitle{Explanations based on the missing: Towards contrastive explanations with pertinent negatives}.
\newblock \bibinfo{journal}{\emph{Advances in neural information processing systems}}  \bibinfo{volume}{31} (\bibinfo{year}{2018}).
\newblock


\bibitem[\protect\citeauthoryear{Dias, Tabb, and Medeiros}{Dias et~al\mbox{.}}{2018a}]%
        {dias2018multispecies}
\bibfield{author}{\bibinfo{person}{Philipe~A Dias}, \bibinfo{person}{Amy Tabb}, {and} \bibinfo{person}{Henry Medeiros}.} \bibinfo{year}{2018}\natexlab{a}.
\newblock \showarticletitle{Multispecies fruit flower detection using a refined semantic segmentation network}.
\newblock \bibinfo{journal}{\emph{IEEE robotics and automation letters}} \bibinfo{volume}{3}, \bibinfo{number}{4} (\bibinfo{year}{2018}), \bibinfo{pages}{3003--3010}.
\newblock


\bibitem[\protect\citeauthoryear{Dias, Tabb, and Medeiros}{Dias et~al\mbox{.}}{2018b}]%
        {8392727}
\bibfield{author}{\bibinfo{person}{Philipe~A. Dias}, \bibinfo{person}{Amy Tabb}, {and} \bibinfo{person}{Henry Medeiros}.} \bibinfo{year}{2018}\natexlab{b}.
\newblock \showarticletitle{Multispecies Fruit Flower Detection Using a Refined Semantic Segmentation Network}.
\newblock \bibinfo{journal}{\emph{IEEE Robotics and Automation Letters}} \bibinfo{volume}{3}, \bibinfo{number}{4} (\bibinfo{year}{2018}), \bibinfo{pages}{3003--3010}.
\newblock
\urldef\tempurl%
\url{https://doi.org/10.1109/LRA.2018.2849498}
\showDOI{\tempurl}


\bibitem[\protect\citeauthoryear{Ding, Han, Liu, and Niethammer}{Ding et~al\mbox{.}}{2021}]%
        {Ding_2021_ICCV}
\bibfield{author}{\bibinfo{person}{Zhipeng Ding}, \bibinfo{person}{Xu Han}, \bibinfo{person}{Peirong Liu}, {and} \bibinfo{person}{Marc Niethammer}.} \bibinfo{year}{2021}\natexlab{}.
\newblock \showarticletitle{Local Temperature Scaling for Probability Calibration}. In \bibinfo{booktitle}{\emph{Proceedings of the IEEE/CVF International Conference on Computer Vision (ICCV)}}. \bibinfo{pages}{6889--6899}.
\newblock


\bibitem[\protect\citeauthoryear{Dogan, Li, Sigal, and Gross}{Dogan et~al\mbox{.}}{2018}]%
        {PelinDogan-2018}
\bibfield{author}{\bibinfo{person}{Pelin Dogan}, \bibinfo{person}{Boyang Li}, \bibinfo{person}{Leonid Sigal}, {and} \bibinfo{person}{Markus Gross}.} \bibinfo{year}{2018}\natexlab{}.
\newblock \showarticletitle{A Neural Multi-sequence Alignment TeCHnique (NeuMATCH)}. In \bibinfo{booktitle}{\emph{The Conference on Computer Vision and Pattern Recognition (CVPR)}}.
\newblock


\bibitem[\protect\citeauthoryear{Domingues, Brand{\~a}o, and Ferreira}{Domingues et~al\mbox{.}}{2022}]%
        {domingues2022machine}
\bibfield{author}{\bibinfo{person}{Tiago Domingues}, \bibinfo{person}{Tom{\'a}s Brand{\~a}o}, {and} \bibinfo{person}{Jo{\~a}o~C Ferreira}.} \bibinfo{year}{2022}\natexlab{}.
\newblock \showarticletitle{Machine Learning for Detection and Prediction of Crop Diseases and Pests: A Comprehensive Survey}.
\newblock \bibinfo{journal}{\emph{Agriculture}} \bibinfo{volume}{12}, \bibinfo{number}{9} (\bibinfo{year}{2022}), \bibinfo{pages}{1350}.
\newblock


\bibitem[\protect\citeauthoryear{Dosovitskiy, Beyer, Kolesnikov, Weissenborn, Zhai, Unterthiner, Dehghani, Minderer, Heigold, Gelly, et~al\mbox{.}}{Dosovitskiy et~al\mbox{.}}{2021}]%
        {dosovitskiy2020:ViT}
\bibfield{author}{\bibinfo{person}{Alexey Dosovitskiy}, \bibinfo{person}{Lucas Beyer}, \bibinfo{person}{Alexander Kolesnikov}, \bibinfo{person}{Dirk Weissenborn}, \bibinfo{person}{Xiaohua Zhai}, \bibinfo{person}{Thomas Unterthiner}, \bibinfo{person}{Mostafa Dehghani}, \bibinfo{person}{Matthias Minderer}, \bibinfo{person}{Georg Heigold}, \bibinfo{person}{Sylvain Gelly}, {et~al\mbox{.}}} \bibinfo{year}{2021}\natexlab{}.
\newblock \showarticletitle{An image is worth 16x16 words: Transformers for image recognition at scale}. In \bibinfo{booktitle}{\emph{International Conference on Learning Representations}}.
\newblock


\bibitem[\protect\citeauthoryear{Duan, Bai, Xie, Qi, Huang, and Tian}{Duan et~al\mbox{.}}{2019}]%
        {CornerNet2019}
\bibfield{author}{\bibinfo{person}{Kaiwen Duan}, \bibinfo{person}{Song Bai}, \bibinfo{person}{Lingxi Xie}, \bibinfo{person}{Honggang Qi}, \bibinfo{person}{Qingming Huang}, {and} \bibinfo{person}{Qi Tian}.} \bibinfo{year}{2019}\natexlab{}.
\newblock \showarticletitle{CenterNet: Keypoint Triplets for Object Detection}. In \bibinfo{booktitle}{\emph{Proceedings of the IEEE/CVF International Conference on Computer Vision (ICCV)}}.
\newblock


\bibitem[\protect\citeauthoryear{Duckett, Pearson, Blackmore, Grieve, Chen, Cielniak, Cleaversmith, Dai, Davis, Fox, et~al\mbox{.}}{Duckett et~al\mbox{.}}{2018}]%
        {duckett2018agricultural}
\bibfield{author}{\bibinfo{person}{Tom Duckett}, \bibinfo{person}{Simon Pearson}, \bibinfo{person}{Simon Blackmore}, \bibinfo{person}{Bruce Grieve}, \bibinfo{person}{Wen-Hua Chen}, \bibinfo{person}{Grzegorz Cielniak}, \bibinfo{person}{Jason Cleaversmith}, \bibinfo{person}{Jian Dai}, \bibinfo{person}{Steve Davis}, \bibinfo{person}{Charles Fox}, {et~al\mbox{.}}} \bibinfo{year}{2018}\natexlab{}.
\newblock \showarticletitle{Agricultural robotics: the future of robotic agriculture}.
\newblock \bibinfo{journal}{\emph{arXiv preprint arXiv:1806.06762}} (\bibinfo{year}{2018}).
\newblock


\bibitem[\protect\citeauthoryear{Erhan, Bengio, Courville, and Vincent}{Erhan et~al\mbox{.}}{2009}]%
        {erhan2009visualizing}
\bibfield{author}{\bibinfo{person}{Dumitru Erhan}, \bibinfo{person}{Yoshua Bengio}, \bibinfo{person}{Aaron Courville}, {and} \bibinfo{person}{Pascal Vincent}.} \bibinfo{year}{2009}\natexlab{}.
\newblock \showarticletitle{Visualizing higher-layer features of a deep network}.
\newblock \bibinfo{journal}{\emph{University of Montreal}} \bibinfo{volume}{1341}, \bibinfo{number}{3} (\bibinfo{year}{2009}), \bibinfo{pages}{1}.
\newblock


\bibitem[\protect\citeauthoryear{Farabet, Couprie, Najman, and LeCun}{Farabet et~al\mbox{.}}{2013}]%
        {Farabet:TPAMI2013}
\bibfield{author}{\bibinfo{person}{Clement Farabet}, \bibinfo{person}{Camille Couprie}, \bibinfo{person}{Laurent Najman}, {and} \bibinfo{person}{Yann LeCun}.} \bibinfo{year}{2013}\natexlab{}.
\newblock \showarticletitle{Learning Hierarchical Features for Scene Labeling}.
\newblock \bibinfo{journal}{\emph{IEEE Transactions on Pattern Analysis and Machine Intelligence}} \bibinfo{volume}{35}, \bibinfo{number}{8} (\bibinfo{year}{2013}), \bibinfo{pages}{1915--1929}.
\newblock
\urldef\tempurl%
\url{https://doi.org/10.1109/TPAMI.2012.231}
\showDOI{\tempurl}


\bibitem[\protect\citeauthoryear{Ferentinos}{Ferentinos}{2018}]%
        {ferentinos2018deep}
\bibfield{author}{\bibinfo{person}{Konstantinos~P Ferentinos}.} \bibinfo{year}{2018}\natexlab{}.
\newblock \showarticletitle{Deep learning models for plant disease detection and diagnosis}.
\newblock \bibinfo{journal}{\emph{Computers and electronics in agriculture}}  \bibinfo{volume}{145} (\bibinfo{year}{2018}), \bibinfo{pages}{311--318}.
\newblock


\bibitem[\protect\citeauthoryear{Fern{\'a}ndez, Salinas, Montes, and Sarria}{Fern{\'a}ndez et~al\mbox{.}}{2014}]%
        {fernandez2014multisensory}
\bibfield{author}{\bibinfo{person}{Roemi Fern{\'a}ndez}, \bibinfo{person}{Carlota Salinas}, \bibinfo{person}{H{\'e}ctor Montes}, {and} \bibinfo{person}{Javier Sarria}.} \bibinfo{year}{2014}\natexlab{}.
\newblock \showarticletitle{Multisensory system for fruit harvesting robots. Experimental testing in natural scenarios and with different kinds of crops}.
\newblock \bibinfo{journal}{\emph{Sensors}} \bibinfo{volume}{14}, \bibinfo{number}{12} (\bibinfo{year}{2014}), \bibinfo{pages}{23885--23904}.
\newblock


\bibitem[\protect\citeauthoryear{Fong and Vedaldi}{Fong and Vedaldi}{2018}]%
        {fong2018net2vec}
\bibfield{author}{\bibinfo{person}{Ruth Fong} {and} \bibinfo{person}{Andrea Vedaldi}.} \bibinfo{year}{2018}\natexlab{}.
\newblock \showarticletitle{Net2vec: Quantifying and explaining how concepts are encoded by filters in deep neural networks}. In \bibinfo{booktitle}{\emph{Proceedings of the IEEE conference on computer vision and pattern recognition}}. \bibinfo{pages}{8730--8738}.
\newblock


\bibitem[\protect\citeauthoryear{F{\"o}rster, Behley, Behmann, and Roscher}{F{\"o}rster et~al\mbox{.}}{2019}]%
        {forster2019hyperspectral}
\bibfield{author}{\bibinfo{person}{Alina F{\"o}rster}, \bibinfo{person}{Jens Behley}, \bibinfo{person}{Jan Behmann}, {and} \bibinfo{person}{Ribana Roscher}.} \bibinfo{year}{2019}\natexlab{}.
\newblock \showarticletitle{Hyperspectral plant disease forecasting using generative adversarial networks}. In \bibinfo{booktitle}{\emph{IGARSS 2019-2019 IEEE International Geoscience and Remote Sensing Symposium}}. IEEE, \bibinfo{pages}{1793--1796}.
\newblock


\bibitem[\protect\citeauthoryear{Frank, Wiegman, Davis, and Shearer}{Frank et~al\mbox{.}}{2021}]%
        {frank2021confidence}
\bibfield{author}{\bibinfo{person}{Logan Frank}, \bibinfo{person}{Christopher Wiegman}, \bibinfo{person}{Jim Davis}, {and} \bibinfo{person}{Scott Shearer}.} \bibinfo{year}{2021}\natexlab{}.
\newblock \showarticletitle{Confidence-Driven Hierarchical Classification of Cultivated Plant Stresses}. In \bibinfo{booktitle}{\emph{Proceedings of the IEEE/CVF Winter Conference on Applications of Computer Vision}}. \bibinfo{pages}{2503--2512}.
\newblock


\bibitem[\protect\citeauthoryear{Fraser and Campbell}{Fraser and Campbell}{2019}]%
        {fraser2019agriculture}
\bibfield{author}{\bibinfo{person}{Evan~DG Fraser} {and} \bibinfo{person}{Malcolm Campbell}.} \bibinfo{year}{2019}\natexlab{}.
\newblock \showarticletitle{Agriculture 5.0: reconciling production with planetary health}.
\newblock \bibinfo{journal}{\emph{One Earth}} \bibinfo{volume}{1}, \bibinfo{number}{3} (\bibinfo{year}{2019}), \bibinfo{pages}{278--280}.
\newblock


\bibitem[\protect\citeauthoryear{Fu, Liu, Ranga, Tyagi, and Berg}{Fu et~al\mbox{.}}{2017}]%
        {fu2017dssd}
\bibfield{author}{\bibinfo{person}{Cheng-Yang Fu}, \bibinfo{person}{Wei Liu}, \bibinfo{person}{Ananth Ranga}, \bibinfo{person}{Ambrish Tyagi}, {and} \bibinfo{person}{Alexander~C. Berg}.} \bibinfo{year}{2017}\natexlab{}.
\newblock \showarticletitle{DSSD : Deconvolutional Single Shot Detector}.
\newblock \bibinfo{journal}{\emph{arXiv Preprint 1701.06659}} (\bibinfo{year}{2017}).
\newblock


\bibitem[\protect\citeauthoryear{Fu}{Fu}{1991}]%
        {fu1991rule}
\bibfield{author}{\bibinfo{person}{LiMin Fu}.} \bibinfo{year}{1991}\natexlab{}.
\newblock \showarticletitle{Rule Learning by Searching on Adapted Nets.}. In \bibinfo{booktitle}{\emph{AAAI}}, Vol.~\bibinfo{volume}{91}. \bibinfo{pages}{590--595}.
\newblock


\bibitem[\protect\citeauthoryear{Fuentes, Yoon, Kim, and Park}{Fuentes et~al\mbox{.}}{2017}]%
        {fuentes2017robust}
\bibfield{author}{\bibinfo{person}{Alvaro Fuentes}, \bibinfo{person}{Sook Yoon}, \bibinfo{person}{Sang~Cheol Kim}, {and} \bibinfo{person}{Dong~Sun Park}.} \bibinfo{year}{2017}\natexlab{}.
\newblock \showarticletitle{A robust deep-learning-based detector for real-time tomato plant diseases and pests recognition}.
\newblock \bibinfo{journal}{\emph{Sensors}} \bibinfo{volume}{17}, \bibinfo{number}{9} (\bibinfo{year}{2017}), \bibinfo{pages}{2022}.
\newblock


\bibitem[\protect\citeauthoryear{Galeano, Joseph, and Lillo}{Galeano et~al\mbox{.}}{2015}]%
        {galeano2015mahalanobis}
\bibfield{author}{\bibinfo{person}{Pedro Galeano}, \bibinfo{person}{Esdras Joseph}, {and} \bibinfo{person}{Rosa~E Lillo}.} \bibinfo{year}{2015}\natexlab{}.
\newblock \showarticletitle{The Mahalanobis distance for functional data with applications to classification}.
\newblock \bibinfo{journal}{\emph{Technometrics}} \bibinfo{volume}{57}, \bibinfo{number}{2} (\bibinfo{year}{2015}), \bibinfo{pages}{281--291}.
\newblock


\bibitem[\protect\citeauthoryear{Garreau and Mardaoui}{Garreau and Mardaoui}{2021}]%
        {pmlr-v139-garreau21a}
\bibfield{author}{\bibinfo{person}{Damien Garreau} {and} \bibinfo{person}{Dina Mardaoui}.} \bibinfo{year}{2021}\natexlab{}.
\newblock \showarticletitle{What does LIME really see in images?}. In \bibinfo{booktitle}{\emph{Proceedings of the 38th International Conference on Machine Learning}} \emph{(\bibinfo{series}{Proceedings of Machine Learning Research}, Vol.~\bibinfo{volume}{139})}, \bibfield{editor}{\bibinfo{person}{Marina Meila} {and} \bibinfo{person}{Tong Zhang}} (Eds.). \bibinfo{publisher}{PMLR}, \bibinfo{pages}{3620--3629}.
\newblock
\urldef\tempurl%
\url{https://proceedings.mlr.press/v139/garreau21a.html}
\showURL{%
\tempurl}


\bibitem[\protect\citeauthoryear{Garreau and von Luxburg}{Garreau and von Luxburg}{2020}]%
        {pmlr-v108-garreau20a}
\bibfield{author}{\bibinfo{person}{Damien Garreau} {and} \bibinfo{person}{Ulrike von Luxburg}.} \bibinfo{year}{2020}\natexlab{}.
\newblock \showarticletitle{Explaining the Explainer: A First Theoretical Analysis of LIME}. In \bibinfo{booktitle}{\emph{Proceedings of the Twenty Third International Conference on Artificial Intelligence and Statistics}} \emph{(\bibinfo{series}{Proceedings of Machine Learning Research}, Vol.~\bibinfo{volume}{108})}, \bibfield{editor}{\bibinfo{person}{Silvia Chiappa} {and} \bibinfo{person}{Roberto Calandra}} (Eds.). \bibinfo{publisher}{PMLR}, \bibinfo{pages}{1287--1296}.
\newblock
\urldef\tempurl%
\url{https://proceedings.mlr.press/v108/garreau20a.html}
\showURL{%
\tempurl}


\bibitem[\protect\citeauthoryear{Ge, Xiong, Tenorio, and From}{Ge et~al\mbox{.}}{2019}]%
        {8863343}
\bibfield{author}{\bibinfo{person}{Yuanyue Ge}, \bibinfo{person}{Ya Xiong}, \bibinfo{person}{Gabriel~Lins Tenorio}, {and} \bibinfo{person}{Pål~Johan From}.} \bibinfo{year}{2019}\natexlab{}.
\newblock \showarticletitle{Fruit Localization and Environment Perception for Strawberry Harvesting Robots}.
\newblock \bibinfo{journal}{\emph{IEEE Access}}  \bibinfo{volume}{7} (\bibinfo{year}{2019}), \bibinfo{pages}{147642--147652}.
\newblock
\urldef\tempurl%
\url{https://doi.org/10.1109/ACCESS.2019.2946369}
\showDOI{\tempurl}


\bibitem[\protect\citeauthoryear{Gen{\'e}-Mola, Sanz-Cortiella, Rosell-Polo, Morros, Ruiz-Hidalgo, Vilaplana, and Gregorio}{Gen{\'e}-Mola et~al\mbox{.}}{2020}]%
        {gene2020fruit}
\bibfield{author}{\bibinfo{person}{Jordi Gen{\'e}-Mola}, \bibinfo{person}{Ricardo Sanz-Cortiella}, \bibinfo{person}{Joan~R Rosell-Polo}, \bibinfo{person}{Josep-Ramon Morros}, \bibinfo{person}{Javier Ruiz-Hidalgo}, \bibinfo{person}{Ver{\'o}nica Vilaplana}, {and} \bibinfo{person}{Eduard Gregorio}.} \bibinfo{year}{2020}\natexlab{}.
\newblock \showarticletitle{Fruit detection and 3D location using instance segmentation neural networks and structure-from-motion photogrammetry}.
\newblock \bibinfo{journal}{\emph{Computers and Electronics in Agriculture}}  \bibinfo{volume}{169} (\bibinfo{year}{2020}), \bibinfo{pages}{105165}.
\newblock


\bibitem[\protect\citeauthoryear{Geus, Meletis, Lu, Wen, and Dubbelman}{Geus et~al\mbox{.}}{2021}]%
        {Geus2021}
\bibfield{author}{\bibinfo{person}{Daan~de Geus}, \bibinfo{person}{Panagiotis Meletis}, \bibinfo{person}{Chenyang Lu}, \bibinfo{person}{Xiaoxiao Wen}, {and} \bibinfo{person}{Gijs Dubbelman}.} \bibinfo{year}{2021}\natexlab{}.
\newblock \showarticletitle{Part-aware Panoptic Segmentation}. In \bibinfo{booktitle}{\emph{2021 IEEE/CVF Conference on Computer Vision and Pattern Recognition (CVPR)}}. \bibinfo{pages}{5481--5490}.
\newblock
\urldef\tempurl%
\url{https://doi.org/10.1109/CVPR46437.2021.00544}
\showDOI{\tempurl}


\bibitem[\protect\citeauthoryear{Girshick}{Girshick}{2015}]%
        {FastRCNN}
\bibfield{author}{\bibinfo{person}{Ross Girshick}.} \bibinfo{year}{2015}\natexlab{}.
\newblock \showarticletitle{Fast R-CNN}. In \bibinfo{booktitle}{\emph{ICCV}}.
\newblock


\bibitem[\protect\citeauthoryear{Gomes and Leta}{Gomes and Leta}{2012}]%
        {gomes2012applications}
\bibfield{author}{\bibinfo{person}{Juliana Freitas~Santos Gomes} {and} \bibinfo{person}{Fabiana~Rodrigues Leta}.} \bibinfo{year}{2012}\natexlab{}.
\newblock \showarticletitle{Applications of computer vision techniques in the agriculture and food industry: a review}.
\newblock \bibinfo{journal}{\emph{European Food Research and Technology}}  \bibinfo{volume}{235} (\bibinfo{year}{2012}), \bibinfo{pages}{989--1000}.
\newblock


\bibitem[\protect\citeauthoryear{Gomez-Zavaglia, Mejuto, and Simal-Gandara}{Gomez-Zavaglia et~al\mbox{.}}{2020}]%
        {gomez2020mitigation}
\bibfield{author}{\bibinfo{person}{Andrea Gomez-Zavaglia}, \bibinfo{person}{Juan~Carlos Mejuto}, {and} \bibinfo{person}{Jesus Simal-Gandara}.} \bibinfo{year}{2020}\natexlab{}.
\newblock \showarticletitle{Mitigation of emerging implications of climate change on food production systems}.
\newblock \bibinfo{journal}{\emph{Food Research International}}  \bibinfo{volume}{134} (\bibinfo{year}{2020}), \bibinfo{pages}{109256}.
\newblock


\bibitem[\protect\citeauthoryear{Gould, Fulton, and Koller}{Gould et~al\mbox{.}}{2009}]%
        {Gould2009}
\bibfield{author}{\bibinfo{person}{Stephen Gould}, \bibinfo{person}{Richard Fulton}, {and} \bibinfo{person}{Daphne Koller}.} \bibinfo{year}{2009}\natexlab{}.
\newblock \showarticletitle{Decomposing a scene into geometric and semantically consistent regions}. In \bibinfo{booktitle}{\emph{2009 IEEE 12th International Conference on Computer Vision}}. \bibinfo{pages}{1--8}.
\newblock
\urldef\tempurl%
\url{https://doi.org/10.1109/ICCV.2009.5459211}
\showDOI{\tempurl}


\bibitem[\protect\citeauthoryear{Goyal, Dollár, Girshick, Noordhuis, Wesolowski, Kyrola, Tulloch, Jia, and He}{Goyal et~al\mbox{.}}{2018}]%
        {goyal2018accurate}
\bibfield{author}{\bibinfo{person}{Priya Goyal}, \bibinfo{person}{Piotr Dollár}, \bibinfo{person}{Ross Girshick}, \bibinfo{person}{Pieter Noordhuis}, \bibinfo{person}{Lukasz Wesolowski}, \bibinfo{person}{Aapo Kyrola}, \bibinfo{person}{Andrew Tulloch}, \bibinfo{person}{Yangqing Jia}, {and} \bibinfo{person}{Kaiming He}.} \bibinfo{year}{2018}\natexlab{}.
\newblock \showarticletitle{Accurate, Large Minibatch SGD: Training ImageNet in 1 Hour}.
\newblock \bibinfo{journal}{\emph{arXiv 1706.02677}} (\bibinfo{year}{2018}).
\newblock


\bibitem[\protect\citeauthoryear{Goyal, Wu, Ernst, Batra, Parikh, and Lee}{Goyal et~al\mbox{.}}{2019}]%
        {goyal2019counterfactual}
\bibfield{author}{\bibinfo{person}{Yash Goyal}, \bibinfo{person}{Ziyan Wu}, \bibinfo{person}{Jan Ernst}, \bibinfo{person}{Dhruv Batra}, \bibinfo{person}{Devi Parikh}, {and} \bibinfo{person}{Stefan Lee}.} \bibinfo{year}{2019}\natexlab{}.
\newblock \showarticletitle{Counterfactual visual explanations}. In \bibinfo{booktitle}{\emph{International Conference on Machine Learning}}. PMLR, \bibinfo{pages}{2376--2384}.
\newblock


\bibitem[\protect\citeauthoryear{Gozzovelli, Franchetti, Bekmurat, and Pirri}{Gozzovelli et~al\mbox{.}}{2021}]%
        {gozzovelli2021tip}
\bibfield{author}{\bibinfo{person}{Riccardo Gozzovelli}, \bibinfo{person}{Benjamin Franchetti}, \bibinfo{person}{Malik Bekmurat}, {and} \bibinfo{person}{Fiora Pirri}.} \bibinfo{year}{2021}\natexlab{}.
\newblock \showarticletitle{Tip-burn stress detection of lettuce canopy grown in Plant Factories}. In \bibinfo{booktitle}{\emph{Proceedings of the IEEE/CVF International Conference on Computer Vision}}. \bibinfo{pages}{1259--1268}.
\newblock


\bibitem[\protect\citeauthoryear{Guo, Pleiss, Sun, and Weinberger}{Guo et~al\mbox{.}}{2017}]%
        {Guo-Weinberger-2017:Calibration}
\bibfield{author}{\bibinfo{person}{Chuan Guo}, \bibinfo{person}{Geoff Pleiss}, \bibinfo{person}{Yu Sun}, {and} \bibinfo{person}{Kilian~Q. Weinberger}.} \bibinfo{year}{2017}\natexlab{}.
\newblock \showarticletitle{On Calibration of Modern Neural Networks}. In \bibinfo{booktitle}{\emph{Proceedings of the 34th International Conference on Machine Learning - Volume 70}} (Sydney, NSW, Australia) \emph{(\bibinfo{series}{ICML'17})}. \bibinfo{publisher}{JMLR.org}, \bibinfo{pages}{1321–1330}.
\newblock


\bibitem[\protect\citeauthoryear{Guo, Li, Yu, and Miao}{Guo et~al\mbox{.}}{2021a}]%
        {XuGuo-2021}
\bibfield{author}{\bibinfo{person}{Xu Guo}, \bibinfo{person}{Boyang Li}, \bibinfo{person}{Han Yu}, {and} \bibinfo{person}{Chunyan Miao}.} \bibinfo{year}{2021}\natexlab{a}.
\newblock \showarticletitle{Latent-Optimized Adversarial Neural Transfer for Sarcasm Detection}. In \bibinfo{booktitle}{\emph{2021 Annual Conference of the North American Chapter of the Association for Computational Linguistics (NAACL-HLT 2021)}}.
\newblock


\bibitem[\protect\citeauthoryear{Guo, Li, Yu, and Miao}{Guo et~al\mbox{.}}{2021b}]%
        {XuGuo-BoyangLi-NAACL-2021}
\bibfield{author}{\bibinfo{person}{Xu Guo}, \bibinfo{person}{Boyang Li}, \bibinfo{person}{Han Yu}, {and} \bibinfo{person}{Chunyan Miao}.} \bibinfo{year}{2021}\natexlab{b}.
\newblock \showarticletitle{Latent-Optimized Adversarial Neural Transfer for Sarcasm Detection}. In \bibinfo{booktitle}{\emph{2021 Annual Conference of the North American Chapter of the Association for Computational Linguistics (NAACL-HLT 2021)}}.
\newblock
\urldef\tempurl%
\url{http://www.boyangli.org/paper/XuGuo-NAACL-2021.pdf}
\showURL{%
\tempurl}


\bibitem[\protect\citeauthoryear{Gupta, Kamath, Kembhavi, and Hoiem}{Gupta et~al\mbox{.}}{2022}]%
        {gupta2022towards}
\bibfield{author}{\bibinfo{person}{Tanmay Gupta}, \bibinfo{person}{Amita Kamath}, \bibinfo{person}{Aniruddha Kembhavi}, {and} \bibinfo{person}{Derek Hoiem}.} \bibinfo{year}{2022}\natexlab{}.
\newblock \showarticletitle{Towards General Purpose Vision Systems: An End-to-End Task-Agnostic Vision-Language Architecture}. In \bibinfo{booktitle}{\emph{Proceedings of the IEEE/CVF Conference on Computer Vision and Pattern Recognition}}. \bibinfo{pages}{16399--16409}.
\newblock


\bibitem[\protect\citeauthoryear{Gustafson and Stoldt}{Gustafson and Stoldt}{1936}]%
        {gustafson1936some}
\bibfield{author}{\bibinfo{person}{Felix~G Gustafson} {and} \bibinfo{person}{Elnore Stoldt}.} \bibinfo{year}{1936}\natexlab{}.
\newblock \showarticletitle{Some relations between leaf area and fruit size in tomatoes}.
\newblock \bibinfo{journal}{\emph{Plant Physiology}} \bibinfo{volume}{11}, \bibinfo{number}{2} (\bibinfo{year}{1936}), \bibinfo{pages}{445}.
\newblock


\bibitem[\protect\citeauthoryear{Guti{\'e}rrez, Wendel, and Underwood}{Guti{\'e}rrez et~al\mbox{.}}{2019}]%
        {gutierrez2019ground}
\bibfield{author}{\bibinfo{person}{Salvador Guti{\'e}rrez}, \bibinfo{person}{Alexander Wendel}, {and} \bibinfo{person}{James Underwood}.} \bibinfo{year}{2019}\natexlab{}.
\newblock \showarticletitle{Ground based hyperspectral imaging for extensive mango yield estimation}.
\newblock \bibinfo{journal}{\emph{Computers and Electronics in Agriculture}}  \bibinfo{volume}{157} (\bibinfo{year}{2019}), \bibinfo{pages}{126--135}.
\newblock


\bibitem[\protect\citeauthoryear{Habib, Arif, Shorif, Uddin, and Ahmed}{Habib et~al\mbox{.}}{2021}]%
        {habib2021machine}
\bibfield{author}{\bibinfo{person}{Md~Tarek Habib}, \bibinfo{person}{Md~Ariful~Islam Arif}, \bibinfo{person}{Sumaita~Binte Shorif}, \bibinfo{person}{Mohammad~Shorif Uddin}, {and} \bibinfo{person}{Farruk Ahmed}.} \bibinfo{year}{2021}\natexlab{}.
\newblock \showarticletitle{Machine Vision-Based Fruit and Vegetable Disease Recognition: A Review}.
\newblock \bibinfo{journal}{\emph{Computer Vision and Machine Learning in Agriculture}} (\bibinfo{year}{2021}), \bibinfo{pages}{143--157}.
\newblock


\bibitem[\protect\citeauthoryear{Halstead, McCool, Denman, Perez, and Fookes}{Halstead et~al\mbox{.}}{2018}]%
        {halstead2018fruit}
\bibfield{author}{\bibinfo{person}{Michael Halstead}, \bibinfo{person}{Christopher McCool}, \bibinfo{person}{Simon Denman}, \bibinfo{person}{Tristan Perez}, {and} \bibinfo{person}{Clinton Fookes}.} \bibinfo{year}{2018}\natexlab{}.
\newblock \showarticletitle{Fruit quantity and ripeness estimation using a robotic vision system}.
\newblock \bibinfo{journal}{\emph{IEEE robotics and automation LETTERS}} \bibinfo{volume}{3}, \bibinfo{number}{4} (\bibinfo{year}{2018}), \bibinfo{pages}{2995--3002}.
\newblock


\bibitem[\protect\citeauthoryear{Han, Xie, and Zisserman}{Han et~al\mbox{.}}{2020}]%
        {han2020self}
\bibfield{author}{\bibinfo{person}{Tengda Han}, \bibinfo{person}{Weidi Xie}, {and} \bibinfo{person}{Andrew Zisserman}.} \bibinfo{year}{2020}\natexlab{}.
\newblock \showarticletitle{Self-supervised co-training for video representation learning}.
\newblock \bibinfo{journal}{\emph{Advances in Neural Information Processing Systems}}  \bibinfo{volume}{33} (\bibinfo{year}{2020}), \bibinfo{pages}{5679--5690}.
\newblock


\bibitem[\protect\citeauthoryear{H{\"a}ni, Roy, and Isler}{H{\"a}ni et~al\mbox{.}}{2020}]%
        {hani2020minneapple}
\bibfield{author}{\bibinfo{person}{Nicolai H{\"a}ni}, \bibinfo{person}{Pravakar Roy}, {and} \bibinfo{person}{Volkan Isler}.} \bibinfo{year}{2020}\natexlab{}.
\newblock \showarticletitle{MinneApple: a benchmark dataset for apple detection and segmentation}.
\newblock \bibinfo{journal}{\emph{IEEE Robotics and Automation Letters}} \bibinfo{volume}{5}, \bibinfo{number}{2} (\bibinfo{year}{2020}), \bibinfo{pages}{852--858}.
\newblock


\bibitem[\protect\citeauthoryear{Hao, Guo, Zheng, Celeste, Kholsa, and Chen}{Hao et~al\mbox{.}}{2015}]%
        {hao2015response}
\bibfield{author}{\bibinfo{person}{X Hao}, \bibinfo{person}{X Guo}, \bibinfo{person}{J Zheng}, \bibinfo{person}{L Celeste}, \bibinfo{person}{S Kholsa}, {and} \bibinfo{person}{X Chen}.} \bibinfo{year}{2015}\natexlab{}.
\newblock \showarticletitle{Response of greenhouse tomato to different vertical spectra of LED lighting under overhead high pressure sodium and plasma lighting}. In \bibinfo{booktitle}{\emph{International Symposium on New Technologies and Management for Greenhouses-GreenSys2015 1170}}. \bibinfo{pages}{1003--1110}.
\newblock


\bibitem[\protect\citeauthoryear{Hao and Papadopoulos}{Hao and Papadopoulos}{1999}]%
        {hao1999effects}
\bibfield{author}{\bibinfo{person}{Xiuming Hao} {and} \bibinfo{person}{Athanasios~P Papadopoulos}.} \bibinfo{year}{1999}\natexlab{}.
\newblock \showarticletitle{Effects of supplemental lighting and cover materials on growth, photosynthesis, biomass partitioning, early yield and quality of greenhouse cucumber}.
\newblock \bibinfo{journal}{\emph{Scientia Horticulturae}} \bibinfo{volume}{80}, \bibinfo{number}{1-2} (\bibinfo{year}{1999}), \bibinfo{pages}{1--18}.
\newblock


\bibitem[\protect\citeauthoryear{Hariharan, Arbel{\'a}ez, Girshick, and Malik}{Hariharan et~al\mbox{.}}{2014}]%
        {Hariharan2014}
\bibfield{author}{\bibinfo{person}{Bharath Hariharan}, \bibinfo{person}{Pablo Arbel{\'a}ez}, \bibinfo{person}{Ross Girshick}, {and} \bibinfo{person}{Jitendra Malik}.} \bibinfo{year}{2014}\natexlab{}.
\newblock \showarticletitle{Simultaneous Detection and Segmentation}. In \bibinfo{booktitle}{\emph{Computer Vision -- ECCV 2014}}, \bibfield{editor}{\bibinfo{person}{David Fleet}, \bibinfo{person}{Tomas Pajdla}, \bibinfo{person}{Bernt Schiele}, {and} \bibinfo{person}{Tinne Tuytelaars}} (Eds.). \bibinfo{publisher}{Springer International Publishing}, \bibinfo{address}{Cham}, \bibinfo{pages}{297--312}.
\newblock
\showISBNx{978-3-319-10584-0}


\bibitem[\protect\citeauthoryear{Havasi, Jenatton, Fort, Liu, Snoek, Lakshminarayanan, Dai, and Tran}{Havasi et~al\mbox{.}}{2021}]%
        {havasi2021training}
\bibfield{author}{\bibinfo{person}{Marton Havasi}, \bibinfo{person}{Rodolphe Jenatton}, \bibinfo{person}{Stanislav Fort}, \bibinfo{person}{Jeremiah~Zhe Liu}, \bibinfo{person}{Jasper Snoek}, \bibinfo{person}{Balaji Lakshminarayanan}, \bibinfo{person}{Andrew~M. Dai}, {and} \bibinfo{person}{Dustin Tran}.} \bibinfo{year}{2021}\natexlab{}.
\newblock \showarticletitle{Training independent subnetworks for robust prediction}. In \bibinfo{booktitle}{\emph{ICLR}}.
\newblock


\bibitem[\protect\citeauthoryear{Hayder, He, and Salzmann}{Hayder et~al\mbox{.}}{2016}]%
        {Hayder2016:boundary-aware}
\bibfield{author}{\bibinfo{person}{Zeeshan Hayder}, \bibinfo{person}{Xuming He}, {and} \bibinfo{person}{Mathieu Salzmann}.} \bibinfo{year}{2016}\natexlab{}.
\newblock \showarticletitle{Boundary-aware Instance Segmentation}.
\newblock \bibinfo{journal}{\emph{arXiv Preprint 1612.03129}} (\bibinfo{year}{2016}).
\newblock
\urldef\tempurl%
\url{https://doi.org/10.48550/ARXIV.1612.03129}
\showDOI{\tempurl}


\bibitem[\protect\citeauthoryear{He, Gkioxari, Dollár, and Girshick}{He et~al\mbox{.}}{2017}]%
        {MaskRCNN2017}
\bibfield{author}{\bibinfo{person}{Kaiming He}, \bibinfo{person}{Georgia Gkioxari}, \bibinfo{person}{Piotr Dollár}, {and} \bibinfo{person}{Ross Girshick}.} \bibinfo{year}{2017}\natexlab{}.
\newblock \showarticletitle{Mask R-CNN}. In \bibinfo{booktitle}{\emph{ICCV}}.
\newblock


\bibitem[\protect\citeauthoryear{He, Gkioxari, Dollár, and Girshick}{He et~al\mbox{.}}{2018}]%
        {he2018mask}
\bibfield{author}{\bibinfo{person}{Kaiming He}, \bibinfo{person}{Georgia Gkioxari}, \bibinfo{person}{Piotr Dollár}, {and} \bibinfo{person}{Ross Girshick}.} \bibinfo{year}{2018}\natexlab{}.
\newblock \showarticletitle{Mask R-CNN}.
\newblock \bibinfo{journal}{\emph{arXiv 1703.06870}} (\bibinfo{year}{2018}).
\newblock


\bibitem[\protect\citeauthoryear{He, Zhang, Ren, and Sun}{He et~al\mbox{.}}{2015a}]%
        {he2015deep}
\bibfield{author}{\bibinfo{person}{Kaiming He}, \bibinfo{person}{Xiangyu Zhang}, \bibinfo{person}{Shaoqing Ren}, {and} \bibinfo{person}{Jian Sun}.} \bibinfo{year}{2015}\natexlab{a}.
\newblock \showarticletitle{Deep Residual Learning for Image Recognition}.
\newblock \bibinfo{journal}{\emph{arXiv 1512.03385}} (\bibinfo{year}{2015}).
\newblock


\bibitem[\protect\citeauthoryear{He, Zhang, Ren, and Sun}{He et~al\mbox{.}}{2015b}]%
        {he2015-resnet-report}
\bibfield{author}{\bibinfo{person}{Kaiming He}, \bibinfo{person}{Xiangyu Zhang}, \bibinfo{person}{Shaoqing Ren}, {and} \bibinfo{person}{Jian Sun}.} \bibinfo{year}{2015}\natexlab{b}.
\newblock \showarticletitle{Deep Residual Learning for Image Recognition}.
\newblock  (\bibinfo{year}{2015}).
\newblock
\showeprint[arxiv]{1512.03385}~[cs.CV]


\bibitem[\protect\citeauthoryear{He, Fang, Zhao, Wu, Fu, Li, Majeed, and Dhupia}{He et~al\mbox{.}}{2022}]%
        {he2022fruit}
\bibfield{author}{\bibinfo{person}{Leilei He}, \bibinfo{person}{Wentai Fang}, \bibinfo{person}{Guanao Zhao}, \bibinfo{person}{Zhenchao Wu}, \bibinfo{person}{Longsheng Fu}, \bibinfo{person}{Rui Li}, \bibinfo{person}{Yaqoob Majeed}, {and} \bibinfo{person}{Jaspreet Dhupia}.} \bibinfo{year}{2022}\natexlab{}.
\newblock \showarticletitle{Fruit yield prediction and estimation in orchards: A state-of-the-art comprehensive review for both direct and indirect methods}.
\newblock \bibinfo{journal}{\emph{Computers and Electronics in Agriculture}}  \bibinfo{volume}{195} (\bibinfo{year}{2022}), \bibinfo{pages}{106812}.
\newblock


\bibitem[\protect\citeauthoryear{Hermann and Lampinen}{Hermann and Lampinen}{2020}]%
        {hermann2020shapes}
\bibfield{author}{\bibinfo{person}{Katherine~L. Hermann} {and} \bibinfo{person}{Andrew~K. Lampinen}.} \bibinfo{year}{2020}\natexlab{}.
\newblock \showarticletitle{What shapes feature representations? Exploring datasets, architectures, and training}.
\newblock \bibinfo{journal}{\emph{arXiv2006.12433}} (\bibinfo{year}{2020}).
\newblock


\bibitem[\protect\citeauthoryear{Hongyi~Zhang}{Hongyi~Zhang}{2018}]%
        {zhang2018mixup}
\bibfield{author}{\bibinfo{person}{Yann N. Dauphin David Lopez-Paz Hongyi~Zhang, Moustapha~Cisse}.} \bibinfo{year}{2018}\natexlab{}.
\newblock \showarticletitle{mixup: Beyond Empirical Risk Minimization}. In \bibinfo{booktitle}{\emph{International Conference on Learning Representations}}.
\newblock
\urldef\tempurl%
\url{https://openreview.net/forum?id=r1Ddp1-Rb}
\showURL{%
\tempurl}


\bibitem[\protect\citeauthoryear{Howard, Zhu, Chen, Kalenichenko, Wang, Weyand, Andreetto, and Adam}{Howard et~al\mbox{.}}{2017a}]%
        {howard2017mobilenets}
\bibfield{author}{\bibinfo{person}{Andrew~G. Howard}, \bibinfo{person}{Menglong Zhu}, \bibinfo{person}{Bo Chen}, \bibinfo{person}{Dmitry Kalenichenko}, \bibinfo{person}{Weijun Wang}, \bibinfo{person}{Tobias Weyand}, \bibinfo{person}{Marco Andreetto}, {and} \bibinfo{person}{Hartwig Adam}.} \bibinfo{year}{2017}\natexlab{a}.
\newblock \showarticletitle{MobileNets: Efficient Convolutional Neural Networks for Mobile Vision Applications}.
\newblock \bibinfo{journal}{\emph{arXiv 1704.04861}} (\bibinfo{year}{2017}).
\newblock


\bibitem[\protect\citeauthoryear{Howard, Zhu, Chen, Kalenichenko, Wang, Weyand, Andreetto, and Adam}{Howard et~al\mbox{.}}{2017b}]%
        {MobileNet-2017}
\bibfield{author}{\bibinfo{person}{Andrew~G. Howard}, \bibinfo{person}{Menglong Zhu}, \bibinfo{person}{Bo Chen}, \bibinfo{person}{Dmitry Kalenichenko}, \bibinfo{person}{Weijun Wang}, \bibinfo{person}{Tobias Weyand}, \bibinfo{person}{Marco Andreetto}, {and} \bibinfo{person}{Hartwig Adam}.} \bibinfo{year}{2017}\natexlab{b}.
\newblock \showarticletitle{MobileNets: Efficient Convolutional Neural Networks for Mobile Vision Applications}.
\newblock \bibinfo{journal}{\emph{arXiv PrePrint 1704.04861}} (\bibinfo{year}{2017}).
\newblock


\bibitem[\protect\citeauthoryear{Hu, Liu, Pan, and Li}{Hu et~al\mbox{.}}{2019}]%
        {hu2019automatic}
\bibfield{author}{\bibinfo{person}{Chunhua Hu}, \bibinfo{person}{Xuan Liu}, \bibinfo{person}{Zhou Pan}, {and} \bibinfo{person}{Pingping Li}.} \bibinfo{year}{2019}\natexlab{}.
\newblock \showarticletitle{Automatic detection of single ripe tomato on plant combining faster R-CNN and intuitionistic fuzzy set}.
\newblock \bibinfo{journal}{\emph{IEEE Access}}  \bibinfo{volume}{7} (\bibinfo{year}{2019}), \bibinfo{pages}{154683--154696}.
\newblock


\bibitem[\protect\citeauthoryear{Huang, Hu, Wang, Yang, Zhang, and Shi}{Huang et~al\mbox{.}}{2019}]%
        {ani9070470}
\bibfield{author}{\bibinfo{person}{Xiaoping Huang}, \bibinfo{person}{Zelin Hu}, \bibinfo{person}{Xiaorun Wang}, \bibinfo{person}{Xuanjiang Yang}, \bibinfo{person}{Jian Zhang}, {and} \bibinfo{person}{Daoling Shi}.} \bibinfo{year}{2019}\natexlab{}.
\newblock \showarticletitle{An Improved Single Shot Multibox Detector Method Applied in Body Condition Score for Dairy Cows}.
\newblock \bibinfo{journal}{\emph{Animals}} \bibinfo{volume}{9}, \bibinfo{number}{7} (\bibinfo{year}{2019}).
\newblock
\showISSN{2076-2615}
\urldef\tempurl%
\url{https://doi.org/10.3390/ani9070470}
\showDOI{\tempurl}


\bibitem[\protect\citeauthoryear{Huang, Wang, and Basanta}{Huang et~al\mbox{.}}{2020}]%
        {huang2020using}
\bibfield{author}{\bibinfo{person}{Yo-Ping Huang}, \bibinfo{person}{Tzu-Hao Wang}, {and} \bibinfo{person}{Haobijam Basanta}.} \bibinfo{year}{2020}\natexlab{}.
\newblock \showarticletitle{Using fuzzy mask R-CNN model to automatically identify tomato ripeness}.
\newblock \bibinfo{journal}{\emph{IEEE Access}}  \bibinfo{volume}{8} (\bibinfo{year}{2020}), \bibinfo{pages}{207672--207682}.
\newblock


\bibitem[\protect\citeauthoryear{Huffman}{Huffman}{2012}]%
        {huffman2012status}
\bibfield{author}{\bibinfo{person}{Wallace~E Huffman}.} \bibinfo{year}{2012}\natexlab{}.
\newblock \showarticletitle{The status of labor-saving mechanization in US fruit and vegetable harvesting}.
\newblock \bibinfo{journal}{\emph{Choices}} \bibinfo{volume}{27}, \bibinfo{number}{316-2016-6262} (\bibinfo{year}{2012}).
\newblock


\bibitem[\protect\citeauthoryear{Hughes, Salath{\'e}, et~al\mbox{.}}{Hughes et~al\mbox{.}}{2015}]%
        {hughes2015open}
\bibfield{author}{\bibinfo{person}{David Hughes}, \bibinfo{person}{Marcel Salath{\'e}}, {et~al\mbox{.}}} \bibinfo{year}{2015}\natexlab{}.
\newblock \showarticletitle{An open access repository of images on plant health to enable the development of mobile disease diagnostics}.
\newblock \bibinfo{journal}{\emph{arXiv preprint arXiv:1511.08060}} (\bibinfo{year}{2015}).
\newblock


\bibitem[\protect\citeauthoryear{H{\"u}llermeier and Waegeman}{H{\"u}llermeier and Waegeman}{2021}]%
        {hullermeier2021aleatoric}
\bibfield{author}{\bibinfo{person}{Eyke H{\"u}llermeier} {and} \bibinfo{person}{Willem Waegeman}.} \bibinfo{year}{2021}\natexlab{}.
\newblock \showarticletitle{Aleatoric and epistemic uncertainty in machine learning: An introduction to concepts and methods}.
\newblock \bibinfo{journal}{\emph{Machine Learning}} \bibinfo{volume}{110}, \bibinfo{number}{3} (\bibinfo{year}{2021}), \bibinfo{pages}{457--506}.
\newblock


\bibitem[\protect\citeauthoryear{Ibrahimi, Sors, de~Rezende, and Clinchant}{Ibrahimi et~al\mbox{.}}{2022}]%
        {ibrahimi2022learning}
\bibfield{author}{\bibinfo{person}{Sarah Ibrahimi}, \bibinfo{person}{Arnaud Sors}, \bibinfo{person}{Rafael~Sampaio de Rezende}, {and} \bibinfo{person}{St{\'e}phane Clinchant}.} \bibinfo{year}{2022}\natexlab{}.
\newblock \showarticletitle{Learning with label noise for image retrieval by selecting interactions}. In \bibinfo{booktitle}{\emph{Proceedings of the IEEE/CVF Winter Conference on Applications of Computer Vision}}. \bibinfo{pages}{2181--2190}.
\newblock


\bibitem[\protect\citeauthoryear{Iljazi}{Iljazi}{2017}]%
        {iljazi2017deep}
\bibfield{author}{\bibinfo{person}{Joana Iljazi}.} \bibinfo{year}{2017}\natexlab{}.
\newblock \showarticletitle{Deep learning for image-based prediction of plant growth in City Farms}.
\newblock  (\bibinfo{year}{2017}).
\newblock


\bibitem[\protect\citeauthoryear{Iqbal, Khan, Sharif, Shah, ur~Rehman, and Javed}{Iqbal et~al\mbox{.}}{2018}]%
        {iqbal2018automated}
\bibfield{author}{\bibinfo{person}{Zahid Iqbal}, \bibinfo{person}{Muhammad~Attique Khan}, \bibinfo{person}{Muhammad Sharif}, \bibinfo{person}{Jamal~Hussain Shah}, \bibinfo{person}{Muhammad~Habib ur Rehman}, {and} \bibinfo{person}{Kashif Javed}.} \bibinfo{year}{2018}\natexlab{}.
\newblock \showarticletitle{An automated detection and classification of citrus plant diseases using image processing techniques: A review}.
\newblock \bibinfo{journal}{\emph{Computers and electronics in agriculture}}  \bibinfo{volume}{153} (\bibinfo{year}{2018}), \bibinfo{pages}{12--32}.
\newblock


\bibitem[\protect\citeauthoryear{Ireri, Belal, Okinda, Makange, and Ji}{Ireri et~al\mbox{.}}{2019}]%
        {ireri2019computer}
\bibfield{author}{\bibinfo{person}{David Ireri}, \bibinfo{person}{Eisa Belal}, \bibinfo{person}{Cedric Okinda}, \bibinfo{person}{Nelson Makange}, {and} \bibinfo{person}{Changying Ji}.} \bibinfo{year}{2019}\natexlab{}.
\newblock \showarticletitle{A computer vision system for defect discrimination and grading in tomatoes using machine learning and image processing}.
\newblock \bibinfo{journal}{\emph{Artificial Intelligence in Agriculture}}  \bibinfo{volume}{2} (\bibinfo{year}{2019}), \bibinfo{pages}{28--37}.
\newblock


\bibitem[\protect\citeauthoryear{Jacobsen, Smeulders, and Oyallon}{Jacobsen et~al\mbox{.}}{2018}]%
        {jacobsen2018irevnet}
\bibfield{author}{\bibinfo{person}{Jörn-Henrik Jacobsen}, \bibinfo{person}{Arnold Smeulders}, {and} \bibinfo{person}{Edouard Oyallon}.} \bibinfo{year}{2018}\natexlab{}.
\newblock \showarticletitle{i-RevNet: Deep Invertible Networks}. In \bibinfo{booktitle}{\emph{International Conference on Learning Representations (ICLR)}}.
\newblock


\bibitem[\protect\citeauthoryear{Jaegle, Borgeaud, Alayrac, Doersch, Ionescu, Ding, Koppula, Zoran, Brock, Shelhamer, et~al\mbox{.}}{Jaegle et~al\mbox{.}}{2021}]%
        {jaegle2021perceiver}
\bibfield{author}{\bibinfo{person}{Andrew Jaegle}, \bibinfo{person}{Sebastian Borgeaud}, \bibinfo{person}{Jean-Baptiste Alayrac}, \bibinfo{person}{Carl Doersch}, \bibinfo{person}{Catalin Ionescu}, \bibinfo{person}{David Ding}, \bibinfo{person}{Skanda Koppula}, \bibinfo{person}{Daniel Zoran}, \bibinfo{person}{Andrew Brock}, \bibinfo{person}{Evan Shelhamer}, {et~al\mbox{.}}} \bibinfo{year}{2021}\natexlab{}.
\newblock \showarticletitle{Perceiver io: A general architecture for structured inputs \& outputs}.
\newblock \bibinfo{journal}{\emph{arXiv preprint arXiv:2107.14795}} (\bibinfo{year}{2021}).
\newblock


\bibitem[\protect\citeauthoryear{Jiang, Di~Huang, and Yang}{Jiang et~al\mbox{.}}{2020}]%
        {jiang2020beyond}
\bibfield{author}{\bibinfo{person}{Lu Jiang}, \bibinfo{person}{Mason~Liu Di~Huang}, {and} \bibinfo{person}{Weilong Yang}.} \bibinfo{year}{2020}\natexlab{}.
\newblock \showarticletitle{Beyond synthetic noise: Deep learning on controlled noisy labels}. In \bibinfo{booktitle}{\emph{ICML}}.
\newblock


\bibitem[\protect\citeauthoryear{Kakani, Nguyen, Kumar, Kim, and Pasupuleti}{Kakani et~al\mbox{.}}{2020}]%
        {kakani2020critical}
\bibfield{author}{\bibinfo{person}{Vijay Kakani}, \bibinfo{person}{Van~Huan Nguyen}, \bibinfo{person}{Basivi~Praveen Kumar}, \bibinfo{person}{Hakil Kim}, {and} \bibinfo{person}{Visweswara~Rao Pasupuleti}.} \bibinfo{year}{2020}\natexlab{}.
\newblock \showarticletitle{A critical review on computer vision and artificial intelligence in food industry}.
\newblock \bibinfo{journal}{\emph{Journal of Agriculture and Food Research}}  \bibinfo{volume}{2} (\bibinfo{year}{2020}), \bibinfo{pages}{100033}.
\newblock


\bibitem[\protect\citeauthoryear{Kamath, Clark, Gupta, Kolve, Hoiem, and Kembhavi}{Kamath et~al\mbox{.}}{2022}]%
        {kamath2022webly}
\bibfield{author}{\bibinfo{person}{Amita Kamath}, \bibinfo{person}{Christopher Clark}, \bibinfo{person}{Tanmay Gupta}, \bibinfo{person}{Eric Kolve}, \bibinfo{person}{Derek Hoiem}, {and} \bibinfo{person}{Aniruddha Kembhavi}.} \bibinfo{year}{2022}\natexlab{}.
\newblock \showarticletitle{Webly Supervised Concept Expansion for General Purpose Vision Models}.
\newblock \bibinfo{journal}{\emph{arXiv preprint arXiv:2202.02317}} (\bibinfo{year}{2022}).
\newblock


\bibitem[\protect\citeauthoryear{Kanamori, Takagi, Kobayashi, and Arimura}{Kanamori et~al\mbox{.}}{2020}]%
        {kanamori2020dace}
\bibfield{author}{\bibinfo{person}{Kentaro Kanamori}, \bibinfo{person}{Takuya Takagi}, \bibinfo{person}{Ken Kobayashi}, {and} \bibinfo{person}{Hiroki Arimura}.} \bibinfo{year}{2020}\natexlab{}.
\newblock \showarticletitle{DACE: Distribution-Aware Counterfactual Explanation by Mixed-Integer Linear Optimization.}. In \bibinfo{booktitle}{\emph{IJCAI}}. \bibinfo{pages}{2855--2862}.
\newblock


\bibitem[\protect\citeauthoryear{Kestur, Meduri, and Narasipura}{Kestur et~al\mbox{.}}{2019}]%
        {kestur2019mangonet}
\bibfield{author}{\bibinfo{person}{Ramesh Kestur}, \bibinfo{person}{Avadesh Meduri}, {and} \bibinfo{person}{Omkar Narasipura}.} \bibinfo{year}{2019}\natexlab{}.
\newblock \showarticletitle{MangoNet: A deep semantic segmentation architecture for a method to detect and count mangoes in an open orchard}.
\newblock \bibinfo{journal}{\emph{Engineering Applications of Artificial Intelligence}}  \bibinfo{volume}{77} (\bibinfo{year}{2019}), \bibinfo{pages}{59--69}.
\newblock


\bibitem[\protect\citeauthoryear{Kirillov, He, Girshick, Rother, and Dollar}{Kirillov et~al\mbox{.}}{2019}]%
        {Kirillov_2019_CVPR}
\bibfield{author}{\bibinfo{person}{Alexander Kirillov}, \bibinfo{person}{Kaiming He}, \bibinfo{person}{Ross Girshick}, \bibinfo{person}{Carsten Rother}, {and} \bibinfo{person}{Piotr Dollar}.} \bibinfo{year}{2019}\natexlab{}.
\newblock \showarticletitle{Panoptic Segmentation}. In \bibinfo{booktitle}{\emph{Proceedings of the IEEE/CVF Conference on Computer Vision and Pattern Recognition (CVPR)}}.
\newblock


\bibitem[\protect\citeauthoryear{Koh and Liang}{Koh and Liang}{2017}]%
        {koh2017understanding}
\bibfield{author}{\bibinfo{person}{Pang~Wei Koh} {and} \bibinfo{person}{Percy Liang}.} \bibinfo{year}{2017}\natexlab{}.
\newblock \showarticletitle{Understanding black-box predictions via influence functions}. In \bibinfo{booktitle}{\emph{International conference on machine learning}}. PMLR, \bibinfo{pages}{1885--1894}.
\newblock


\bibitem[\protect\citeauthoryear{Koirala, Walsh, Wang, and McCarthy}{Koirala et~al\mbox{.}}{2019}]%
        {koirala2019deep}
\bibfield{author}{\bibinfo{person}{Anand Koirala}, \bibinfo{person}{KB Walsh}, \bibinfo{person}{Zhenglin Wang}, {and} \bibinfo{person}{C McCarthy}.} \bibinfo{year}{2019}\natexlab{}.
\newblock \showarticletitle{Deep learning for real-time fruit detection and orchard fruit load estimation: Benchmarking of ‘MangoYOLO’}.
\newblock \bibinfo{journal}{\emph{Precision Agriculture}} \bibinfo{volume}{20}, \bibinfo{number}{6} (\bibinfo{year}{2019}), \bibinfo{pages}{1107--1135}.
\newblock


\bibitem[\protect\citeauthoryear{Kong, Sun, Liu, Jiang, Li, and Shi}{Kong et~al\mbox{.}}{2020}]%
        {FoveaBox2020}
\bibfield{author}{\bibinfo{person}{Tao Kong}, \bibinfo{person}{Fuchun Sun}, \bibinfo{person}{Huaping Liu}, \bibinfo{person}{Yuning Jiang}, \bibinfo{person}{Lei Li}, {and} \bibinfo{person}{Jianbo Shi}.} \bibinfo{year}{2020}\natexlab{}.
\newblock \showarticletitle{FoveaBox: Beyound Anchor-Based Object Detection}.
\newblock \bibinfo{journal}{\emph{IEEE Transactions on Image Processing}}  \bibinfo{volume}{29} (\bibinfo{year}{2020}), \bibinfo{pages}{7389--7398}.
\newblock
\urldef\tempurl%
\url{https://doi.org/10.1109/TIP.2020.3002345}
\showDOI{\tempurl}


\bibitem[\protect\citeauthoryear{Kopsell, Sams, and Morrow}{Kopsell et~al\mbox{.}}{2015}]%
        {kopsell2015blue}
\bibfield{author}{\bibinfo{person}{Dean~A Kopsell}, \bibinfo{person}{Carl~E Sams}, {and} \bibinfo{person}{Robert~C Morrow}.} \bibinfo{year}{2015}\natexlab{}.
\newblock \showarticletitle{Blue wavelengths from LED lighting increase nutritionally important metabolites in specialty crops}.
\newblock \bibinfo{journal}{\emph{HortScience}} \bibinfo{volume}{50}, \bibinfo{number}{9} (\bibinfo{year}{2015}), \bibinfo{pages}{1285--1288}.
\newblock


\bibitem[\protect\citeauthoryear{Kovalev, Utkin, and Kasimov}{Kovalev et~al\mbox{.}}{2020}]%
        {KOVALEV2020106164}
\bibfield{author}{\bibinfo{person}{Maxim~S. Kovalev}, \bibinfo{person}{Lev~V. Utkin}, {and} \bibinfo{person}{Ernest~M. Kasimov}.} \bibinfo{year}{2020}\natexlab{}.
\newblock \showarticletitle{SurvLIME: A method for explaining machine learning survival models}.
\newblock \bibinfo{journal}{\emph{Knowledge-Based Systems}}  \bibinfo{volume}{203} (\bibinfo{year}{2020}), \bibinfo{pages}{106164}.
\newblock
\showISSN{0950-7051}
\urldef\tempurl%
\url{https://doi.org/10.1016/j.knosys.2020.106164}
\showDOI{\tempurl}


\bibitem[\protect\citeauthoryear{Krishnamurthy}{Krishnamurthy}{2014}]%
        {krishnamurthy2014vertical}
\bibfield{author}{\bibinfo{person}{R Krishnamurthy}.} \bibinfo{year}{2014}\natexlab{}.
\newblock \showarticletitle{vertical farming: Singapore’s Solution to feed the local urban Population}.
\newblock \bibinfo{journal}{\emph{Permaculture Research Institute}} (\bibinfo{year}{2014}).
\newblock


\bibitem[\protect\citeauthoryear{Krizhevsky}{Krizhevsky}{2014}]%
        {krizhevsky2014one}
\bibfield{author}{\bibinfo{person}{Alex Krizhevsky}.} \bibinfo{year}{2014}\natexlab{}.
\newblock \showarticletitle{One weird trick for parallelizing convolutional neural networks}.
\newblock \bibinfo{journal}{\emph{arXiv preprint arXiv:1404.5997}} (\bibinfo{year}{2014}).
\newblock


\bibitem[\protect\citeauthoryear{Krizhevsky, Sutskever, and Hinton}{Krizhevsky et~al\mbox{.}}{2012}]%
        {AlexNet-2012}
\bibfield{author}{\bibinfo{person}{Alex Krizhevsky}, \bibinfo{person}{Ilya Sutskever}, {and} \bibinfo{person}{Geoffrey~E. Hinton}.} \bibinfo{year}{2012}\natexlab{}.
\newblock \showarticletitle{ImageNet Classification with Deep Convolutional Neural Networks}. In \bibinfo{booktitle}{\emph{Proceedings of the 25th International Conference on Neural Information Processing Systems}} (Lake Tahoe, Nevada) \emph{(\bibinfo{series}{NIPS'12})}. \bibinfo{publisher}{Curran Associates Inc.}, \bibinfo{address}{Red Hook, NY, USA}, \bibinfo{pages}{1097–1105}.
\newblock


\bibitem[\protect\citeauthoryear{Krogh~Mortensen, Skovsen, Karstoft, and Gislum}{Krogh~Mortensen et~al\mbox{.}}{2019}]%
        {Mortensen_2019_CVPR_Workshops}
\bibfield{author}{\bibinfo{person}{Anders Krogh~Mortensen}, \bibinfo{person}{Soren Skovsen}, \bibinfo{person}{Henrik Karstoft}, {and} \bibinfo{person}{Rene Gislum}.} \bibinfo{year}{2019}\natexlab{}.
\newblock \showarticletitle{The Oil Radish Growth Dataset for Semantic Segmentation and Yield Estimation}. In \bibinfo{booktitle}{\emph{Proceedings of the IEEE/CVF Conference on Computer Vision and Pattern Recognition (CVPR) Workshops}}.
\newblock


\bibitem[\protect\citeauthoryear{Kurtulmus, Lee, and Vardar}{Kurtulmus et~al\mbox{.}}{2014}]%
        {kurtulmus2014immature}
\bibfield{author}{\bibinfo{person}{Ferhat Kurtulmus}, \bibinfo{person}{Won~Suk Lee}, {and} \bibinfo{person}{Ali Vardar}.} \bibinfo{year}{2014}\natexlab{}.
\newblock \showarticletitle{Immature peach detection in colour images acquired in natural illumination conditions using statistical classifiers and neural network}.
\newblock \bibinfo{journal}{\emph{Precision agriculture}} \bibinfo{volume}{15}, \bibinfo{number}{1} (\bibinfo{year}{2014}), \bibinfo{pages}{57--79}.
\newblock


\bibitem[\protect\citeauthoryear{Kusumam, Krajn{\'\i}k, Pearson, Duckett, and Cielniak}{Kusumam et~al\mbox{.}}{2017}]%
        {kusumam20173d}
\bibfield{author}{\bibinfo{person}{Keerthy Kusumam}, \bibinfo{person}{Tom{\'a}{\v{s}} Krajn{\'\i}k}, \bibinfo{person}{Simon Pearson}, \bibinfo{person}{Tom Duckett}, {and} \bibinfo{person}{Grzegorz Cielniak}.} \bibinfo{year}{2017}\natexlab{}.
\newblock \showarticletitle{3D-vision based detection, localization, and sizing of broccoli heads in the field}.
\newblock \bibinfo{journal}{\emph{Journal of Field Robotics}} \bibinfo{volume}{34}, \bibinfo{number}{8} (\bibinfo{year}{2017}), \bibinfo{pages}{1505--1518}.
\newblock


\bibitem[\protect\citeauthoryear{Ladický, Russell, Kohli, and Torr}{Ladický et~al\mbox{.}}{2009}]%
        {Ladicky2009}
\bibfield{author}{\bibinfo{person}{L'ubor Ladický}, \bibinfo{person}{Chris Russell}, \bibinfo{person}{Pushmeet Kohli}, {and} \bibinfo{person}{Philip~H.S. Torr}.} \bibinfo{year}{2009}\natexlab{}.
\newblock \showarticletitle{Associative hierarchical CRFs for object class image segmentation}. In \bibinfo{booktitle}{\emph{2009 IEEE 12th International Conference on Computer Vision}}. \bibinfo{pages}{739--746}.
\newblock
\urldef\tempurl%
\url{https://doi.org/10.1109/ICCV.2009.5459248}
\showDOI{\tempurl}


\bibitem[\protect\citeauthoryear{Lahat, Adali, and Jutten}{Lahat et~al\mbox{.}}{2015}]%
        {lahat2015multimodal}
\bibfield{author}{\bibinfo{person}{Dana Lahat}, \bibinfo{person}{T{\"u}lay Adali}, {and} \bibinfo{person}{Christian Jutten}.} \bibinfo{year}{2015}\natexlab{}.
\newblock \showarticletitle{Multimodal data fusion: an overview of methods, challenges, and prospects}.
\newblock \bibinfo{journal}{\emph{Proc. IEEE}} \bibinfo{volume}{103}, \bibinfo{number}{9} (\bibinfo{year}{2015}), \bibinfo{pages}{1449--1477}.
\newblock


\bibitem[\protect\citeauthoryear{Lancashire, Bleiholder, Boom, Langel{\"u}ddeke, Stauss, Weber, and Witzenberger}{Lancashire et~al\mbox{.}}{1991}]%
        {lancashire1991uniform}
\bibfield{author}{\bibinfo{person}{Peter~D Lancashire}, \bibinfo{person}{Hermann Bleiholder}, \bibinfo{person}{T~van~den Boom}, \bibinfo{person}{P Langel{\"u}ddeke}, \bibinfo{person}{Reinhold Stauss}, \bibinfo{person}{Elfriede Weber}, {and} \bibinfo{person}{A Witzenberger}.} \bibinfo{year}{1991}\natexlab{}.
\newblock \showarticletitle{A uniform decimal code for growth stages of crops and weeds}.
\newblock \bibinfo{journal}{\emph{Annals of applied Biology}} \bibinfo{volume}{119}, \bibinfo{number}{3} (\bibinfo{year}{1991}), \bibinfo{pages}{561--601}.
\newblock


\bibitem[\protect\citeauthoryear{Larsson, Maire, and Shakhnarovich}{Larsson et~al\mbox{.}}{2017}]%
        {larsson2017fractalnet}
\bibfield{author}{\bibinfo{person}{Gustav Larsson}, \bibinfo{person}{Michael Maire}, {and} \bibinfo{person}{Gregory Shakhnarovich}.} \bibinfo{year}{2017}\natexlab{}.
\newblock \showarticletitle{FractalNet: Ultra-Deep Neural Networks without Residuals}. In \bibinfo{booktitle}{\emph{ICLR}}.
\newblock


\bibitem[\protect\citeauthoryear{Law and Deng}{Law and Deng}{2018}]%
        {CornerNet2018}
\bibfield{author}{\bibinfo{person}{Hei Law} {and} \bibinfo{person}{Jia Deng}.} \bibinfo{year}{2018}\natexlab{}.
\newblock \showarticletitle{CornerNet: Detecting Objects as Paired Keypoints}. In \bibinfo{booktitle}{\emph{Proceedings of the European Conference on Computer Vision (ECCV)}}.
\newblock


\bibitem[\protect\citeauthoryear{LeCun, Bengio, and Hinton}{LeCun et~al\mbox{.}}{2015}]%
        {DLReview-2015}
\bibfield{author}{\bibinfo{person}{Yann LeCun}, \bibinfo{person}{Yoshua Bengio}, {and} \bibinfo{person}{Geoffrey Hinton}.} \bibinfo{year}{2015}\natexlab{}.
\newblock \showarticletitle{Deep learning}.
\newblock \bibinfo{journal}{\emph{Nature}}  \bibinfo{volume}{521} (\bibinfo{year}{2015}), \bibinfo{pages}{436–444}.
\newblock


\bibitem[\protect\citeauthoryear{Lecun, Bottou, Bengio, and Haffner}{Lecun et~al\mbox{.}}{1998}]%
        {LeNet-1998}
\bibfield{author}{\bibinfo{person}{Y. Lecun}, \bibinfo{person}{L. Bottou}, \bibinfo{person}{Y. Bengio}, {and} \bibinfo{person}{P. Haffner}.} \bibinfo{year}{1998}\natexlab{}.
\newblock \showarticletitle{Gradient-based learning applied to document recognition}.
\newblock \bibinfo{journal}{\emph{Proc. IEEE}} \bibinfo{volume}{86}, \bibinfo{number}{11} (\bibinfo{year}{1998}), \bibinfo{pages}{2278--2324}.
\newblock
\urldef\tempurl%
\url{https://doi.org/10.1109/5.726791}
\showDOI{\tempurl}


\bibitem[\protect\citeauthoryear{Lee, Moon, and Son}{Lee et~al\mbox{.}}{2021}]%
        {horticulturae7090284}
\bibfield{author}{\bibinfo{person}{Joon-Woo Lee}, \bibinfo{person}{Taewon Moon}, {and} \bibinfo{person}{Jung-Eek Son}.} \bibinfo{year}{2021}\natexlab{}.
\newblock \showarticletitle{Development of Growth Estimation Algorithms for Hydroponic Bell Peppers Using Recurrent Neural Networks}.
\newblock \bibinfo{journal}{\emph{Horticulturae}} \bibinfo{volume}{7}, \bibinfo{number}{9} (\bibinfo{year}{2021}).
\newblock
\showISSN{2311-7524}
\urldef\tempurl%
\url{https://doi.org/10.3390/horticulturae7090284}
\showDOI{\tempurl}


\bibitem[\protect\citeauthoryear{Lei, Barzilay, and Jaakkola}{Lei et~al\mbox{.}}{2016}]%
        {lei2016rationalizing}
\bibfield{author}{\bibinfo{person}{Tao Lei}, \bibinfo{person}{Regina Barzilay}, {and} \bibinfo{person}{Tommi Jaakkola}.} \bibinfo{year}{2016}\natexlab{}.
\newblock \showarticletitle{Rationalizing neural predictions}.
\newblock \bibinfo{journal}{\emph{arXiv preprint arXiv:1606.04155}} (\bibinfo{year}{2016}).
\newblock


\bibitem[\protect\citeauthoryear{Li, Lee, and Wang}{Li et~al\mbox{.}}{2016}]%
        {li2016immature}
\bibfield{author}{\bibinfo{person}{Han Li}, \bibinfo{person}{Won~Suk Lee}, {and} \bibinfo{person}{Ku Wang}.} \bibinfo{year}{2016}\natexlab{}.
\newblock \showarticletitle{Immature green citrus fruit detection and counting based on fast normalized cross correlation (FNCC) using natural outdoor colour images}.
\newblock \bibinfo{journal}{\emph{Precision Agriculture}} \bibinfo{volume}{17}, \bibinfo{number}{6} (\bibinfo{year}{2016}), \bibinfo{pages}{678--697}.
\newblock


\bibitem[\protect\citeauthoryear{Li, Min, and Fu}{Li et~al\mbox{.}}{2019b}]%
        {li2019rethinking}
\bibfield{author}{\bibinfo{person}{Kai Li}, \bibinfo{person}{Martin~Renqiang Min}, {and} \bibinfo{person}{Yun Fu}.} \bibinfo{year}{2019}\natexlab{b}.
\newblock \showarticletitle{Rethinking zero-shot learning: A conditional visual classification perspective}. In \bibinfo{booktitle}{\emph{Proceedings of the IEEE/CVF international conference on computer vision}}. \bibinfo{pages}{3583--3592}.
\newblock


\bibitem[\protect\citeauthoryear{Li, Chen, Zhu, Xie, Huang, Du, and Wang}{Li et~al\mbox{.}}{2019a}]%
        {LiYanwei2019}
\bibfield{author}{\bibinfo{person}{Yanwei Li}, \bibinfo{person}{Xinze Chen}, \bibinfo{person}{Zheng Zhu}, \bibinfo{person}{Lingxi Xie}, \bibinfo{person}{Guan Huang}, \bibinfo{person}{Dalong Du}, {and} \bibinfo{person}{Xingang Wang}.} \bibinfo{year}{2019}\natexlab{a}.
\newblock \showarticletitle{Attention-Guided Unified Network for Panoptic Segmentation}. In \bibinfo{booktitle}{\emph{2019 IEEE/CVF Conference on Computer Vision and Pattern Recognition (CVPR)}}. \bibinfo{pages}{7019--7028}.
\newblock
\urldef\tempurl%
\url{https://doi.org/10.1109/CVPR.2019.00719}
\showDOI{\tempurl}


\bibitem[\protect\citeauthoryear{Li and Liang}{Li and Liang}{2018}]%
        {LiYuanzhi:NEURIPS2018}
\bibfield{author}{\bibinfo{person}{Yuanzhi Li} {and} \bibinfo{person}{Yingyu Liang}.} \bibinfo{year}{2018}\natexlab{}.
\newblock \showarticletitle{Learning Overparameterized Neural Networks via Stochastic Gradient Descent on Structured Data}. In \bibinfo{booktitle}{\emph{Advances in Neural Information Processing Systems}}, \bibfield{editor}{\bibinfo{person}{S.~Bengio}, \bibinfo{person}{H.~Wallach}, \bibinfo{person}{H.~Larochelle}, \bibinfo{person}{K.~Grauman}, \bibinfo{person}{N.~Cesa-Bianchi}, {and} \bibinfo{person}{R.~Garnett}} (Eds.), Vol.~\bibinfo{volume}{31}. \bibinfo{publisher}{Curran Associates, Inc.}
\newblock
\urldef\tempurl%
\url{https://proceedings.neurips.cc/paper/2018/file/54fe976ba170c19ebae453679b362263-Paper.pdf}
\showURL{%
\tempurl}


\bibitem[\protect\citeauthoryear{Li, Qi, Dai, Ji, and Wei}{Li et~al\mbox{.}}{[n.\,d.]}]%
        {LiYi2017}
\bibfield{author}{\bibinfo{person}{Yi Li}, \bibinfo{person}{Haozhi Qi}, \bibinfo{person}{Jifeng Dai}, \bibinfo{person}{Xiangyang Ji}, {and} \bibinfo{person}{Yichen Wei}.} \bibinfo{year}{[n.\,d.]}\natexlab{}.
\newblock \showarticletitle{Fully Convolutional Instance-Aware Semantic Segmentation}. In \bibinfo{booktitle}{\emph{2017 IEEE Conference on Computer Vision and Pattern Recognition (CVPR)}}.
\newblock


\bibitem[\protect\citeauthoryear{Li, Wang, Dang, Sadeghi-Niaraki, and Moon}{Li et~al\mbox{.}}{2020}]%
        {li2020crop}
\bibfield{author}{\bibinfo{person}{Yanfen Li}, \bibinfo{person}{Hanxiang Wang}, \bibinfo{person}{L~Minh Dang}, \bibinfo{person}{Abolghasem Sadeghi-Niaraki}, {and} \bibinfo{person}{Hyeonjoon Moon}.} \bibinfo{year}{2020}\natexlab{}.
\newblock \showarticletitle{Crop pest recognition in natural scenes using convolutional neural networks}.
\newblock \bibinfo{journal}{\emph{Computers and Electronics in Agriculture}}  \bibinfo{volume}{169} (\bibinfo{year}{2020}), \bibinfo{pages}{105174}.
\newblock


\bibitem[\protect\citeauthoryear{Li and Yang}{Li and Yang}{2021}]%
        {li2021meta}
\bibfield{author}{\bibinfo{person}{Yang Li} {and} \bibinfo{person}{Jiachen Yang}.} \bibinfo{year}{2021}\natexlab{}.
\newblock \showarticletitle{Meta-learning baselines and database for few-shot classification in agriculture}.
\newblock \bibinfo{journal}{\emph{Computers and Electronics in Agriculture}}  \bibinfo{volume}{182} (\bibinfo{year}{2021}), \bibinfo{pages}{106055}.
\newblock


\bibitem[\protect\citeauthoryear{Li and Hoiem}{Li and Hoiem}{2020}]%
        {LiZhizhong2020}
\bibfield{author}{\bibinfo{person}{Zhizhong Li} {and} \bibinfo{person}{Derek Hoiem}.} \bibinfo{year}{2020}\natexlab{}.
\newblock \showarticletitle{Improving Confidence Estimates for Unfamiliar Examples}. In \bibinfo{booktitle}{\emph{2020 IEEE/CVF Conference on Computer Vision and Pattern Recognition (CVPR)}}. \bibinfo{pages}{2683--2692}.
\newblock
\urldef\tempurl%
\url{https://doi.org/10.1109/CVPR42600.2020.00276}
\showDOI{\tempurl}


\bibitem[\protect\citeauthoryear{Lin, Tang, Zou, Xiong, and Fang}{Lin et~al\mbox{.}}{2020}]%
        {lin2020color}
\bibfield{author}{\bibinfo{person}{Guichao Lin}, \bibinfo{person}{Yunchao Tang}, \bibinfo{person}{Xiangjun Zou}, \bibinfo{person}{Juntao Xiong}, {and} \bibinfo{person}{Yamei Fang}.} \bibinfo{year}{2020}\natexlab{}.
\newblock \showarticletitle{Color-, depth-, and shape-based 3D fruit detection}.
\newblock \bibinfo{journal}{\emph{Precision Agriculture}} \bibinfo{volume}{21}, \bibinfo{number}{1} (\bibinfo{year}{2020}), \bibinfo{pages}{1--17}.
\newblock


\bibitem[\protect\citeauthoryear{Lin, Tang, Zou, Xiong, and Li}{Lin et~al\mbox{.}}{2019}]%
        {lin2019guava}
\bibfield{author}{\bibinfo{person}{Guichao Lin}, \bibinfo{person}{Yunchao Tang}, \bibinfo{person}{Xiangjun Zou}, \bibinfo{person}{Juntao Xiong}, {and} \bibinfo{person}{Jinhui Li}.} \bibinfo{year}{2019}\natexlab{}.
\newblock \showarticletitle{Guava detection and pose estimation using a low-cost RGB-D sensor in the field}.
\newblock \bibinfo{journal}{\emph{Sensors}} \bibinfo{volume}{19}, \bibinfo{number}{2} (\bibinfo{year}{2019}), \bibinfo{pages}{428}.
\newblock


\bibitem[\protect\citeauthoryear{Lin, Dollar, Girshick, He, Hariharan, and Belongie}{Lin et~al\mbox{.}}{2017a}]%
        {Lin_2017_CVPR}
\bibfield{author}{\bibinfo{person}{Tsung-Yi Lin}, \bibinfo{person}{Piotr Dollar}, \bibinfo{person}{Ross Girshick}, \bibinfo{person}{Kaiming He}, \bibinfo{person}{Bharath Hariharan}, {and} \bibinfo{person}{Serge Belongie}.} \bibinfo{year}{2017}\natexlab{a}.
\newblock \showarticletitle{Feature Pyramid Networks for Object Detection}. In \bibinfo{booktitle}{\emph{Proceedings of the IEEE Conference on Computer Vision and Pattern Recognition (CVPR)}}.
\newblock


\bibitem[\protect\citeauthoryear{Lin, Goyal, Girshick, He, and Dollar}{Lin et~al\mbox{.}}{2017b}]%
        {RetinaNet2017}
\bibfield{author}{\bibinfo{person}{Tsung-Yi Lin}, \bibinfo{person}{Priya Goyal}, \bibinfo{person}{Ross Girshick}, \bibinfo{person}{Kaiming He}, {and} \bibinfo{person}{Piotr Dollar}.} \bibinfo{year}{2017}\natexlab{b}.
\newblock \showarticletitle{Focal Loss for Dense Object Detection}. In \bibinfo{booktitle}{\emph{Proceedings of the IEEE International Conference on Computer Vision (ICCV)}}.
\newblock


\bibitem[\protect\citeauthoryear{Liu, Yu, Li, Shen, Gao, Ren, Xie, Cui, and Miao}{Liu et~al\mbox{.}}{2021b}]%
        {ChangLiu-BoyangLi-PRISM-2021}
\bibfield{author}{\bibinfo{person}{Chang Liu}, \bibinfo{person}{Han Yu}, \bibinfo{person}{Boyang Li}, \bibinfo{person}{Zhiqi Shen}, \bibinfo{person}{Zhanning Gao}, \bibinfo{person}{Peiran Ren}, \bibinfo{person}{Xuansong Xie}, \bibinfo{person}{Lizhen Cui}, {and} \bibinfo{person}{Chunyan Miao}.} \bibinfo{year}{2021}\natexlab{b}.
\newblock \showarticletitle{Noise-resistant Deep Metric Learning with Ranking-based Instance Selection}. In \bibinfo{booktitle}{\emph{The IEEE/CVF Conference on Computer Vision and Pattern Recognition (CVPR)}}.
\newblock
\urldef\tempurl%
\url{http://www.boyangli.org/paper/ChangLiu-CVPR-2021.pdf}
\showURL{%
\tempurl}


\bibitem[\protect\citeauthoryear{Liu, Peng, Yu, Wang, Liu, Yu, and Jiang}{Liu et~al\mbox{.}}{2019a}]%
        {LiuHuanyu2019}
\bibfield{author}{\bibinfo{person}{Huanyu Liu}, \bibinfo{person}{Chao Peng}, \bibinfo{person}{Changqian Yu}, \bibinfo{person}{Jingbo Wang}, \bibinfo{person}{Xu Liu}, \bibinfo{person}{Gang Yu}, {and} \bibinfo{person}{Wei Jiang}.} \bibinfo{year}{2019}\natexlab{a}.
\newblock \showarticletitle{An End-To-End Network for Panoptic Segmentation}. In \bibinfo{booktitle}{\emph{2019 IEEE/CVF Conference on Computer Vision and Pattern Recognition (CVPR)}}. \bibinfo{pages}{6165--6174}.
\newblock
\urldef\tempurl%
\url{https://doi.org/10.1109/CVPR.2019.00633}
\showDOI{\tempurl}


\bibitem[\protect\citeauthoryear{Liu, Sun, Han, Dou, and Li}{Liu et~al\mbox{.}}{2020}]%
        {liu2020deep}
\bibfield{author}{\bibinfo{person}{Jialun Liu}, \bibinfo{person}{Yifan Sun}, \bibinfo{person}{Chuchu Han}, \bibinfo{person}{Zhaopeng Dou}, {and} \bibinfo{person}{Wenhui Li}.} \bibinfo{year}{2020}\natexlab{}.
\newblock \showarticletitle{Deep Representation Learning on Long-tailed Data: A Learnable Embedding Augmentation Perspective}. In \bibinfo{booktitle}{\emph{CVPR}}. \bibinfo{pages}{2970--2979}.
\newblock


\bibitem[\protect\citeauthoryear{Liu, Liao, and Carin}{Liu et~al\mbox{.}}{2007}]%
        {LiuQiuhua2007}
\bibfield{author}{\bibinfo{person}{Qiuhua Liu}, \bibinfo{person}{Xuejun Liao}, {and} \bibinfo{person}{Lawrence Carin}.} \bibinfo{year}{2007}\natexlab{}.
\newblock \showarticletitle{Semi-Supervised Multitask Learning}. In \bibinfo{booktitle}{\emph{Advances in Neural Information Processing Systems}}, \bibfield{editor}{\bibinfo{person}{J.~Platt}, \bibinfo{person}{D.~Koller}, \bibinfo{person}{Y.~Singer}, {and} \bibinfo{person}{S.~Roweis}} (Eds.), Vol.~\bibinfo{volume}{20}. \bibinfo{publisher}{Curran Associates, Inc.}
\newblock
\urldef\tempurl%
\url{https://proceedings.neurips.cc/paper/2007/file/a34bacf839b923770b2c360eefa26748-Paper.pdf}
\showURL{%
\tempurl}


\bibitem[\protect\citeauthoryear{Liu, Ehsani, Toudeshki, Zou, and Wang}{Liu et~al\mbox{.}}{2018}]%
        {liu2018detection}
\bibfield{author}{\bibinfo{person}{Tian-Hu Liu}, \bibinfo{person}{Reza Ehsani}, \bibinfo{person}{Arash Toudeshki}, \bibinfo{person}{Xiang-Jun Zou}, {and} \bibinfo{person}{Hong-Jun Wang}.} \bibinfo{year}{2018}\natexlab{}.
\newblock \showarticletitle{Detection of citrus fruit and tree trunks in natural environments using a multi-elliptical boundary model}.
\newblock \bibinfo{journal}{\emph{Computers in Industry}}  \bibinfo{volume}{99} (\bibinfo{year}{2018}), \bibinfo{pages}{9--16}.
\newblock


\bibitem[\protect\citeauthoryear{Liu, Anguelov, Erhan, Szegedy, Reed, Fu, and Berg}{Liu et~al\mbox{.}}{2016}]%
        {SSD2016}
\bibfield{author}{\bibinfo{person}{Wei Liu}, \bibinfo{person}{Dragomir Anguelov}, \bibinfo{person}{Dumitru Erhan}, \bibinfo{person}{Christian Szegedy}, \bibinfo{person}{Scott Reed}, \bibinfo{person}{Cheng-Yang Fu}, {and} \bibinfo{person}{Alexander~C. Berg}.} \bibinfo{year}{2016}\natexlab{}.
\newblock \showarticletitle{SSD: Single Shot MultiBox Detector}. In \bibinfo{booktitle}{\emph{Computer Vision -- ECCV 2016}}, \bibfield{editor}{\bibinfo{person}{Bastian Leibe}, \bibinfo{person}{Jiri Matas}, \bibinfo{person}{Nicu Sebe}, {and} \bibinfo{person}{Max Welling}} (Eds.). \bibinfo{pages}{21--37}.
\newblock


\bibitem[\protect\citeauthoryear{Liu, {Wang}, He, Wang, Joshi, and Xu}{Liu et~al\mbox{.}}{2019b}]%
        {10.3389}
\bibfield{author}{\bibinfo{person}{Yang Liu}, \bibinfo{person}{Duolin {Wang}}, \bibinfo{person}{Fei He}, \bibinfo{person}{Juexin Wang}, \bibinfo{person}{Trupti Joshi}, {and} \bibinfo{person}{Dong Xu}.} \bibinfo{year}{2019}\natexlab{b}.
\newblock \showarticletitle{Phenotype Prediction and Genome-Wide Association Study Using Deep Convolutional Neural Network of Soybean}.
\newblock \bibinfo{journal}{\emph{Frontiers in Genetics}}  \bibinfo{volume}{10} (\bibinfo{year}{2019}), \bibinfo{pages}{1091}.
\newblock
\showISSN{1664-8021}
\urldef\tempurl%
\url{https://doi.org/10.3389/fgene.2019.01091}
\showDOI{\tempurl}


\bibitem[\protect\citeauthoryear{Liu, Lin, Cao, Hu, Wei, Zhang, Lin, and Guo}{Liu et~al\mbox{.}}{2021a}]%
        {liu2021:SwinTransformer}
\bibfield{author}{\bibinfo{person}{Ze Liu}, \bibinfo{person}{Yutong Lin}, \bibinfo{person}{Yue Cao}, \bibinfo{person}{Han Hu}, \bibinfo{person}{Yixuan Wei}, \bibinfo{person}{Zheng Zhang}, \bibinfo{person}{Stephen Lin}, {and} \bibinfo{person}{Baining Guo}.} \bibinfo{year}{2021}\natexlab{a}.
\newblock \showarticletitle{Swin transformer: Hierarchical vision transformer using shifted windows}. In \bibinfo{booktitle}{\emph{Proceedings of the IEEE/CVF International Conference on Computer Vision}}. \bibinfo{pages}{10012--10022}.
\newblock


\bibitem[\protect\citeauthoryear{Long, Shelhamer, and Darrell}{Long et~al\mbox{.}}{2015}]%
        {long2015FCN}
\bibfield{author}{\bibinfo{person}{Jonathan Long}, \bibinfo{person}{Evan Shelhamer}, {and} \bibinfo{person}{Trevor Darrell}.} \bibinfo{year}{2015}\natexlab{}.
\newblock \showarticletitle{Fully convolutional networks for semantic segmentation}. In \bibinfo{booktitle}{\emph{Proceedings of the IEEE/CVF conference on computer vision and pattern recognition (CVPR)}}. \bibinfo{pages}{3431--3440}.
\newblock


\bibitem[\protect\citeauthoryear{Lu, Clark, Zellers, Mottaghi, and Kembhavi}{Lu et~al\mbox{.}}{2022}]%
        {lu2022unified}
\bibfield{author}{\bibinfo{person}{Jiasen Lu}, \bibinfo{person}{Christopher Clark}, \bibinfo{person}{Rowan Zellers}, \bibinfo{person}{Roozbeh Mottaghi}, {and} \bibinfo{person}{Aniruddha Kembhavi}.} \bibinfo{year}{2022}\natexlab{}.
\newblock \showarticletitle{Unified-IO: A Unified Model for Vision, Language, and Multi-Modal Tasks}.
\newblock \bibinfo{journal}{\emph{arXiv preprint arXiv:2206.08916}} (\bibinfo{year}{2022}).
\newblock


\bibitem[\protect\citeauthoryear{Lu, Chang, and Kuo}{Lu et~al\mbox{.}}{2019}]%
        {lu2019monitoring}
\bibfield{author}{\bibinfo{person}{Jie-Yan Lu}, \bibinfo{person}{Chung-Liang Chang}, {and} \bibinfo{person}{Yan-Fu Kuo}.} \bibinfo{year}{2019}\natexlab{}.
\newblock \showarticletitle{Monitoring growth rate of lettuce using deep convolutional neural networks}. In \bibinfo{booktitle}{\emph{2019 ASABE Annual International Meeting}}. American Society of Agricultural and Biological Engineers, \bibinfo{pages}{1}.
\newblock


\bibitem[\protect\citeauthoryear{Lyu, Zhao, Li, Li, Fan, and Li}{Lyu et~al\mbox{.}}{2022}]%
        {lyu2022embedded}
\bibfield{author}{\bibinfo{person}{Shilei Lyu}, \bibinfo{person}{Yawen Zhao}, \bibinfo{person}{Ruiyao Li}, \bibinfo{person}{Zhen Li}, \bibinfo{person}{Renjie Fan}, {and} \bibinfo{person}{Qiafeng Li}.} \bibinfo{year}{2022}\natexlab{}.
\newblock \showarticletitle{Embedded Sensing System for Recognizing Citrus Flowers Using Cascaded Fusion YOLOv4-CF+ FPGA}.
\newblock \bibinfo{journal}{\emph{Sensors}} \bibinfo{volume}{22}, \bibinfo{number}{3} (\bibinfo{year}{2022}), \bibinfo{pages}{1255}.
\newblock


\bibitem[\protect\citeauthoryear{Ma, Du, Zhang, Zheng, Chu, and Sun}{Ma et~al\mbox{.}}{2017}]%
        {ma2017segmentation}
\bibfield{author}{\bibinfo{person}{Juncheng Ma}, \bibinfo{person}{Keming Du}, \bibinfo{person}{Lingxian Zhang}, \bibinfo{person}{Feixiang Zheng}, \bibinfo{person}{Jinxiang Chu}, {and} \bibinfo{person}{Zhongfu Sun}.} \bibinfo{year}{2017}\natexlab{}.
\newblock \showarticletitle{A segmentation method for greenhouse vegetable foliar disease spots images using color information and region growing}.
\newblock \bibinfo{journal}{\emph{Computers and Electronics in Agriculture}}  \bibinfo{volume}{142} (\bibinfo{year}{2017}), \bibinfo{pages}{110--117}.
\newblock


\bibitem[\protect\citeauthoryear{Ma, Du, Zheng, Zhang, Gong, and Sun}{Ma et~al\mbox{.}}{2018}]%
        {ma2018recognition}
\bibfield{author}{\bibinfo{person}{Juncheng Ma}, \bibinfo{person}{Keming Du}, \bibinfo{person}{Feixiang Zheng}, \bibinfo{person}{Lingxian Zhang}, \bibinfo{person}{Zhihong Gong}, {and} \bibinfo{person}{Zhongfu Sun}.} \bibinfo{year}{2018}\natexlab{}.
\newblock \showarticletitle{A recognition method for cucumber diseases using leaf symptom images based on deep convolutional neural network}.
\newblock \bibinfo{journal}{\emph{Computers and electronics in agriculture}}  \bibinfo{volume}{154} (\bibinfo{year}{2018}), \bibinfo{pages}{18--24}.
\newblock


\bibitem[\protect\citeauthoryear{Mahendran and Vedaldi}{Mahendran and Vedaldi}{2015}]%
        {mahendran2015understanding}
\bibfield{author}{\bibinfo{person}{Aravindh Mahendran} {and} \bibinfo{person}{Andrea Vedaldi}.} \bibinfo{year}{2015}\natexlab{}.
\newblock \showarticletitle{Understanding deep image representations by inverting them}. In \bibinfo{booktitle}{\emph{Proceedings of the IEEE conference on computer vision and pattern recognition}}. \bibinfo{pages}{5188--5196}.
\newblock


\bibitem[\protect\citeauthoryear{Mehrtash, Wells, Tempany, Abolmaesumi, and Kapur}{Mehrtash et~al\mbox{.}}{2020}]%
        {Mehrtash2020:Dice-loss-calibration}
\bibfield{author}{\bibinfo{person}{Alireza Mehrtash}, \bibinfo{person}{William~M. Wells}, \bibinfo{person}{Clare~M. Tempany}, \bibinfo{person}{Purang Abolmaesumi}, {and} \bibinfo{person}{Tina Kapur}.} \bibinfo{year}{2020}\natexlab{}.
\newblock \showarticletitle{Confidence Calibration and Predictive Uncertainty Estimation for Deep Medical Image Segmentation}.
\newblock \bibinfo{journal}{\emph{IEEE Transactions on Medical Imaging}} \bibinfo{volume}{39}, \bibinfo{number}{12} (\bibinfo{year}{2020}), \bibinfo{pages}{3868--3878}.
\newblock
\urldef\tempurl%
\url{https://doi.org/10.1109/TMI.2020.3006437}
\showDOI{\tempurl}


\bibitem[\protect\citeauthoryear{Minervini, Fischbach, Scharr, and Tsaftaris}{Minervini et~al\mbox{.}}{2016}]%
        {minervini2016finely}
\bibfield{author}{\bibinfo{person}{Massimo Minervini}, \bibinfo{person}{Andreas Fischbach}, \bibinfo{person}{Hanno Scharr}, {and} \bibinfo{person}{Sotirios~A Tsaftaris}.} \bibinfo{year}{2016}\natexlab{}.
\newblock \showarticletitle{Finely-grained annotated datasets for image-based plant phenotyping}.
\newblock \bibinfo{journal}{\emph{Pattern recognition letters}}  \bibinfo{volume}{81} (\bibinfo{year}{2016}), \bibinfo{pages}{80--89}.
\newblock


\bibitem[\protect\citeauthoryear{Miranda}{Miranda}{[n.\,d.]}]%
        {Miranda2021:Data-Centric}
\bibfield{author}{\bibinfo{person}{Lj Miranda}.} \bibinfo{year}{[n.\,d.]}\natexlab{}.
\newblock \bibinfo{booktitle}{\emph{Towards data-centric machine learning: a short review}}.
\newblock
\urldef\tempurl%
\url{https://ljvmiranda921.github.io/notebook/2021/07/30/data-centric-ml/}
\showURL{%
\tempurl}


\bibitem[\protect\citeauthoryear{Molnar}{Molnar}{2020}]%
        {molnar2020interpretable}
\bibfield{author}{\bibinfo{person}{Christoph Molnar}.} \bibinfo{year}{2020}\natexlab{}.
\newblock \bibinfo{booktitle}{\emph{Interpretable machine learning}}.
\newblock
\urldef\tempurl%
\url{https://christophm.github.io/interpretable-ml-book/}
\showURL{%
\tempurl}


\bibitem[\protect\citeauthoryear{Montavon, Binder, Lapuschkin, Samek, and M{\"u}ller}{Montavon et~al\mbox{.}}{2019}]%
        {montavon2019layer}
\bibfield{author}{\bibinfo{person}{Gr{\'e}goire Montavon}, \bibinfo{person}{Alexander Binder}, \bibinfo{person}{Sebastian Lapuschkin}, \bibinfo{person}{Wojciech Samek}, {and} \bibinfo{person}{Klaus-Robert M{\"u}ller}.} \bibinfo{year}{2019}\natexlab{}.
\newblock \showarticletitle{Layer-wise relevance propagation: an overview}.
\newblock \bibinfo{journal}{\emph{Explainable AI: interpreting, explaining and visualizing deep learning}} (\bibinfo{year}{2019}), \bibinfo{pages}{193--209}.
\newblock


\bibitem[\protect\citeauthoryear{Moonrinta, Chaivivatrakul, Dailey, and Ekpanyapong}{Moonrinta et~al\mbox{.}}{2010}]%
        {moonrinta2010fruit}
\bibfield{author}{\bibinfo{person}{Jednipat Moonrinta}, \bibinfo{person}{Supawadee Chaivivatrakul}, \bibinfo{person}{Matthew~N Dailey}, {and} \bibinfo{person}{Mongkol Ekpanyapong}.} \bibinfo{year}{2010}\natexlab{}.
\newblock \showarticletitle{Fruit detection, tracking, and 3D reconstruction for crop mapping and yield estimation}. In \bibinfo{booktitle}{\emph{2010 11th International Conference on Control Automation Robotics \& Vision}}. IEEE, \bibinfo{pages}{1181--1186}.
\newblock


\bibitem[\protect\citeauthoryear{Mordvintsev, Olah, and Tyka}{Mordvintsev et~al\mbox{.}}{2015}]%
        {mordvintsev2015inceptionism}
\bibfield{author}{\bibinfo{person}{Alexander Mordvintsev}, \bibinfo{person}{Christopher Olah}, {and} \bibinfo{person}{Mike Tyka}.} \bibinfo{year}{2015}\natexlab{}.
\newblock \showarticletitle{Inceptionism: Going deeper into neural networks}.
\newblock  (\bibinfo{year}{2015}).
\newblock


\bibitem[\protect\citeauthoryear{Mukhoti, Kulharia, Sanyal, Golodetz, Torr, and Dokania}{Mukhoti et~al\mbox{.}}{2020}]%
        {Mukhoti:NEURIPS2020}
\bibfield{author}{\bibinfo{person}{Jishnu Mukhoti}, \bibinfo{person}{Viveka Kulharia}, \bibinfo{person}{Amartya Sanyal}, \bibinfo{person}{Stuart Golodetz}, \bibinfo{person}{Philip Torr}, {and} \bibinfo{person}{Puneet Dokania}.} \bibinfo{year}{2020}\natexlab{}.
\newblock \showarticletitle{Calibrating Deep Neural Networks using Focal Loss}. In \bibinfo{booktitle}{\emph{Advances in Neural Information Processing Systems}}, \bibfield{editor}{\bibinfo{person}{H.~Larochelle}, \bibinfo{person}{M.~Ranzato}, \bibinfo{person}{R.~Hadsell}, \bibinfo{person}{M.F. Balcan}, {and} \bibinfo{person}{H.~Lin}} (Eds.), Vol.~\bibinfo{volume}{33}. \bibinfo{publisher}{Curran Associates, Inc.}, \bibinfo{pages}{15288--15299}.
\newblock
\urldef\tempurl%
\url{https://proceedings.neurips.cc/paper/2020/file/aeb7b30ef1d024a76f21a1d40e30c302-Paper.pdf}
\showURL{%
\tempurl}


\bibitem[\protect\citeauthoryear{M\"{u}ller, Kornblith, and Hinton}{M\"{u}ller et~al\mbox{.}}{2019}]%
        {Muller:NEURIPS2019}
\bibfield{author}{\bibinfo{person}{Rafael M\"{u}ller}, \bibinfo{person}{Simon Kornblith}, {and} \bibinfo{person}{Geoffrey~E Hinton}.} \bibinfo{year}{2019}\natexlab{}.
\newblock \showarticletitle{When does label smoothing help?}. In \bibinfo{booktitle}{\emph{Advances in Neural Information Processing Systems}}, \bibfield{editor}{\bibinfo{person}{H.~Wallach}, \bibinfo{person}{H.~Larochelle}, \bibinfo{person}{A.~Beygelzimer}, \bibinfo{person}{F.~d\textquotesingle Alch\'{e}-Buc}, \bibinfo{person}{E.~Fox}, {and} \bibinfo{person}{R.~Garnett}} (Eds.), Vol.~\bibinfo{volume}{32}. \bibinfo{publisher}{Curran Associates, Inc.}
\newblock
\urldef\tempurl%
\url{https://proceedings.neurips.cc/paper/2019/file/f1748d6b0fd9d439f71450117eba2725-Paper.pdf}
\showURL{%
\tempurl}


\bibitem[\protect\citeauthoryear{Nassar, Khan, Villalva, Nour, Almuslem, and Hussain}{Nassar et~al\mbox{.}}{2018}]%
        {nassar2018compliant}
\bibfield{author}{\bibinfo{person}{Joanna~M Nassar}, \bibinfo{person}{Sherjeel~M Khan}, \bibinfo{person}{Diego~Rosas Villalva}, \bibinfo{person}{Maha~M Nour}, \bibinfo{person}{Amani~S Almuslem}, {and} \bibinfo{person}{Muhammad~M Hussain}.} \bibinfo{year}{2018}\natexlab{}.
\newblock \showarticletitle{Compliant plant wearables for localized microclimate and plant growth monitoring}.
\newblock \bibinfo{journal}{\emph{npj Flexible Electronics}} \bibinfo{volume}{2}, \bibinfo{number}{1} (\bibinfo{year}{2018}), \bibinfo{pages}{1--12}.
\newblock


\bibitem[\protect\citeauthoryear{Neven, Brabandere, Proesmans, and Van~Gool}{Neven et~al\mbox{.}}{2019}]%
        {Davy2019}
\bibfield{author}{\bibinfo{person}{Davy Neven}, \bibinfo{person}{Bert~De Brabandere}, \bibinfo{person}{Marc Proesmans}, {and} \bibinfo{person}{Luc Van~Gool}.} \bibinfo{year}{2019}\natexlab{}.
\newblock \showarticletitle{Instance Segmentation by Jointly Optimizing Spatial Embeddings and Clustering Bandwidth}. In \bibinfo{booktitle}{\emph{2019 IEEE/CVF Conference on Computer Vision and Pattern Recognition (CVPR)}}. \bibinfo{pages}{8829--8837}.
\newblock
\urldef\tempurl%
\url{https://doi.org/10.1109/CVPR.2019.00904}
\showDOI{\tempurl}


\bibitem[\protect\citeauthoryear{Ng}{Ng}{[n.\,d.]}]%
        {AndrewNg2021:Data-Centric}
\bibfield{author}{\bibinfo{person}{Andrew Ng}.} \bibinfo{year}{[n.\,d.]}\natexlab{}.
\newblock \bibinfo{booktitle}{\emph{A Chat with Andrew on MLOps: From Model-centric to Data-centric AI}}.
\newblock
\urldef\tempurl%
\url{https://www.youtube.com/watch?v=06-AZXmwHjo}
\showURL{%
\tempurl}


\bibitem[\protect\citeauthoryear{Nguyen, Sagan, Maimaitiyiming, Maimaitijiang, Bhadra, and Kwasniewski}{Nguyen et~al\mbox{.}}{2021}]%
        {s21030742}
\bibfield{author}{\bibinfo{person}{Canh Nguyen}, \bibinfo{person}{Vasit Sagan}, \bibinfo{person}{Matthew Maimaitiyiming}, \bibinfo{person}{Maitiniyazi Maimaitijiang}, \bibinfo{person}{Sourav Bhadra}, {and} \bibinfo{person}{Misha~T. Kwasniewski}.} \bibinfo{year}{2021}\natexlab{}.
\newblock \showarticletitle{Early Detection of Plant Viral Disease Using Hyperspectral Imaging and Deep Learning}.
\newblock \bibinfo{journal}{\emph{Sensors}} \bibinfo{volume}{21}, \bibinfo{number}{3} (\bibinfo{year}{2021}).
\newblock
\showISSN{1424-8220}
\urldef\tempurl%
\url{https://doi.org/10.3390/s21030742}
\showDOI{\tempurl}


\bibitem[\protect\citeauthoryear{Ni, Li, Jiang, and Takeda}{Ni et~al\mbox{.}}{2020}]%
        {ni2020deep}
\bibfield{author}{\bibinfo{person}{Xueping Ni}, \bibinfo{person}{Changying Li}, \bibinfo{person}{Huanyu Jiang}, {and} \bibinfo{person}{Fumiomi Takeda}.} \bibinfo{year}{2020}\natexlab{}.
\newblock \showarticletitle{Deep learning image segmentation and extraction of blueberry fruit traits associated with harvestability and yield}.
\newblock \bibinfo{journal}{\emph{Horticulture research}}  \bibinfo{volume}{7} (\bibinfo{year}{2020}).
\newblock


\bibitem[\protect\citeauthoryear{Nuthalapati and Tunga}{Nuthalapati and Tunga}{2021}]%
        {nuthalapati2021multi}
\bibfield{author}{\bibinfo{person}{Sai~Vidyaranya Nuthalapati} {and} \bibinfo{person}{Anirudh Tunga}.} \bibinfo{year}{2021}\natexlab{}.
\newblock \showarticletitle{Multi-domain few-shot learning and dataset for agricultural applications}. In \bibinfo{booktitle}{\emph{Proceedings of the IEEE/CVF International Conference on Computer Vision}}. \bibinfo{pages}{1399--1408}.
\newblock


\bibitem[\protect\citeauthoryear{Nyarko, Vidovi{\'c}, Rado{\v{c}}aj, and Cupec}{Nyarko et~al\mbox{.}}{2018}]%
        {nyarko2018nearest}
\bibfield{author}{\bibinfo{person}{Emmanuel~Karlo Nyarko}, \bibinfo{person}{Ivan Vidovi{\'c}}, \bibinfo{person}{Kristijan Rado{\v{c}}aj}, {and} \bibinfo{person}{Robert Cupec}.} \bibinfo{year}{2018}\natexlab{}.
\newblock \showarticletitle{A nearest neighbor approach for fruit recognition in RGB-D images based on detection of convex surfaces}.
\newblock \bibinfo{journal}{\emph{Expert Systems with Applications}}  \bibinfo{volume}{114} (\bibinfo{year}{2018}), \bibinfo{pages}{454--466}.
\newblock


\bibitem[\protect\citeauthoryear{Oquab, Bottou, Laptev, and Sivic}{Oquab et~al\mbox{.}}{2015}]%
        {oquab2015object}
\bibfield{author}{\bibinfo{person}{Maxime Oquab}, \bibinfo{person}{L{\'e}on Bottou}, \bibinfo{person}{Ivan Laptev}, {and} \bibinfo{person}{Josef Sivic}.} \bibinfo{year}{2015}\natexlab{}.
\newblock \showarticletitle{Is object localization for free?-weakly-supervised learning with convolutional neural networks}. In \bibinfo{booktitle}{\emph{Proceedings of the IEEE conference on computer vision and pattern recognition}}. \bibinfo{pages}{685--694}.
\newblock


\bibitem[\protect\citeauthoryear{Ostovar, Ringdahl, and Hellstr{\"o}m}{Ostovar et~al\mbox{.}}{2018}]%
        {ostovar2018adaptive}
\bibfield{author}{\bibinfo{person}{Ahmad Ostovar}, \bibinfo{person}{Ola Ringdahl}, {and} \bibinfo{person}{Thomas Hellstr{\"o}m}.} \bibinfo{year}{2018}\natexlab{}.
\newblock \showarticletitle{Adaptive image thresholding of yellow peppers for a harvesting robot}.
\newblock \bibinfo{journal}{\emph{Robotics}} \bibinfo{volume}{7}, \bibinfo{number}{1} (\bibinfo{year}{2018}), \bibinfo{pages}{11}.
\newblock


\bibitem[\protect\citeauthoryear{Otsu}{Otsu}{1979}]%
        {otsu1979threshold}
\bibfield{author}{\bibinfo{person}{Nobuyuki Otsu}.} \bibinfo{year}{1979}\natexlab{}.
\newblock \showarticletitle{A threshold selection method from gray-level histograms}.
\newblock \bibinfo{journal}{\emph{IEEE transactions on systems, man, and cybernetics}} \bibinfo{volume}{9}, \bibinfo{number}{1} (\bibinfo{year}{1979}), \bibinfo{pages}{62--66}.
\newblock


\bibitem[\protect\citeauthoryear{Payer, Štern, Neff, Bischof, and Urschler}{Payer et~al\mbox{.}}{2018}]%
        {Payer-Instance-Segmentation:2018}
\bibfield{author}{\bibinfo{person}{Christian Payer}, \bibinfo{person}{Darko Štern}, \bibinfo{person}{Thomas Neff}, \bibinfo{person}{Horst Bischof}, {and} \bibinfo{person}{Martin Urschler}.} \bibinfo{year}{2018}\natexlab{}.
\newblock \showarticletitle{Instance Segmentation and Tracking with Cosine Embeddings and Recurrent Hourglass Networks}. In \bibinfo{booktitle}{\emph{International Conference on Medical Image Computing and Computer-Assisted Intervention}}.
\newblock


\bibitem[\protect\citeauthoryear{Pedapati, Balakrishnan, Shanmugam, and Dhurandhar}{Pedapati et~al\mbox{.}}{2020}]%
        {pedapati2020learning}
\bibfield{author}{\bibinfo{person}{Tejaswini Pedapati}, \bibinfo{person}{Avinash Balakrishnan}, \bibinfo{person}{Karthikeyan Shanmugam}, {and} \bibinfo{person}{Amit Dhurandhar}.} \bibinfo{year}{2020}\natexlab{}.
\newblock \showarticletitle{Learning global transparent models consistent with local contrastive explanations}.
\newblock \bibinfo{journal}{\emph{Advances in neural information processing systems}}  \bibinfo{volume}{33} (\bibinfo{year}{2020}), \bibinfo{pages}{3592--3602}.
\newblock


\bibitem[\protect\citeauthoryear{Peng, Dallas, Ascencio-Ib{\'a}{\~n}ez, Hoyer, Legg, Hanley-Bowdoin, Grieve, and Yin}{Peng et~al\mbox{.}}{2022}]%
        {peng2022early}
\bibfield{author}{\bibinfo{person}{Yao Peng}, \bibinfo{person}{Mary~M Dallas}, \bibinfo{person}{Jos{\'e}~T Ascencio-Ib{\'a}{\~n}ez}, \bibinfo{person}{J~Steen Hoyer}, \bibinfo{person}{James Legg}, \bibinfo{person}{Linda Hanley-Bowdoin}, \bibinfo{person}{Bruce Grieve}, {and} \bibinfo{person}{Hujun Yin}.} \bibinfo{year}{2022}\natexlab{}.
\newblock \showarticletitle{Early detection of plant virus infection using multispectral imaging and spatial--spectral machine learning}.
\newblock \bibinfo{journal}{\emph{Scientific Reports}} \bibinfo{volume}{12}, \bibinfo{number}{1} (\bibinfo{year}{2022}), \bibinfo{pages}{3113}.
\newblock


\bibitem[\protect\citeauthoryear{Pereyra, Tucker, Chorowski, Łukasz Kaiser, and Hinton}{Pereyra et~al\mbox{.}}{2017}]%
        {pereyra2017regularizing}
\bibfield{author}{\bibinfo{person}{Gabriel Pereyra}, \bibinfo{person}{George Tucker}, \bibinfo{person}{Jan Chorowski}, \bibinfo{person}{Łukasz Kaiser}, {and} \bibinfo{person}{Geoffrey Hinton}.} \bibinfo{year}{2017}\natexlab{}.
\newblock \showarticletitle{Regularizing Neural Networks by Penalizing Confident Output Distributions}.
\newblock \bibinfo{journal}{\emph{arXiv 1701.06548}} (\bibinfo{year}{2017}).
\newblock


\bibitem[\protect\citeauthoryear{Pfeffer}{Pfeffer}{1900}]%
        {pfeffer1900physiology}
\bibfield{author}{\bibinfo{person}{Wilhelm Pfeffer}.} \bibinfo{year}{1900}\natexlab{}.
\newblock \bibinfo{booktitle}{\emph{The physiology of plants: a treatise upon the metabolism and sources of energy in plants}}. Vol.~\bibinfo{volume}{1}.
\newblock \bibinfo{publisher}{Clarendon Press}.
\newblock


\bibitem[\protect\citeauthoryear{Pinheiro and Collobert}{Pinheiro and Collobert}{2015}]%
        {pinheiro2015doll}
\bibfield{author}{\bibinfo{person}{Pedro~O Pinheiro} {and} \bibinfo{person}{Ronan Collobert}.} \bibinfo{year}{2015}\natexlab{}.
\newblock \showarticletitle{Doll{\'{}} ar P. Learning to segment object candidates}. In \bibinfo{booktitle}{\emph{Proc. the 28th Int. Conf. Neural Information Processing Systems}}. \bibinfo{pages}{1990--1998}.
\newblock


\bibitem[\protect\citeauthoryear{Pinheiro, Lin, Collobert, and Doll{\'a}r}{Pinheiro et~al\mbox{.}}{2016}]%
        {Pinheiro2016}
\bibfield{author}{\bibinfo{person}{Pedro~O. Pinheiro}, \bibinfo{person}{Tsung-Yi Lin}, \bibinfo{person}{Ronan Collobert}, {and} \bibinfo{person}{Piotr Doll{\'a}r}.} \bibinfo{year}{2016}\natexlab{}.
\newblock \showarticletitle{Learning to Refine Object Segments}. In \bibinfo{booktitle}{\emph{Computer Vision -- ECCV 2016}}, \bibfield{editor}{\bibinfo{person}{Bastian Leibe}, \bibinfo{person}{Jiri Matas}, \bibinfo{person}{Nicu Sebe}, {and} \bibinfo{person}{Max Welling}} (Eds.). \bibinfo{publisher}{Springer International Publishing}, \bibinfo{address}{Cham}, \bibinfo{pages}{75--91}.
\newblock
\showISBNx{978-3-319-46448-0}


\bibitem[\protect\citeauthoryear{Prusinkiewicz}{Prusinkiewicz}{2002}]%
        {prusinkiewicz2002art}
\bibfield{author}{\bibinfo{person}{Przemyslaw Prusinkiewicz}.} \bibinfo{year}{2002}\natexlab{}.
\newblock \showarticletitle{Art and science of life: designing and growing virtual plants with L-systems}. In \bibinfo{booktitle}{\emph{XXVI International Horticultural Congress: Nursery Crops; Development, Evaluation, Production and Use 630}}. \bibinfo{pages}{15--28}.
\newblock


\bibitem[\protect\citeauthoryear{Pruthi, Liu, Kale, and Sundararajan}{Pruthi et~al\mbox{.}}{2020}]%
        {pruthi2020estimating}
\bibfield{author}{\bibinfo{person}{Garima Pruthi}, \bibinfo{person}{Frederick Liu}, \bibinfo{person}{Satyen Kale}, {and} \bibinfo{person}{Mukund Sundararajan}.} \bibinfo{year}{2020}\natexlab{}.
\newblock \showarticletitle{Estimating training data influence by tracing gradient descent}.
\newblock \bibinfo{journal}{\emph{Advances in Neural Information Processing Systems}}  \bibinfo{volume}{33} (\bibinfo{year}{2020}), \bibinfo{pages}{19920--19930}.
\newblock


\bibitem[\protect\citeauthoryear{Qi, Khorram, and Fuxin}{Qi et~al\mbox{.}}{2021}]%
        {qi2021embedding}
\bibfield{author}{\bibinfo{person}{Zhongang Qi}, \bibinfo{person}{Saeed Khorram}, {and} \bibinfo{person}{Li Fuxin}.} \bibinfo{year}{2021}\natexlab{}.
\newblock \showarticletitle{Embedding deep networks into visual explanations}.
\newblock \bibinfo{journal}{\emph{Artificial Intelligence}}  \bibinfo{volume}{292} (\bibinfo{year}{2021}), \bibinfo{pages}{103435}.
\newblock


\bibitem[\protect\citeauthoryear{R~Shamshiri, Weltzien, Hameed, J~Yule, E~Grift, Balasundram, Pitonakova, Ahmad, and Chowdhary}{R~Shamshiri et~al\mbox{.}}{2018}]%
        {r2018research}
\bibfield{author}{\bibinfo{person}{Redmond R~Shamshiri}, \bibinfo{person}{Cornelia Weltzien}, \bibinfo{person}{Ibrahim~A Hameed}, \bibinfo{person}{Ian J~Yule}, \bibinfo{person}{Tony E~Grift}, \bibinfo{person}{Siva~K Balasundram}, \bibinfo{person}{Lenka Pitonakova}, \bibinfo{person}{Desa Ahmad}, {and} \bibinfo{person}{Girish Chowdhary}.} \bibinfo{year}{2018}\natexlab{}.
\newblock \showarticletitle{Research and development in agricultural robotics: A perspective of digital farming}.
\newblock  (\bibinfo{year}{2018}).
\newblock


\bibitem[\protect\citeauthoryear{Ragazou, Garefalakis, Zafeiriou, and Passas}{Ragazou et~al\mbox{.}}{2022}]%
        {ragazou2022agriculture}
\bibfield{author}{\bibinfo{person}{K Ragazou}, \bibinfo{person}{A Garefalakis}, \bibinfo{person}{E Zafeiriou}, {and} \bibinfo{person}{I Passas}.} \bibinfo{year}{2022}\natexlab{}.
\newblock \bibinfo{title}{Agriculture 5.0: A New Strategic Management Mode for a Cut Cost and an Energy Efficient Agriculture Sector. Energies 2022, 15, 3113}.
\newblock
\newblock


\bibitem[\protect\citeauthoryear{Rahimi, Islam, Duarte, Tazerji, Sobur, El~Zowalaty, Ashour, and Rahman}{Rahimi et~al\mbox{.}}{2021}]%
        {rahimi2021impact}
\bibfield{author}{\bibinfo{person}{Parastoo Rahimi}, \bibinfo{person}{Md~Saiful Islam}, \bibinfo{person}{Phelipe~Magalh{\~a}es Duarte}, \bibinfo{person}{Sina~Salajegheh Tazerji}, \bibinfo{person}{Md~Abdus Sobur}, \bibinfo{person}{Mohamed~E El~Zowalaty}, \bibinfo{person}{Hossam~M Ashour}, {and} \bibinfo{person}{Md~Tanvir Rahman}.} \bibinfo{year}{2021}\natexlab{}.
\newblock \showarticletitle{Impact of the COVID-19 pandemic on food production and animal health}.
\newblock \bibinfo{journal}{\emph{Trends in Food Science \& Technology}} (\bibinfo{year}{2021}).
\newblock


\bibitem[\protect\citeauthoryear{Rahnemoonfar and Sheppard}{Rahnemoonfar and Sheppard}{2017}]%
        {rahnemoonfar2017deep}
\bibfield{author}{\bibinfo{person}{Maryam Rahnemoonfar} {and} \bibinfo{person}{Clay Sheppard}.} \bibinfo{year}{2017}\natexlab{}.
\newblock \showarticletitle{Deep count: fruit counting based on deep simulated learning}.
\newblock \bibinfo{journal}{\emph{Sensors}} \bibinfo{volume}{17}, \bibinfo{number}{4} (\bibinfo{year}{2017}), \bibinfo{pages}{905}.
\newblock


\bibitem[\protect\citeauthoryear{Redmon, Divvala, Girshick, and Farhadi}{Redmon et~al\mbox{.}}{2016}]%
        {redmon2016look}
\bibfield{author}{\bibinfo{person}{Joseph Redmon}, \bibinfo{person}{Santosh Divvala}, \bibinfo{person}{Ross Girshick}, {and} \bibinfo{person}{Ali Farhadi}.} \bibinfo{year}{2016}\natexlab{}.
\newblock \showarticletitle{You Only Look Once: Unified, Real-Time Object Detection}.
\newblock \bibinfo{journal}{\emph{arXiv 1506.02640}} (\bibinfo{year}{2016}).
\newblock


\bibitem[\protect\citeauthoryear{Redmon and Farhadi}{Redmon and Farhadi}{2017}]%
        {YOLO9000}
\bibfield{author}{\bibinfo{person}{Joseph Redmon} {and} \bibinfo{person}{Ali Farhadi}.} \bibinfo{year}{2017}\natexlab{}.
\newblock \showarticletitle{YOLO9000: Better, Faster, Stronger}. In \bibinfo{booktitle}{\emph{Proceedings of the IEEE Conference on Computer Vision and Pattern Recognition (CVPR)}}.
\newblock


\bibitem[\protect\citeauthoryear{Rehman, Mahmud, Chang, Jin, and Shin}{Rehman et~al\mbox{.}}{2019}]%
        {rehman2019current}
\bibfield{author}{\bibinfo{person}{Tanzeel~U Rehman}, \bibinfo{person}{Md~Sultan Mahmud}, \bibinfo{person}{Young~K Chang}, \bibinfo{person}{Jian Jin}, {and} \bibinfo{person}{Jaemyung Shin}.} \bibinfo{year}{2019}\natexlab{}.
\newblock \showarticletitle{Current and future applications of statistical machine learning algorithms for agricultural machine vision systems}.
\newblock \bibinfo{journal}{\emph{Computers and electronics in agriculture}}  \bibinfo{volume}{156} (\bibinfo{year}{2019}), \bibinfo{pages}{585--605}.
\newblock


\bibitem[\protect\citeauthoryear{Ren and Zemel}{Ren and Zemel}{2017}]%
        {ren2017end}
\bibfield{author}{\bibinfo{person}{Mengye Ren} {and} \bibinfo{person}{Richard~S Zemel}.} \bibinfo{year}{2017}\natexlab{}.
\newblock \showarticletitle{End-to-end instance segmentation with recurrent attention}. In \bibinfo{booktitle}{\emph{Proceedings of the IEEE conference on computer vision and pattern recognition}}. \bibinfo{pages}{6656--6664}.
\newblock


\bibitem[\protect\citeauthoryear{Ren, He, Girshick, and Sun}{Ren et~al\mbox{.}}{2015}]%
        {FasterRCNN2015}
\bibfield{author}{\bibinfo{person}{Shaoqing Ren}, \bibinfo{person}{Kaiming He}, \bibinfo{person}{Ross Girshick}, {and} \bibinfo{person}{Jian Sun}.} \bibinfo{year}{2015}\natexlab{}.
\newblock \showarticletitle{Faster R-CNN: Towards Real-Time Object Detection with Region Proposal Networks}. In \bibinfo{booktitle}{\emph{Advances in Neural Information Processing Systems}}.
\newblock


\bibitem[\protect\citeauthoryear{Ren, He, Girshick, and Sun}{Ren et~al\mbox{.}}{2016}]%
        {ren2016faster}
\bibfield{author}{\bibinfo{person}{Shaoqing Ren}, \bibinfo{person}{Kaiming He}, \bibinfo{person}{Ross Girshick}, {and} \bibinfo{person}{Jian Sun}.} \bibinfo{year}{2016}\natexlab{}.
\newblock \showarticletitle{Faster R-CNN: Towards Real-Time Object Detection with Region Proposal Networks}.
\newblock \bibinfo{journal}{\emph{arXiv 1506.01497}} (\bibinfo{year}{2016}).
\newblock


\bibitem[\protect\citeauthoryear{Reyes-Yanes, Martinez, and Ahmad}{Reyes-Yanes et~al\mbox{.}}{2020a}]%
        {REYESYANES2020105827}
\bibfield{author}{\bibinfo{person}{A. Reyes-Yanes}, \bibinfo{person}{P. Martinez}, {and} \bibinfo{person}{R. Ahmad}.} \bibinfo{year}{2020}\natexlab{a}.
\newblock \showarticletitle{Real-time growth rate and fresh weight estimation for little gem romaine lettuce in aquaponic grow beds}.
\newblock \bibinfo{journal}{\emph{Computers and Electronics in Agriculture}}  \bibinfo{volume}{179} (\bibinfo{year}{2020}), \bibinfo{pages}{105827}.
\newblock
\showISSN{0168-1699}
\urldef\tempurl%
\url{https://doi.org/10.1016/j.compag.2020.105827}
\showDOI{\tempurl}


\bibitem[\protect\citeauthoryear{Reyes-Yanes, Martinez, and Ahmad}{Reyes-Yanes et~al\mbox{.}}{2020b}]%
        {reyes2020real}
\bibfield{author}{\bibinfo{person}{A Reyes-Yanes}, \bibinfo{person}{Pablo Martinez}, {and} \bibinfo{person}{R Ahmad}.} \bibinfo{year}{2020}\natexlab{b}.
\newblock \showarticletitle{Real-time growth rate and fresh weight estimation for little gem romaine lettuce in aquaponic grow beds}.
\newblock \bibinfo{journal}{\emph{Computers and Electronics in Agriculture}}  \bibinfo{volume}{179} (\bibinfo{year}{2020}), \bibinfo{pages}{105827}.
\newblock


\bibitem[\protect\citeauthoryear{Ribeiro, Singh, and Guestrin}{Ribeiro et~al\mbox{.}}{2016}]%
        {ribeiro2016should}
\bibfield{author}{\bibinfo{person}{Marco~Tulio Ribeiro}, \bibinfo{person}{Sameer Singh}, {and} \bibinfo{person}{Carlos Guestrin}.} \bibinfo{year}{2016}\natexlab{}.
\newblock \showarticletitle{" Why should i trust you?" Explaining the predictions of any classifier}. In \bibinfo{booktitle}{\emph{Proceedings of the 22nd ACM SIGKDD international conference on knowledge discovery and data mining}}. \bibinfo{pages}{1135--1144}.
\newblock


\bibitem[\protect\citeauthoryear{Roberts, Bruce, Monaghan, Pope, Leather, and Beacham}{Roberts et~al\mbox{.}}{2020}]%
        {roberts2020vertical}
\bibfield{author}{\bibinfo{person}{Joe~M Roberts}, \bibinfo{person}{Toby~JA Bruce}, \bibinfo{person}{James~M Monaghan}, \bibinfo{person}{Tom~W Pope}, \bibinfo{person}{Simon~R Leather}, {and} \bibinfo{person}{Andrew~M Beacham}.} \bibinfo{year}{2020}\natexlab{}.
\newblock \showarticletitle{Vertical farming systems bring new considerations for pest and disease management}.
\newblock \bibinfo{journal}{\emph{Annals of Applied Biology}} \bibinfo{volume}{176}, \bibinfo{number}{3} (\bibinfo{year}{2020}), \bibinfo{pages}{226--232}.
\newblock


\bibitem[\protect\citeauthoryear{Rolnick, Donti, Kaack, Kochanski, Lacoste, Sankaran, Ross, Milojevic-Dupont, Jaques, Waldman-Brown, Luccioni, Maharaj, Sherwin, Mukkavilli, Kording, Gomes, Ng, Hassabis, Platt, Creutzig, Chayes, and Bengio}{Rolnick et~al\mbox{.}}{2022}]%
        {10.1145/3485128}
\bibfield{author}{\bibinfo{person}{David Rolnick}, \bibinfo{person}{Priya~L. Donti}, \bibinfo{person}{Lynn~H. Kaack}, \bibinfo{person}{Kelly Kochanski}, \bibinfo{person}{Alexandre Lacoste}, \bibinfo{person}{Kris Sankaran}, \bibinfo{person}{Andrew~Slavin Ross}, \bibinfo{person}{Nikola Milojevic-Dupont}, \bibinfo{person}{Natasha Jaques}, \bibinfo{person}{Anna Waldman-Brown}, \bibinfo{person}{Alexandra~Sasha Luccioni}, \bibinfo{person}{Tegan Maharaj}, \bibinfo{person}{Evan~D. Sherwin}, \bibinfo{person}{S.~Karthik Mukkavilli}, \bibinfo{person}{Konrad~P. Kording}, \bibinfo{person}{Carla~P. Gomes}, \bibinfo{person}{Andrew~Y. Ng}, \bibinfo{person}{Demis Hassabis}, \bibinfo{person}{John~C. Platt}, \bibinfo{person}{Felix Creutzig}, \bibinfo{person}{Jennifer Chayes}, {and} \bibinfo{person}{Yoshua Bengio}.} \bibinfo{year}{2022}\natexlab{}.
\newblock \showarticletitle{Tackling Climate Change with Machine Learning}.
\newblock \bibinfo{journal}{\emph{ACM Comput. Surv.}} \bibinfo{volume}{55}, \bibinfo{number}{2}, Article \bibinfo{articleno}{42} (\bibinfo{date}{feb} \bibinfo{year}{2022}), \bibinfo{numpages}{96}~pages.
\newblock
\showISSN{0360-0300}
\urldef\tempurl%
\url{https://doi.org/10.1145/3485128}
\showDOI{\tempurl}


\bibitem[\protect\citeauthoryear{Romera, Alvarez, Bergasa, and Arroyo}{Romera et~al\mbox{.}}{2017}]%
        {romera2017erfnet}
\bibfield{author}{\bibinfo{person}{Eduardo Romera}, \bibinfo{person}{Jos{\'e}~M Alvarez}, \bibinfo{person}{Luis~M Bergasa}, {and} \bibinfo{person}{Roberto Arroyo}.} \bibinfo{year}{2017}\natexlab{}.
\newblock \showarticletitle{Erfnet: Efficient residual factorized convnet for real-time semantic segmentation}.
\newblock \bibinfo{journal}{\emph{IEEE Transactions on Intelligent Transportation Systems}} \bibinfo{volume}{19}, \bibinfo{number}{1} (\bibinfo{year}{2017}), \bibinfo{pages}{263--272}.
\newblock


\bibitem[\protect\citeauthoryear{Romera-Paredes and Torr}{Romera-Paredes and Torr}{2016}]%
        {romera2016recurrent}
\bibfield{author}{\bibinfo{person}{Bernardino Romera-Paredes} {and} \bibinfo{person}{Philip Hilaire~Sean Torr}.} \bibinfo{year}{2016}\natexlab{}.
\newblock \showarticletitle{Recurrent instance segmentation}. In \bibinfo{booktitle}{\emph{European conference on computer vision}}. Springer, \bibinfo{pages}{312--329}.
\newblock


\bibitem[\protect\citeauthoryear{Ronneberger, Fischer, and Brox}{Ronneberger et~al\mbox{.}}{2015}]%
        {Ronneberger2015}
\bibfield{author}{\bibinfo{person}{Olaf Ronneberger}, \bibinfo{person}{Philipp Fischer}, {and} \bibinfo{person}{Thomas Brox}.} \bibinfo{year}{2015}\natexlab{}.
\newblock \showarticletitle{U-Net: Convolutional Networks for Biomedical Image Segmentation}. In \bibinfo{booktitle}{\emph{Medical Image Computing and Computer-Assisted Intervention -- MICCAI 2015}}, \bibfield{editor}{\bibinfo{person}{Nassir Navab}, \bibinfo{person}{Joachim Hornegger}, \bibinfo{person}{William~M. Wells}, {and} \bibinfo{person}{Alejandro~F. Frangi}} (Eds.). \bibinfo{publisher}{Springer International Publishing}, \bibinfo{address}{Cham}, \bibinfo{pages}{234--241}.
\newblock
\showISBNx{978-3-319-24574-4}


\bibitem[\protect\citeauthoryear{Russakovsky, Deng, Su, Krause, Satheesh, Ma, Huang, Karpathy, Khosla, Bernstein, Berg, and Fei-Fei}{Russakovsky et~al\mbox{.}}{2015}]%
        {ILSVRC15}
\bibfield{author}{\bibinfo{person}{Olga Russakovsky}, \bibinfo{person}{Jia Deng}, \bibinfo{person}{Hao Su}, \bibinfo{person}{Jonathan Krause}, \bibinfo{person}{Sanjeev Satheesh}, \bibinfo{person}{Sean Ma}, \bibinfo{person}{Zhiheng Huang}, \bibinfo{person}{Andrej Karpathy}, \bibinfo{person}{Aditya Khosla}, \bibinfo{person}{Michael Bernstein}, \bibinfo{person}{Alexander~C. Berg}, {and} \bibinfo{person}{Li Fei-Fei}.} \bibinfo{year}{2015}\natexlab{}.
\newblock \showarticletitle{{ImageNet Large Scale Visual Recognition Challenge}}.
\newblock \bibinfo{journal}{\emph{International Journal of Computer Vision (IJCV)}} \bibinfo{volume}{115}, \bibinfo{number}{3} (\bibinfo{year}{2015}), \bibinfo{pages}{211--252}.
\newblock
\urldef\tempurl%
\url{https://doi.org/10.1007/s11263-015-0816-y}
\showDOI{\tempurl}


\bibitem[\protect\citeauthoryear{Rusu}{Rusu}{2010}]%
        {rusu2010semantic}
\bibfield{author}{\bibinfo{person}{Radu~Bogdan Rusu}.} \bibinfo{year}{2010}\natexlab{}.
\newblock \showarticletitle{Semantic 3D object maps for everyday manipulation in human living environments}.
\newblock \bibinfo{journal}{\emph{KI-K{\"u}nstliche Intelligenz}} \bibinfo{volume}{24}, \bibinfo{number}{4} (\bibinfo{year}{2010}), \bibinfo{pages}{345--348}.
\newblock


\bibitem[\protect\citeauthoryear{Sa, Ge, Dayoub, Upcroft, Perez, and McCool}{Sa et~al\mbox{.}}{2016a}]%
        {sa2016deepfruits}
\bibfield{author}{\bibinfo{person}{Inkyu Sa}, \bibinfo{person}{Zongyuan Ge}, \bibinfo{person}{Feras Dayoub}, \bibinfo{person}{Ben Upcroft}, \bibinfo{person}{Tristan Perez}, {and} \bibinfo{person}{Chris McCool}.} \bibinfo{year}{2016}\natexlab{a}.
\newblock \showarticletitle{Deepfruits: A fruit detection system using deep neural networks}.
\newblock \bibinfo{journal}{\emph{sensors}} \bibinfo{volume}{16}, \bibinfo{number}{8} (\bibinfo{year}{2016}), \bibinfo{pages}{1222}.
\newblock


\bibitem[\protect\citeauthoryear{Sa, Ge, Dayoub, Upcroft, Perez, and McCool}{Sa et~al\mbox{.}}{2016b}]%
        {s16081222}
\bibfield{author}{\bibinfo{person}{Inkyu Sa}, \bibinfo{person}{Zongyuan Ge}, \bibinfo{person}{Feras Dayoub}, \bibinfo{person}{Ben Upcroft}, \bibinfo{person}{Tristan Perez}, {and} \bibinfo{person}{Chris McCool}.} \bibinfo{year}{2016}\natexlab{b}.
\newblock \showarticletitle{DeepFruits: A Fruit Detection System Using Deep Neural Networks}.
\newblock \bibinfo{journal}{\emph{Sensors}} \bibinfo{volume}{16}, \bibinfo{number}{8} (\bibinfo{year}{2016}).
\newblock
\showISSN{1424-8220}
\urldef\tempurl%
\url{https://doi.org/10.3390/s16081222}
\showDOI{\tempurl}


\bibitem[\protect\citeauthoryear{Saiz-Rubio and Rovira-M{\'a}s}{Saiz-Rubio and Rovira-M{\'a}s}{2020}]%
        {saiz2020smart}
\bibfield{author}{\bibinfo{person}{Ver{\'o}nica Saiz-Rubio} {and} \bibinfo{person}{Francisco Rovira-M{\'a}s}.} \bibinfo{year}{2020}\natexlab{}.
\newblock \showarticletitle{From smart farming towards agriculture 5.0: A review on crop data management}.
\newblock \bibinfo{journal}{\emph{Agronomy}} \bibinfo{volume}{10}, \bibinfo{number}{2} (\bibinfo{year}{2020}), \bibinfo{pages}{207}.
\newblock


\bibitem[\protect\citeauthoryear{Salvador, Bellver, Campos, Baradad, Marques, Torres, and Giro-i Nieto}{Salvador et~al\mbox{.}}{2017a}]%
        {Salvador2017}
\bibfield{author}{\bibinfo{person}{Amaia Salvador}, \bibinfo{person}{Miriam Bellver}, \bibinfo{person}{Victor Campos}, \bibinfo{person}{Manel Baradad}, \bibinfo{person}{Ferran Marques}, \bibinfo{person}{Jordi Torres}, {and} \bibinfo{person}{Xavier Giro-i Nieto}.} \bibinfo{year}{2017}\natexlab{a}.
\newblock \showarticletitle{Recurrent Neural Networks for Semantic Instance Segmentation}.
\newblock \bibinfo{journal}{\emph{arXiv Preprint 1712.00617}} (\bibinfo{year}{2017}).
\newblock
\urldef\tempurl%
\url{https://arxiv.org/abs/1712.00617}
\showURL{%
\tempurl}


\bibitem[\protect\citeauthoryear{Salvador, Bellver, Campos, Baradad, Marques, Torres, and Giro-i Nieto}{Salvador et~al\mbox{.}}{2017b}]%
        {salvador2017recurrent}
\bibfield{author}{\bibinfo{person}{Amaia Salvador}, \bibinfo{person}{Miriam Bellver}, \bibinfo{person}{Victor Campos}, \bibinfo{person}{Manel Baradad}, \bibinfo{person}{Ferran Marques}, \bibinfo{person}{Jordi Torres}, {and} \bibinfo{person}{Xavier Giro-i Nieto}.} \bibinfo{year}{2017}\natexlab{b}.
\newblock \showarticletitle{Recurrent neural networks for semantic instance segmentation}.
\newblock \bibinfo{journal}{\emph{arXiv preprint arXiv:1712.00617}} (\bibinfo{year}{2017}).
\newblock


\bibitem[\protect\citeauthoryear{Scharr, Minervini, French, Klukas, Kramer, Liu, Luengo, Pape, Polder, Vukadinovic, et~al\mbox{.}}{Scharr et~al\mbox{.}}{2016}]%
        {scharr2016leaf}
\bibfield{author}{\bibinfo{person}{Hanno Scharr}, \bibinfo{person}{Massimo Minervini}, \bibinfo{person}{Andrew~P French}, \bibinfo{person}{Christian Klukas}, \bibinfo{person}{David~M Kramer}, \bibinfo{person}{Xiaoming Liu}, \bibinfo{person}{Imanol Luengo}, \bibinfo{person}{Jean-Michel Pape}, \bibinfo{person}{Gerrit Polder}, \bibinfo{person}{Danijela Vukadinovic}, {et~al\mbox{.}}} \bibinfo{year}{2016}\natexlab{}.
\newblock \showarticletitle{Leaf segmentation in plant phenotyping: a collation study}.
\newblock \bibinfo{journal}{\emph{Machine vision and applications}} \bibinfo{volume}{27}, \bibinfo{number}{4} (\bibinfo{year}{2016}), \bibinfo{pages}{585--606}.
\newblock


\bibitem[\protect\citeauthoryear{Schmarje, Santarossa, Schr{\"o}der, and Koch}{Schmarje et~al\mbox{.}}{2021}]%
        {schmarje2021survey}
\bibfield{author}{\bibinfo{person}{Lars Schmarje}, \bibinfo{person}{Monty Santarossa}, \bibinfo{person}{Simon-Martin Schr{\"o}der}, {and} \bibinfo{person}{Reinhard Koch}.} \bibinfo{year}{2021}\natexlab{}.
\newblock \showarticletitle{A survey on semi-, self-and unsupervised learning for image classification}.
\newblock \bibinfo{journal}{\emph{IEEE Access}}  \bibinfo{volume}{9} (\bibinfo{year}{2021}), \bibinfo{pages}{82146--82168}.
\newblock


\bibitem[\protect\citeauthoryear{Selvaraj, Vergara, Ruiz, Safari, Elayabalan, Ocimati, and Blomme}{Selvaraj et~al\mbox{.}}{2019}]%
        {selvaraj2019ai}
\bibfield{author}{\bibinfo{person}{Michael~Gomez Selvaraj}, \bibinfo{person}{Alejandro Vergara}, \bibinfo{person}{Henry Ruiz}, \bibinfo{person}{Nancy Safari}, \bibinfo{person}{Sivalingam Elayabalan}, \bibinfo{person}{Walter Ocimati}, {and} \bibinfo{person}{Guy Blomme}.} \bibinfo{year}{2019}\natexlab{}.
\newblock \showarticletitle{AI-powered banana diseases and pest detection}.
\newblock \bibinfo{journal}{\emph{Plant Methods}} \bibinfo{volume}{15}, \bibinfo{number}{1} (\bibinfo{year}{2019}), \bibinfo{pages}{1--11}.
\newblock


\bibitem[\protect\citeauthoryear{Selvaraju, Cogswell, Das, Vedantam, Parikh, and Batra}{Selvaraju et~al\mbox{.}}{2017}]%
        {selvaraju2017grad}
\bibfield{author}{\bibinfo{person}{Ramprasaath~R Selvaraju}, \bibinfo{person}{Michael Cogswell}, \bibinfo{person}{Abhishek Das}, \bibinfo{person}{Ramakrishna Vedantam}, \bibinfo{person}{Devi Parikh}, {and} \bibinfo{person}{Dhruv Batra}.} \bibinfo{year}{2017}\natexlab{}.
\newblock \showarticletitle{Grad-cam: Visual explanations from deep networks via gradient-based localization}. In \bibinfo{booktitle}{\emph{Proceedings of the IEEE international conference on computer vision}}. \bibinfo{pages}{618--626}.
\newblock


\bibitem[\protect\citeauthoryear{Seng and Mirisaee}{Seng and Mirisaee}{2009}]%
        {seng2009new}
\bibfield{author}{\bibinfo{person}{Woo~Chaw Seng} {and} \bibinfo{person}{Seyed~Hadi Mirisaee}.} \bibinfo{year}{2009}\natexlab{}.
\newblock \showarticletitle{A new method for fruits recognition system}. In \bibinfo{booktitle}{\emph{2009 international conference on electrical engineering and informatics}}, Vol.~\bibinfo{volume}{1}. IEEE, \bibinfo{pages}{130--134}.
\newblock


\bibitem[\protect\citeauthoryear{Senthilnath, Dokania, Kandukuri, Ramesh, Anand, and Omkar}{Senthilnath et~al\mbox{.}}{2016}]%
        {senthilnath2016detection}
\bibfield{author}{\bibinfo{person}{Jayavelu Senthilnath}, \bibinfo{person}{Akanksha Dokania}, \bibinfo{person}{Manasa Kandukuri}, \bibinfo{person}{KN Ramesh}, \bibinfo{person}{Gautham Anand}, {and} \bibinfo{person}{SN Omkar}.} \bibinfo{year}{2016}\natexlab{}.
\newblock \showarticletitle{Detection of tomatoes using spectral-spatial methods in remotely sensed RGB images captured by UAV}.
\newblock \bibinfo{journal}{\emph{Biosystems engineering}}  \bibinfo{volume}{146} (\bibinfo{year}{2016}), \bibinfo{pages}{16--32}.
\newblock


\bibitem[\protect\citeauthoryear{Sermanet, Eigen, Zhang, Mathieu, Fergus, and LeCun}{Sermanet et~al\mbox{.}}{2013}]%
        {sermanet2013overfeat}
\bibfield{author}{\bibinfo{person}{Pierre Sermanet}, \bibinfo{person}{David Eigen}, \bibinfo{person}{Xiang Zhang}, \bibinfo{person}{Micha{\"e}l Mathieu}, \bibinfo{person}{Rob Fergus}, {and} \bibinfo{person}{Yann LeCun}.} \bibinfo{year}{2013}\natexlab{}.
\newblock \showarticletitle{Overfeat: Integrated recognition, localization and detection using convolutional networks}.
\newblock \bibinfo{journal}{\emph{arXiv preprint arXiv:1312.6229}} (\bibinfo{year}{2013}).
\newblock


\bibitem[\protect\citeauthoryear{Sha, Camburu, and Lukasiewicz}{Sha et~al\mbox{.}}{2021}]%
        {sha2021learning}
\bibfield{author}{\bibinfo{person}{Lei Sha}, \bibinfo{person}{Oana-Maria Camburu}, {and} \bibinfo{person}{Thomas Lukasiewicz}.} \bibinfo{year}{2021}\natexlab{}.
\newblock \showarticletitle{Learning from the Best: Rationalizing Predictions by Adversarial Information Calibration.}. In \bibinfo{booktitle}{\emph{AAAI}}. \bibinfo{pages}{13771--13779}.
\newblock


\bibitem[\protect\citeauthoryear{Sharma, Dhanaraj, Karnam, Chachlakis, Ptucha, Markopoulos, and Saber}{Sharma et~al\mbox{.}}{2020}]%
        {sharma2020yolors}
\bibfield{author}{\bibinfo{person}{Manish Sharma}, \bibinfo{person}{Mayur Dhanaraj}, \bibinfo{person}{Srivallabha Karnam}, \bibinfo{person}{Dimitris~G Chachlakis}, \bibinfo{person}{Raymond Ptucha}, \bibinfo{person}{Panos~P Markopoulos}, {and} \bibinfo{person}{Eli Saber}.} \bibinfo{year}{2020}\natexlab{}.
\newblock \showarticletitle{YOLOrs: Object detection in multimodal remote sensing imagery}.
\newblock \bibinfo{journal}{\emph{IEEE Journal of Selected Topics in Applied Earth Observations and Remote Sensing}}  \bibinfo{volume}{14} (\bibinfo{year}{2020}), \bibinfo{pages}{1497--1508}.
\newblock


\bibitem[\protect\citeauthoryear{Shen, Lin, and Huang}{Shen et~al\mbox{.}}{2016}]%
        {shen2016relay}
\bibfield{author}{\bibinfo{person}{Li Shen}, \bibinfo{person}{Zhouchen Lin}, {and} \bibinfo{person}{Qingming Huang}.} \bibinfo{year}{2016}\natexlab{}.
\newblock \showarticletitle{Relay backpropagation for effective learning of deep convolutional neural networks}. In \bibinfo{booktitle}{\emph{ECCV}}. Springer, \bibinfo{pages}{467--482}.
\newblock


\bibitem[\protect\citeauthoryear{Shi, Li, and Yamaguchi}{Shi et~al\mbox{.}}{2020a}]%
        {shi2020attribution}
\bibfield{author}{\bibinfo{person}{Rui Shi}, \bibinfo{person}{Tianxing Li}, {and} \bibinfo{person}{Yasushi Yamaguchi}.} \bibinfo{year}{2020}\natexlab{a}.
\newblock \showarticletitle{An attribution-based pruning method for real-time mango detection with YOLO network}.
\newblock \bibinfo{journal}{\emph{Computers and electronics in agriculture}}  \bibinfo{volume}{169} (\bibinfo{year}{2020}), \bibinfo{pages}{105214}.
\newblock


\bibitem[\protect\citeauthoryear{Shi, Zhai, Liu, Jiang, and Gao}{Shi et~al\mbox{.}}{2020b}]%
        {shi2020rectified}
\bibfield{author}{\bibinfo{person}{Ruifeng Shi}, \bibinfo{person}{Deming Zhai}, \bibinfo{person}{Xianming Liu}, \bibinfo{person}{Junjun Jiang}, {and} \bibinfo{person}{Wen Gao}.} \bibinfo{year}{2020}\natexlab{b}.
\newblock \showarticletitle{Rectified meta-learning from noisy labels for robust image-based plant disease diagnosis}.
\newblock \bibinfo{journal}{\emph{arXiv preprint arXiv:2003.07603}} (\bibinfo{year}{2020}).
\newblock


\bibitem[\protect\citeauthoryear{Shimamura}{Shimamura}{[n.\,d.]}]%
        {JapanVertifarm}
\bibfield{author}{\bibinfo{person}{Shigeharu Shimamura}.} \bibinfo{year}{[n.\,d.]}\natexlab{}.
\newblock \bibinfo{title}{Indoor Cultivation for the Future}.
\newblock
\newblock
\newblock
\shownote{\url{https://frc.ri.cmu.edu/~ssingh/VF/Challenges_in_Vertical_Farming/Schedule_files/SHIMAMURA.pdf}}.


\bibitem[\protect\citeauthoryear{Shitole, Li, Kahng, Tadepalli, and Fern}{Shitole et~al\mbox{.}}{2021}]%
        {shitole2021one}
\bibfield{author}{\bibinfo{person}{Vivswan Shitole}, \bibinfo{person}{Fuxin Li}, \bibinfo{person}{Minsuk Kahng}, \bibinfo{person}{Prasad Tadepalli}, {and} \bibinfo{person}{Alan Fern}.} \bibinfo{year}{2021}\natexlab{}.
\newblock \showarticletitle{One explanation is not enough: structured attention graphs for image classification}.
\newblock \bibinfo{journal}{\emph{Advances in Neural Information Processing Systems}}  \bibinfo{volume}{34} (\bibinfo{year}{2021}), \bibinfo{pages}{11352--11363}.
\newblock


\bibitem[\protect\citeauthoryear{Shrikumar, Greenside, Shcherbina, and Kundaje}{Shrikumar et~al\mbox{.}}{2016}]%
        {shrikumar2016not}
\bibfield{author}{\bibinfo{person}{Avanti Shrikumar}, \bibinfo{person}{Peyton Greenside}, \bibinfo{person}{Anna Shcherbina}, {and} \bibinfo{person}{Anshul Kundaje}.} \bibinfo{year}{2016}\natexlab{}.
\newblock \showarticletitle{Not just a black box: Interpretable deep learning by propagating activation differences}.
\newblock \bibinfo{journal}{\emph{arXiv preprint arXiv:1605.01713}}  \bibinfo{volume}{4} (\bibinfo{year}{2016}).
\newblock


\bibitem[\protect\citeauthoryear{Siddharth, Ajay, Uday, and Sanjeev}{Siddharth et~al\mbox{.}}{2019}]%
        {siddharth2019database}
\bibfield{author}{\bibinfo{person}{Singh~Chouhan Siddharth}, \bibinfo{person}{Kaul Ajay}, \bibinfo{person}{Pratap~Singh Uday}, {and} \bibinfo{person}{Jain Sanjeev}.} \bibinfo{year}{2019}\natexlab{}.
\newblock \showarticletitle{A database of leaf images: practice towards plant conservation with plant pathology}.
\newblock \bibinfo{journal}{\emph{Mendeley Data}} (\bibinfo{year}{2019}).
\newblock


\bibitem[\protect\citeauthoryear{Sidor and Rzymski}{Sidor and Rzymski}{2020}]%
        {sidor2020dietary}
\bibfield{author}{\bibinfo{person}{Aleksandra Sidor} {and} \bibinfo{person}{Piotr Rzymski}.} \bibinfo{year}{2020}\natexlab{}.
\newblock \showarticletitle{Dietary choices and habits during COVID-19 lockdown: experience from Poland}.
\newblock \bibinfo{journal}{\emph{Nutrients}} \bibinfo{volume}{12}, \bibinfo{number}{6} (\bibinfo{year}{2020}), \bibinfo{pages}{1657}.
\newblock


\bibitem[\protect\citeauthoryear{Silva, Uesugi, Blum, Marques, and Ferreira}{Silva et~al\mbox{.}}{2016}]%
        {silva2016molecular}
\bibfield{author}{\bibinfo{person}{Claud{\^e}nia Ferreira~da Silva}, \bibinfo{person}{Carlos~Hidemi Uesugi}, \bibinfo{person}{Luiz Eduardo~Bassay Blum}, \bibinfo{person}{Abi Soares dos~Anjos Marques}, {and} \bibinfo{person}{Marisa {\'A}lvares da Silva~Velloso Ferreira}.} \bibinfo{year}{2016}\natexlab{}.
\newblock \showarticletitle{Molecular detection of Erwinia psidii in guava plants under greenhouse and field conditions}.
\newblock \bibinfo{journal}{\emph{Ci{\^e}ncia Rural}}  \bibinfo{volume}{46} (\bibinfo{year}{2016}), \bibinfo{pages}{1528--1534}.
\newblock


\bibitem[\protect\citeauthoryear{Silver, Schrittwieser, Simonyan, Antonoglou, Huang, Guez, Hubert, Baker, Lai, Bolton, Chen, Lillicrap, Hui, Sifre, van~den Driessche, Graepel, and Hassabis}{Silver et~al\mbox{.}}{2017}]%
        {Silver2017}
\bibfield{author}{\bibinfo{person}{David Silver}, \bibinfo{person}{Julian Schrittwieser}, \bibinfo{person}{Karen Simonyan}, \bibinfo{person}{Ioannis Antonoglou}, \bibinfo{person}{Aja Huang}, \bibinfo{person}{Arthur Guez}, \bibinfo{person}{Thomas Hubert}, \bibinfo{person}{Lucas Baker}, \bibinfo{person}{Matthew Lai}, \bibinfo{person}{Adrian Bolton}, \bibinfo{person}{Yutian Chen}, \bibinfo{person}{Timothy Lillicrap}, \bibinfo{person}{Fan Hui}, \bibinfo{person}{Laurent Sifre}, \bibinfo{person}{George van~den Driessche}, \bibinfo{person}{Thore Graepel}, {and} \bibinfo{person}{Demis Hassabis}.} \bibinfo{year}{2017}\natexlab{}.
\newblock \showarticletitle{Mastering the game of Go without human knowledge}.
\newblock \bibinfo{journal}{\emph{Nature}}  \bibinfo{volume}{550} (\bibinfo{year}{2017}), \bibinfo{pages}{354–359}.
\newblock


\bibitem[\protect\citeauthoryear{Simonyan and Zisserman}{Simonyan and Zisserman}{2015}]%
        {Simonyan15:VGG}
\bibfield{author}{\bibinfo{person}{Karen Simonyan} {and} \bibinfo{person}{Andrew Zisserman}.} \bibinfo{year}{2015}\natexlab{}.
\newblock \showarticletitle{Very Deep Convolutional Networks for Large-Scale Image Recognition}. In \bibinfo{booktitle}{\emph{International Conference on Learning Representations}}.
\newblock


\bibitem[\protect\citeauthoryear{Singh, Chouhan, Jain, and Jain}{Singh et~al\mbox{.}}{2019}]%
        {singh2019multilayer}
\bibfield{author}{\bibinfo{person}{Uday~Pratap Singh}, \bibinfo{person}{Siddharth~Singh Chouhan}, \bibinfo{person}{Sukirty Jain}, {and} \bibinfo{person}{Sanjeev Jain}.} \bibinfo{year}{2019}\natexlab{}.
\newblock \showarticletitle{Multilayer convolution neural network for the classification of mango leaves infected by anthracnose disease}.
\newblock \bibinfo{journal}{\emph{IEEE Access}}  \bibinfo{volume}{7} (\bibinfo{year}{2019}), \bibinfo{pages}{43721--43729}.
\newblock


\bibitem[\protect\citeauthoryear{Smilkov, Thorat, Kim, Vi{\'e}gas, and Wattenberg}{Smilkov et~al\mbox{.}}{2017}]%
        {smilkov2017smoothgrad}
\bibfield{author}{\bibinfo{person}{Daniel Smilkov}, \bibinfo{person}{Nikhil Thorat}, \bibinfo{person}{Been Kim}, \bibinfo{person}{Fernanda Vi{\'e}gas}, {and} \bibinfo{person}{Martin Wattenberg}.} \bibinfo{year}{2017}\natexlab{}.
\newblock \showarticletitle{Smoothgrad: removing noise by adding noise}.
\newblock \bibinfo{journal}{\emph{arXiv preprint arXiv:1706.03825}} (\bibinfo{year}{2017}).
\newblock


\bibitem[\protect\citeauthoryear{Smith, Dherin, Barrett, and De}{Smith et~al\mbox{.}}{2021}]%
        {smith2021origin}
\bibfield{author}{\bibinfo{person}{Samuel~L Smith}, \bibinfo{person}{Benoit Dherin}, \bibinfo{person}{David~GT Barrett}, {and} \bibinfo{person}{Soham De}.} \bibinfo{year}{2021}\natexlab{}.
\newblock \showarticletitle{On the origin of implicit regularization in stochastic gradient descent}.
\newblock \bibinfo{journal}{\emph{arXiv preprint arXiv:2101.12176}} (\bibinfo{year}{2021}).
\newblock


\bibitem[\protect\citeauthoryear{Snell, Swersky, and Zemel}{Snell et~al\mbox{.}}{2017}]%
        {snell2017prototypical}
\bibfield{author}{\bibinfo{person}{Jake Snell}, \bibinfo{person}{Kevin Swersky}, {and} \bibinfo{person}{Richard Zemel}.} \bibinfo{year}{2017}\natexlab{}.
\newblock \showarticletitle{Prototypical networks for few-shot learning}.
\newblock \bibinfo{journal}{\emph{Advances in neural information processing systems}}  \bibinfo{volume}{30} (\bibinfo{year}{2017}).
\newblock


\bibitem[\protect\citeauthoryear{Song, Shen, Lei, Zeng, Ou, Tao, and Song}{Song et~al\mbox{.}}{2018a}]%
        {song2018selective}
\bibfield{author}{\bibinfo{person}{Jie Song}, \bibinfo{person}{Chengchao Shen}, \bibinfo{person}{Jie Lei}, \bibinfo{person}{An-Xiang Zeng}, \bibinfo{person}{Kairi Ou}, \bibinfo{person}{Dacheng Tao}, {and} \bibinfo{person}{Mingli Song}.} \bibinfo{year}{2018}\natexlab{a}.
\newblock \showarticletitle{Selective zero-shot classification with augmented attributes}. In \bibinfo{booktitle}{\emph{Proceedings of the European Conference on Computer Vision (ECCV)}}. \bibinfo{pages}{468--483}.
\newblock


\bibitem[\protect\citeauthoryear{Song, Shen, Yang, Liu, and Song}{Song et~al\mbox{.}}{2018b}]%
        {song2018transductive}
\bibfield{author}{\bibinfo{person}{Jie Song}, \bibinfo{person}{Chengchao Shen}, \bibinfo{person}{Yezhou Yang}, \bibinfo{person}{Yang Liu}, {and} \bibinfo{person}{Mingli Song}.} \bibinfo{year}{2018}\natexlab{b}.
\newblock \showarticletitle{Transductive unbiased embedding for zero-shot learning}. In \bibinfo{booktitle}{\emph{Proceedings of the IEEE conference on computer vision and pattern recognition}}. \bibinfo{pages}{1024--1033}.
\newblock


\bibitem[\protect\citeauthoryear{Sornalakshmi, Sujatha, Sindhu, and Hemavathi}{Sornalakshmi et~al\mbox{.}}{2022}]%
        {sornalakshmi2022technical}
\bibfield{author}{\bibinfo{person}{K Sornalakshmi}, \bibinfo{person}{G Sujatha}, \bibinfo{person}{S Sindhu}, {and} \bibinfo{person}{D Hemavathi}.} \bibinfo{year}{2022}\natexlab{}.
\newblock \showarticletitle{A Technical Survey on Deep Learning and AI Solutions for Plant Quality and Health Indicators Monitoring in Agriculture}. In \bibinfo{booktitle}{\emph{2022 3rd International Conference on Smart Electronics and Communication (ICOSEC)}}. IEEE, \bibinfo{pages}{984--988}.
\newblock


\bibitem[\protect\citeauthoryear{Spalding and Miller}{Spalding and Miller}{2013}]%
        {spalding2013image}
\bibfield{author}{\bibinfo{person}{Edgar~P Spalding} {and} \bibinfo{person}{Nathan~D Miller}.} \bibinfo{year}{2013}\natexlab{}.
\newblock \showarticletitle{Image analysis is driving a renaissance in growth measurement}.
\newblock \bibinfo{journal}{\emph{Current opinion in plant biology}} \bibinfo{volume}{16}, \bibinfo{number}{1} (\bibinfo{year}{2013}), \bibinfo{pages}{100--104}.
\newblock


\bibitem[\protect\citeauthoryear{Standley, Zamir, Chen, Guibas, Malik, and Savarese}{Standley et~al\mbox{.}}{2020}]%
        {standley2020tasks}
\bibfield{author}{\bibinfo{person}{Trevor Standley}, \bibinfo{person}{Amir Zamir}, \bibinfo{person}{Dawn Chen}, \bibinfo{person}{Leonidas Guibas}, \bibinfo{person}{Jitendra Malik}, {and} \bibinfo{person}{Silvio Savarese}.} \bibinfo{year}{2020}\natexlab{}.
\newblock \showarticletitle{Which tasks should be learned together in multi-task learning?}. In \bibinfo{booktitle}{\emph{International Conference on Machine Learning}}. PMLR, \bibinfo{pages}{9120--9132}.
\newblock


\bibitem[\protect\citeauthoryear{Steiner, Kolesnikov, Zhai, Wightman, Uszkoreit, and Beyer}{Steiner et~al\mbox{.}}{2022}]%
        {steinerhow}
\bibfield{author}{\bibinfo{person}{Andreas~Peter Steiner}, \bibinfo{person}{Alexander Kolesnikov}, \bibinfo{person}{Xiaohua Zhai}, \bibinfo{person}{Ross Wightman}, \bibinfo{person}{Jakob Uszkoreit}, {and} \bibinfo{person}{Lucas Beyer}.} \bibinfo{year}{2022}\natexlab{}.
\newblock \showarticletitle{How to train your ViT? Data, Augmentation, and Regularization in Vision Transformers}.
\newblock \bibinfo{journal}{\emph{Transactions on Machine Learning Research}} (\bibinfo{year}{2022}).
\newblock
\urldef\tempurl%
\url{https://openreview.net/forum?id=4nPswr1KcP}
\showURL{%
\tempurl}


\bibitem[\protect\citeauthoryear{Strange and Scott}{Strange and Scott}{2005}]%
        {strange2005plant}
\bibfield{author}{\bibinfo{person}{Richard~N Strange} {and} \bibinfo{person}{Peter~R Scott}.} \bibinfo{year}{2005}\natexlab{}.
\newblock \showarticletitle{Plant disease: a threat to global food security}.
\newblock \bibinfo{journal}{\emph{Annual review of phytopathology}} \bibinfo{volume}{43}, \bibinfo{number}{1} (\bibinfo{year}{2005}), \bibinfo{pages}{83--116}.
\newblock


\bibitem[\protect\citeauthoryear{Sun, Shrivastava, Singh, and Gupta}{Sun et~al\mbox{.}}{2017}]%
        {Sun_2017_ICCV}
\bibfield{author}{\bibinfo{person}{Chen Sun}, \bibinfo{person}{Abhinav Shrivastava}, \bibinfo{person}{Saurabh Singh}, {and} \bibinfo{person}{Abhinav Gupta}.} \bibinfo{year}{2017}\natexlab{}.
\newblock \showarticletitle{Revisiting Unreasonable Effectiveness of Data in Deep Learning Era}. In \bibinfo{booktitle}{\emph{Proceedings of the IEEE International Conference on Computer Vision (ICCV)}}.
\newblock


\bibitem[\protect\citeauthoryear{Sun, He, Ge, Wu, Shen, and Song}{Sun et~al\mbox{.}}{2018}]%
        {sun2018detection}
\bibfield{author}{\bibinfo{person}{Jun Sun}, \bibinfo{person}{Xiaofei He}, \bibinfo{person}{Xiao Ge}, \bibinfo{person}{Xiaohong Wu}, \bibinfo{person}{Jifeng Shen}, {and} \bibinfo{person}{Yingying Song}.} \bibinfo{year}{2018}\natexlab{}.
\newblock \showarticletitle{Detection of key organs in tomato based on deep migration learning in a complex background}.
\newblock \bibinfo{journal}{\emph{Agriculture}} \bibinfo{volume}{8}, \bibinfo{number}{12} (\bibinfo{year}{2018}), \bibinfo{pages}{196}.
\newblock


\bibitem[\protect\citeauthoryear{Sun, Wang, Liu, and Liu}{Sun et~al\mbox{.}}{2021}]%
        {sun2021apple}
\bibfield{author}{\bibinfo{person}{Kaiqiong Sun}, \bibinfo{person}{Xuan Wang}, \bibinfo{person}{Shoushuai Liu}, {and} \bibinfo{person}{ChangHua Liu}.} \bibinfo{year}{2021}\natexlab{}.
\newblock \showarticletitle{Apple, peach, and pear flower detection using semantic segmentation network and shape constraint level set}.
\newblock \bibinfo{journal}{\emph{Computers and Electronics in Agriculture}}  \bibinfo{volume}{185} (\bibinfo{year}{2021}), \bibinfo{pages}{106150}.
\newblock


\bibitem[\protect\citeauthoryear{Sun, Liu, Chua, and Schiele}{Sun et~al\mbox{.}}{2019}]%
        {sun2019meta}
\bibfield{author}{\bibinfo{person}{Qianru Sun}, \bibinfo{person}{Yaoyao Liu}, \bibinfo{person}{Tat-Seng Chua}, {and} \bibinfo{person}{Bernt Schiele}.} \bibinfo{year}{2019}\natexlab{}.
\newblock \showarticletitle{Meta-transfer learning for few-shot learning}. In \bibinfo{booktitle}{\emph{Proceedings of the IEEE/CVF Conference on Computer Vision and Pattern Recognition}}. \bibinfo{pages}{403--412}.
\newblock


\bibitem[\protect\citeauthoryear{Sundararajan, Taly, and Yan}{Sundararajan et~al\mbox{.}}{2017}]%
        {sundararajan2017axiomatic}
\bibfield{author}{\bibinfo{person}{Mukund Sundararajan}, \bibinfo{person}{Ankur Taly}, {and} \bibinfo{person}{Qiqi Yan}.} \bibinfo{year}{2017}\natexlab{}.
\newblock \showarticletitle{Axiomatic attribution for deep networks}. In \bibinfo{booktitle}{\emph{International conference on machine learning}}. PMLR, \bibinfo{pages}{3319--3328}.
\newblock


\bibitem[\protect\citeauthoryear{Susi{\v{c}}, {\v{Z}}ibrat, {\v{S}}irca, Strajnar, Razinger, Knapi{\v{c}}, Von{\v{c}}ina, Urek, and Stare}{Susi{\v{c}} et~al\mbox{.}}{2018}]%
        {susivc2018discrimination}
\bibfield{author}{\bibinfo{person}{Nik Susi{\v{c}}}, \bibinfo{person}{Uro{\v{s}} {\v{Z}}ibrat}, \bibinfo{person}{Sa{\v{s}}a {\v{S}}irca}, \bibinfo{person}{Polona Strajnar}, \bibinfo{person}{Jaka Razinger}, \bibinfo{person}{Matej Knapi{\v{c}}}, \bibinfo{person}{Andrej Von{\v{c}}ina}, \bibinfo{person}{Gregor Urek}, {and} \bibinfo{person}{Barbara~Geri{\v{c}} Stare}.} \bibinfo{year}{2018}\natexlab{}.
\newblock \showarticletitle{Discrimination between abiotic and biotic drought stress in tomatoes using hyperspectral imaging}.
\newblock \bibinfo{journal}{\emph{Sensors and actuators B: Chemical}}  \bibinfo{volume}{273} (\bibinfo{year}{2018}), \bibinfo{pages}{842--852}.
\newblock


\bibitem[\protect\citeauthoryear{Szegedy, Vanhoucke, Ioffe, Shlens, and Wojna}{Szegedy et~al\mbox{.}}{2015}]%
        {Inception2015}
\bibfield{author}{\bibinfo{person}{Christian Szegedy}, \bibinfo{person}{Vincent Vanhoucke}, \bibinfo{person}{Sergey Ioffe}, \bibinfo{person}{Jonathon Shlens}, {and} \bibinfo{person}{Zbigniew Wojna}.} \bibinfo{year}{2015}\natexlab{}.
\newblock \showarticletitle{Rethinking the Inception Architecture for Computer Vision}.
\newblock \bibinfo{journal}{\emph{arXiv Preprint 1512.00567}} (\bibinfo{year}{2015}).
\newblock


\bibitem[\protect\citeauthoryear{Szegedy, Vanhoucke, Ioffe, Shlens, and Wojna}{Szegedy et~al\mbox{.}}{2016}]%
        {Szegedy2016:Inception}
\bibfield{author}{\bibinfo{person}{Christian Szegedy}, \bibinfo{person}{Vincent Vanhoucke}, \bibinfo{person}{Sergey Ioffe}, \bibinfo{person}{Jon Shlens}, {and} \bibinfo{person}{Zbigniew Wojna}.} \bibinfo{year}{2016}\natexlab{}.
\newblock \showarticletitle{Rethinking the Inception Architecture for Computer Vision}. In \bibinfo{booktitle}{\emph{2016 IEEE Conference on Computer Vision and Pattern Recognition (CVPR)}}. \bibinfo{pages}{2818--2826}.
\newblock
\urldef\tempurl%
\url{https://doi.org/10.1109/CVPR.2016.308}
\showDOI{\tempurl}


\bibitem[\protect\citeauthoryear{Szegedy, Zaremba, Sutskever, Bruna, Erhan, Goodfellow, and Fergus}{Szegedy et~al\mbox{.}}{2013}]%
        {szegedy2013intriguing}
\bibfield{author}{\bibinfo{person}{Christian Szegedy}, \bibinfo{person}{Wojciech Zaremba}, \bibinfo{person}{Ilya Sutskever}, \bibinfo{person}{Joan Bruna}, \bibinfo{person}{Dumitru Erhan}, \bibinfo{person}{Ian Goodfellow}, {and} \bibinfo{person}{Rob Fergus}.} \bibinfo{year}{2013}\natexlab{}.
\newblock \showarticletitle{Intriguing properties of neural networks}.
\newblock \bibinfo{journal}{\emph{arXiv preprint arXiv:1312.6199}} (\bibinfo{year}{2013}).
\newblock


\bibitem[\protect\citeauthoryear{Sünderhauf, Brock, Scheirer, Hadsell, Fox, Leitner, Upcroft, Abbeel, Burgard, Milford, and Corke}{Sünderhauf et~al\mbox{.}}{2018}]%
        {Niko2018}
\bibfield{author}{\bibinfo{person}{Niko Sünderhauf}, \bibinfo{person}{Oliver Brock}, \bibinfo{person}{Walter Scheirer}, \bibinfo{person}{Raia Hadsell}, \bibinfo{person}{Dieter Fox}, \bibinfo{person}{Jürgen Leitner}, \bibinfo{person}{Ben Upcroft}, \bibinfo{person}{Pieter Abbeel}, \bibinfo{person}{Wolfram Burgard}, \bibinfo{person}{Michael Milford}, {and} \bibinfo{person}{Peter Corke}.} \bibinfo{year}{2018}\natexlab{}.
\newblock \showarticletitle{The limits and potentials of deep learning for robotics}.
\newblock \bibinfo{journal}{\emph{The International Journal of Robotics Research}} \bibinfo{volume}{37}, \bibinfo{number}{4-5} (\bibinfo{year}{2018}), \bibinfo{pages}{405--420}.
\newblock


\bibitem[\protect\citeauthoryear{Tan and Le}{Tan and Le}{2019a}]%
        {EfficientNet-2019}
\bibfield{author}{\bibinfo{person}{Mingxing Tan} {and} \bibinfo{person}{Quoc Le}.} \bibinfo{year}{2019}\natexlab{a}.
\newblock \showarticletitle{{E}fficient{N}et: Rethinking Model Scaling for Convolutional Neural Networks}. In \bibinfo{booktitle}{\emph{Proceedings of the 36th International Conference on Machine Learning}} \emph{(\bibinfo{series}{Proceedings of Machine Learning Research}, Vol.~\bibinfo{volume}{97})}, \bibfield{editor}{\bibinfo{person}{Kamalika Chaudhuri} {and} \bibinfo{person}{Ruslan Salakhutdinov}} (Eds.). \bibinfo{publisher}{PMLR}, \bibinfo{pages}{6105--6114}.
\newblock
\urldef\tempurl%
\url{https://proceedings.mlr.press/v97/tan19a.html}
\showURL{%
\tempurl}


\bibitem[\protect\citeauthoryear{Tan and Le}{Tan and Le}{2019b}]%
        {tan2020efficientnet}
\bibfield{author}{\bibinfo{person}{Mingxing Tan} {and} \bibinfo{person}{Quoc~V. Le}.} \bibinfo{year}{2019}\natexlab{b}.
\newblock \showarticletitle{EfficientNet: Rethinking Model Scaling for Convolutional Neural Networks}. In \bibinfo{booktitle}{\emph{ICML}}.
\newblock


\bibitem[\protect\citeauthoryear{Tang, Yan, Wang, Yang, Wu, Wang, Yue, and Li}{Tang et~al\mbox{.}}{2019}]%
        {tang2019rapid}
\bibfield{author}{\bibinfo{person}{Wenzhi Tang}, \bibinfo{person}{Tingting Yan}, \bibinfo{person}{Fei Wang}, \bibinfo{person}{Jingxian Yang}, \bibinfo{person}{Jian Wu}, \bibinfo{person}{Jianlong Wang}, \bibinfo{person}{Tianli Yue}, {and} \bibinfo{person}{Zhonghong Li}.} \bibinfo{year}{2019}\natexlab{}.
\newblock \showarticletitle{Rapid fabrication of wearable carbon nanotube/graphite strain sensor for real-time monitoring of plant growth}.
\newblock \bibinfo{journal}{\emph{Carbon}}  \bibinfo{volume}{147} (\bibinfo{year}{2019}), \bibinfo{pages}{295--302}.
\newblock


\bibitem[\protect\citeauthoryear{Teimouri, Dyrmann, Nielsen, Mathiassen, Somerville, and Jørgensen}{Teimouri et~al\mbox{.}}{2018}]%
        {s18051580}
\bibfield{author}{\bibinfo{person}{Nima Teimouri}, \bibinfo{person}{Mads Dyrmann}, \bibinfo{person}{Per~Rydahl Nielsen}, \bibinfo{person}{Solvejg~Kopp Mathiassen}, \bibinfo{person}{Gayle~J. Somerville}, {and} \bibinfo{person}{Rasmus~Nyholm Jørgensen}.} \bibinfo{year}{2018}\natexlab{}.
\newblock \showarticletitle{Weed Growth Stage Estimator Using Deep Convolutional Neural Networks}.
\newblock \bibinfo{journal}{\emph{Sensors}} \bibinfo{volume}{18}, \bibinfo{number}{5} (\bibinfo{year}{2018}).
\newblock
\showISSN{1424-8220}
\urldef\tempurl%
\url{http://www.mdpi.com/1424-8220/18/5/1580}
\showURL{%
\tempurl}


\bibitem[\protect\citeauthoryear{Teixid{\'o}, Font, Pallej{\`a}, Tresanchez, Nogu{\'e}s, and Palac{\'\i}n}{Teixid{\'o} et~al\mbox{.}}{2012}]%
        {teixido2012definition}
\bibfield{author}{\bibinfo{person}{Merc{\`e} Teixid{\'o}}, \bibinfo{person}{Davinia Font}, \bibinfo{person}{Tom{\`a}s Pallej{\`a}}, \bibinfo{person}{Marcel Tresanchez}, \bibinfo{person}{Miquel Nogu{\'e}s}, {and} \bibinfo{person}{Jordi Palac{\'\i}n}.} \bibinfo{year}{2012}\natexlab{}.
\newblock \showarticletitle{Definition of linear color models in the RGB vector color space to detect red peaches in orchard images taken under natural illumination}.
\newblock \bibinfo{journal}{\emph{Sensors}} \bibinfo{volume}{12}, \bibinfo{number}{6} (\bibinfo{year}{2012}), \bibinfo{pages}{7701--7718}.
\newblock


\bibitem[\protect\citeauthoryear{Thulasidasan, Chennupati, Bilmes, Bhattacharya, and Michalak}{Thulasidasan et~al\mbox{.}}{2019}]%
        {Thulasidasan:NEURIPS2019}
\bibfield{author}{\bibinfo{person}{Sunil Thulasidasan}, \bibinfo{person}{Gopinath Chennupati}, \bibinfo{person}{Jeff~A Bilmes}, \bibinfo{person}{Tanmoy Bhattacharya}, {and} \bibinfo{person}{Sarah Michalak}.} \bibinfo{year}{2019}\natexlab{}.
\newblock \showarticletitle{On Mixup Training: Improved Calibration and Predictive Uncertainty for Deep Neural Networks}. In \bibinfo{booktitle}{\emph{Advances in Neural Information Processing Systems}}, \bibfield{editor}{\bibinfo{person}{H.~Wallach}, \bibinfo{person}{H.~Larochelle}, \bibinfo{person}{A.~Beygelzimer}, \bibinfo{person}{F.~d\textquotesingle Alch\'{e}-Buc}, \bibinfo{person}{E.~Fox}, {and} \bibinfo{person}{R.~Garnett}} (Eds.), Vol.~\bibinfo{volume}{32}. \bibinfo{publisher}{Curran Associates, Inc.}
\newblock
\urldef\tempurl%
\url{https://proceedings.neurips.cc/paper/2019/file/36ad8b5f42db492827016448975cc22d-Paper.pdf}
\showURL{%
\tempurl}


\bibitem[\protect\citeauthoryear{Tian, Wang, Liu, Qiao, and Li}{Tian et~al\mbox{.}}{2020}]%
        {tian2020computer}
\bibfield{author}{\bibinfo{person}{Hongkun Tian}, \bibinfo{person}{Tianhai Wang}, \bibinfo{person}{Yadong Liu}, \bibinfo{person}{Xi Qiao}, {and} \bibinfo{person}{Yanzhou Li}.} \bibinfo{year}{2020}\natexlab{}.
\newblock \showarticletitle{Computer vision technology in agricultural automation—A review}.
\newblock \bibinfo{journal}{\emph{Information Processing in Agriculture}} \bibinfo{volume}{7}, \bibinfo{number}{1} (\bibinfo{year}{2020}), \bibinfo{pages}{1--19}.
\newblock


\bibitem[\protect\citeauthoryear{Tian, Guo, Chen, Wang, Long, and Ma}{Tian et~al\mbox{.}}{2019a}]%
        {TIAN2019104840}
\bibfield{author}{\bibinfo{person}{Mengxiao Tian}, \bibinfo{person}{Hao Guo}, \bibinfo{person}{Hong Chen}, \bibinfo{person}{Qing Wang}, \bibinfo{person}{Chengjiang Long}, {and} \bibinfo{person}{Yuhao Ma}.} \bibinfo{year}{2019}\natexlab{a}.
\newblock \showarticletitle{Automated pig counting using deep learning}.
\newblock \bibinfo{journal}{\emph{Computers and Electronics in Agriculture}}  \bibinfo{volume}{163} (\bibinfo{year}{2019}), \bibinfo{pages}{104840}.
\newblock
\showISSN{0168-1699}
\urldef\tempurl%
\url{https://doi.org/10.1016/j.compag.2019.05.049}
\showDOI{\tempurl}


\bibitem[\protect\citeauthoryear{Tian, Shen, Chen, and He}{Tian et~al\mbox{.}}{2019b}]%
        {FCOS2019}
\bibfield{author}{\bibinfo{person}{Zhi Tian}, \bibinfo{person}{Chunhua Shen}, \bibinfo{person}{Hao Chen}, {and} \bibinfo{person}{Tong He}.} \bibinfo{year}{2019}\natexlab{b}.
\newblock \showarticletitle{FCOS: Fully Convolutional One-Stage Object Detection}. In \bibinfo{booktitle}{\emph{Proceedings of the IEEE/CVF International Conference on Computer Vision (ICCV)}}.
\newblock


\bibitem[\protect\citeauthoryear{Torres-Tello and Ko}{Torres-Tello and Ko}{2022}]%
        {torres2022optimizing}
\bibfield{author}{\bibinfo{person}{Julio Torres-Tello} {and} \bibinfo{person}{Seok-Bum Ko}.} \bibinfo{year}{2022}\natexlab{}.
\newblock \showarticletitle{Optimizing a Multispectral-Images-Based DL model, through feature selection, pruning and quantization}. In \bibinfo{booktitle}{\emph{2022 IEEE International Symposium on Circuits and Systems (ISCAS)}}. IEEE, \bibinfo{pages}{1352--1356}.
\newblock


\bibitem[\protect\citeauthoryear{Touvron, Cord, Sablayrolles, Synnaeve, and J\'egou}{Touvron et~al\mbox{.}}{2021}]%
        {Touvron_2021:ViT-Cait}
\bibfield{author}{\bibinfo{person}{Hugo Touvron}, \bibinfo{person}{Matthieu Cord}, \bibinfo{person}{Alexandre Sablayrolles}, \bibinfo{person}{Gabriel Synnaeve}, {and} \bibinfo{person}{Herv\'e J\'egou}.} \bibinfo{year}{2021}\natexlab{}.
\newblock \showarticletitle{Going Deeper With Image Transformers}. In \bibinfo{booktitle}{\emph{Proceedings of the IEEE/CVF International Conference on Computer Vision (ICCV)}}. \bibinfo{pages}{32--42}.
\newblock


\bibitem[\protect\citeauthoryear{Tripathi and Maktedar}{Tripathi and Maktedar}{2020}]%
        {tripathi2020role}
\bibfield{author}{\bibinfo{person}{Mukesh~Kumar Tripathi} {and} \bibinfo{person}{Dhananjay~D Maktedar}.} \bibinfo{year}{2020}\natexlab{}.
\newblock \showarticletitle{A role of computer vision in fruits and vegetables among various horticulture products of agriculture fields: A survey}.
\newblock \bibinfo{journal}{\emph{Information Processing in Agriculture}} \bibinfo{volume}{7}, \bibinfo{number}{2} (\bibinfo{year}{2020}), \bibinfo{pages}{183--203}.
\newblock


\bibitem[\protect\citeauthoryear{Trojak, Skowron, Sobala, Kocurek, and Pa{\l}yga}{Trojak et~al\mbox{.}}{2022}]%
        {trojak2022effects}
\bibfield{author}{\bibinfo{person}{Magdalena Trojak}, \bibinfo{person}{Ernest Skowron}, \bibinfo{person}{Tomasz Sobala}, \bibinfo{person}{Maciej Kocurek}, {and} \bibinfo{person}{Jan Pa{\l}yga}.} \bibinfo{year}{2022}\natexlab{}.
\newblock \showarticletitle{Effects of partial replacement of red by green light in the growth spectrum on photomorphogenesis and photosynthesis in tomato plants}.
\newblock \bibinfo{journal}{\emph{Photosynthesis research}} \bibinfo{volume}{151}, \bibinfo{number}{3} (\bibinfo{year}{2022}), \bibinfo{pages}{295--312}.
\newblock


\bibitem[\protect\citeauthoryear{Ubbens, Cieslak, Prusinkiewicz, and Stavness}{Ubbens et~al\mbox{.}}{2018}]%
        {ubbens2018use}
\bibfield{author}{\bibinfo{person}{Jordan Ubbens}, \bibinfo{person}{Mikolaj Cieslak}, \bibinfo{person}{Przemyslaw Prusinkiewicz}, {and} \bibinfo{person}{Ian Stavness}.} \bibinfo{year}{2018}\natexlab{}.
\newblock \showarticletitle{The use of plant models in deep learning: an application to leaf counting in rosette plants}.
\newblock \bibinfo{journal}{\emph{Plant methods}} \bibinfo{volume}{14}, \bibinfo{number}{1} (\bibinfo{year}{2018}), \bibinfo{pages}{1--10}.
\newblock


\bibitem[\protect\citeauthoryear{van Wijkvliet}{van Wijkvliet}{[n.\,d.]}]%
        {Singapore30in30}
\bibfield{author}{\bibinfo{person}{Nathalie van Wijkvliet}.} \bibinfo{year}{[n.\,d.]}\natexlab{}.
\newblock \bibinfo{title}{No space, no problem. How Singapore is turning into an edible paradise}.
\newblock \bibinfo{howpublished}{\url{https://sustainableurbandelta.com/singapore-30-by-30-food-system/}}.
\newblock
\newblock
\shownote{Accessed: 2022-8-15}.


\bibitem[\protect\citeauthoryear{Veys, Chatziavgerinos, AlSuwaidi, Hibbert, Hansen, Bernotas, Smith, Yin, Rolfe, and Grieve}{Veys et~al\mbox{.}}{2019}]%
        {veys2019multispectral}
\bibfield{author}{\bibinfo{person}{Charles Veys}, \bibinfo{person}{Fokion Chatziavgerinos}, \bibinfo{person}{Ali AlSuwaidi}, \bibinfo{person}{James Hibbert}, \bibinfo{person}{Mark Hansen}, \bibinfo{person}{Gytis Bernotas}, \bibinfo{person}{Melvyn Smith}, \bibinfo{person}{Hujun Yin}, \bibinfo{person}{Stephen Rolfe}, {and} \bibinfo{person}{Bruce Grieve}.} \bibinfo{year}{2019}\natexlab{}.
\newblock \showarticletitle{Multispectral imaging for presymptomatic analysis of light leaf spot in oilseed rape}.
\newblock \bibinfo{journal}{\emph{Plant methods}}  \bibinfo{volume}{15} (\bibinfo{year}{2019}), \bibinfo{pages}{1--12}.
\newblock


\bibitem[\protect\citeauthoryear{Vit and Shani}{Vit and Shani}{2018}]%
        {vit2018comparing}
\bibfield{author}{\bibinfo{person}{Adar Vit} {and} \bibinfo{person}{Guy Shani}.} \bibinfo{year}{2018}\natexlab{}.
\newblock \showarticletitle{Comparing rgb-d sensors for close range outdoor agricultural phenotyping}.
\newblock \bibinfo{journal}{\emph{Sensors}} \bibinfo{volume}{18}, \bibinfo{number}{12} (\bibinfo{year}{2018}), \bibinfo{pages}{4413}.
\newblock


\bibitem[\protect\citeauthoryear{Vulpi, Marani, Petitti, Reina, and Milella}{Vulpi et~al\mbox{.}}{2022}]%
        {vulpi2022rgb}
\bibfield{author}{\bibinfo{person}{Fabio Vulpi}, \bibinfo{person}{Roberto Marani}, \bibinfo{person}{Antonio Petitti}, \bibinfo{person}{Giulio Reina}, {and} \bibinfo{person}{Annalisa Milella}.} \bibinfo{year}{2022}\natexlab{}.
\newblock \showarticletitle{An RGB-D multi-view perspective for autonomous agricultural robots}.
\newblock \bibinfo{journal}{\emph{Computers and Electronics in Agriculture}}  \bibinfo{volume}{202} (\bibinfo{year}{2022}), \bibinfo{pages}{107419}.
\newblock


\bibitem[\protect\citeauthoryear{Wang and Tan}{Wang and Tan}{2017}]%
        {wang2017robust}
\bibfield{author}{\bibinfo{person}{Dong Wang} {and} \bibinfo{person}{Xiaoyang Tan}.} \bibinfo{year}{2017}\natexlab{}.
\newblock \showarticletitle{Robust distance metric learning via {B}ayesian inference}.
\newblock \bibinfo{journal}{\emph{IEEE Transactions on Image Processing}} \bibinfo{volume}{27}, \bibinfo{number}{3} (\bibinfo{year}{2017}), \bibinfo{pages}{1542--1553}.
\newblock


\bibitem[\protect\citeauthoryear{Wang, Vinson, Holmes, Seibel, Bechar, Nof, and Tao}{Wang et~al\mbox{.}}{2019a}]%
        {wang2019early}
\bibfield{author}{\bibinfo{person}{Dongyi Wang}, \bibinfo{person}{Robert Vinson}, \bibinfo{person}{Maxwell Holmes}, \bibinfo{person}{Gary Seibel}, \bibinfo{person}{Avital Bechar}, \bibinfo{person}{Shimon Nof}, {and} \bibinfo{person}{Yang Tao}.} \bibinfo{year}{2019}\natexlab{a}.
\newblock \showarticletitle{Early detection of tomato spotted wilt virus by hyperspectral imaging and outlier removal auxiliary classifier generative adversarial nets (OR-AC-GAN)}.
\newblock \bibinfo{journal}{\emph{Scientific reports}} \bibinfo{volume}{9}, \bibinfo{number}{1} (\bibinfo{year}{2019}), \bibinfo{pages}{1--14}.
\newblock


\bibitem[\protect\citeauthoryear{Wang, Feng, and Zhang}{Wang et~al\mbox{.}}{2021}]%
        {WangDengBao2021}
\bibfield{author}{\bibinfo{person}{Deng-Bao Wang}, \bibinfo{person}{Lei Feng}, {and} \bibinfo{person}{Min-Ling Zhang}.} \bibinfo{year}{2021}\natexlab{}.
\newblock \showarticletitle{Rethinking Calibration of Deep Neural Networks: Do Not Be Afraid of Overconfidence}. In \bibinfo{booktitle}{\emph{Advances in Neural Information Processing Systems}}, \bibfield{editor}{\bibinfo{person}{M.~Ranzato}, \bibinfo{person}{A.~Beygelzimer}, \bibinfo{person}{Y.~Dauphin}, \bibinfo{person}{P.S. Liang}, {and} \bibinfo{person}{J.~Wortman Vaughan}} (Eds.), Vol.~\bibinfo{volume}{34}. \bibinfo{publisher}{Curran Associates, Inc.}, \bibinfo{pages}{11809--11820}.
\newblock
\urldef\tempurl%
\url{https://proceedings.neurips.cc/paper/2021/file/61f3a6dbc9120ea78ef75544826c814e-Paper.pdf}
\showURL{%
\tempurl}


\bibitem[\protect\citeauthoryear{Wang, Wang, Zhou, Ji, Gong, Zhou, Li, and Liu}{Wang et~al\mbox{.}}{2018b}]%
        {wang2018cosface}
\bibfield{author}{\bibinfo{person}{Hao Wang}, \bibinfo{person}{Yitong Wang}, \bibinfo{person}{Zheng Zhou}, \bibinfo{person}{Xing Ji}, \bibinfo{person}{Dihong Gong}, \bibinfo{person}{Jingchao Zhou}, \bibinfo{person}{Zhifeng Li}, {and} \bibinfo{person}{Wei Liu}.} \bibinfo{year}{2018}\natexlab{b}.
\newblock \showarticletitle{Cosface: Large margin cosine loss for deep face recognition}. In \bibinfo{booktitle}{\emph{CVPR}}. \bibinfo{pages}{5265--5274}.
\newblock


\bibitem[\protect\citeauthoryear{Wang, Yang, Men, Lin, Bai, Li, Ma, Zhou, Zhou, and Yang}{Wang et~al\mbox{.}}{2022}]%
        {wang2022unifying}
\bibfield{author}{\bibinfo{person}{Peng Wang}, \bibinfo{person}{An Yang}, \bibinfo{person}{Rui Men}, \bibinfo{person}{Junyang Lin}, \bibinfo{person}{Shuai Bai}, \bibinfo{person}{Zhikang Li}, \bibinfo{person}{Jianxin Ma}, \bibinfo{person}{Chang Zhou}, \bibinfo{person}{Jingren Zhou}, {and} \bibinfo{person}{Hongxia Yang}.} \bibinfo{year}{2022}\natexlab{}.
\newblock \showarticletitle{Unifying architectures, tasks, and modalities through a simple sequence-to-sequence learning framework}.
\newblock \bibinfo{journal}{\emph{arXiv preprint arXiv:2202.03052}} (\bibinfo{year}{2022}).
\newblock


\bibitem[\protect\citeauthoryear{Wang, Wu, Zhu, Li, Zuo, and Hu}{Wang et~al\mbox{.}}{2020b}]%
        {9156697}
\bibfield{author}{\bibinfo{person}{Qilong Wang}, \bibinfo{person}{Banggu Wu}, \bibinfo{person}{Pengfei Zhu}, \bibinfo{person}{Peihua Li}, \bibinfo{person}{Wangmeng Zuo}, {and} \bibinfo{person}{Qinghua Hu}.} \bibinfo{year}{2020}\natexlab{b}.
\newblock \showarticletitle{ECA-Net: Efficient Channel Attention for Deep Convolutional Neural Networks}. In \bibinfo{booktitle}{\emph{2020 IEEE/CVF Conference on Computer Vision and Pattern Recognition (CVPR)}}. \bibinfo{pages}{11531--11539}.
\newblock
\urldef\tempurl%
\url{https://doi.org/10.1109/CVPR42600.2020.01155}
\showDOI{\tempurl}


\bibitem[\protect\citeauthoryear{Wang, Kong, Shen, Jiang, and Li}{Wang et~al\mbox{.}}{2020a}]%
        {WangXinlong-2020-solo}
\bibfield{author}{\bibinfo{person}{Xinlong Wang}, \bibinfo{person}{Tao Kong}, \bibinfo{person}{Chunhua Shen}, \bibinfo{person}{Yuning Jiang}, {and} \bibinfo{person}{Lei Li}.} \bibinfo{year}{2020}\natexlab{a}.
\newblock \showarticletitle{SOLO: Segmenting Objects by Locations}. In \bibinfo{booktitle}{\emph{Computer Vision -- ECCV 2020}}, \bibfield{editor}{\bibinfo{person}{Andrea Vedaldi}, \bibinfo{person}{Horst Bischof}, \bibinfo{person}{Thomas Brox}, {and} \bibinfo{person}{Jan-Michael Frahm}} (Eds.). \bibinfo{publisher}{Springer International Publishing}, \bibinfo{address}{Cham}, \bibinfo{pages}{649--665}.
\newblock
\showISBNx{978-3-030-58523-5}


\bibitem[\protect\citeauthoryear{Wang, Zhang, Kong, Li, and Shen}{Wang et~al\mbox{.}}{2020d}]%
        {solov2}
\bibfield{author}{\bibinfo{person}{Xinlong Wang}, \bibinfo{person}{Rufeng Zhang}, \bibinfo{person}{Tao Kong}, \bibinfo{person}{Lei Li}, {and} \bibinfo{person}{Chunhua Shen}.} \bibinfo{year}{2020}\natexlab{d}.
\newblock \showarticletitle{SOLOv2: Dynamic and Fast Instance Segmentation}. In \bibinfo{booktitle}{\emph{Advances in Neural Information Processing Systems}}, \bibfield{editor}{\bibinfo{person}{H.~Larochelle}, \bibinfo{person}{M.~Ranzato}, \bibinfo{person}{R.~Hadsell}, \bibinfo{person}{M.F. Balcan}, {and} \bibinfo{person}{H.~Lin}} (Eds.), Vol.~\bibinfo{volume}{33}. \bibinfo{publisher}{Curran Associates, Inc.}, \bibinfo{pages}{17721--17732}.
\newblock
\urldef\tempurl%
\url{https://proceedings.neurips.cc/paper/2020/file/cd3afef9b8b89558cd56638c3631868a-Paper.pdf}
\showURL{%
\tempurl}


\bibitem[\protect\citeauthoryear{Wang, Su, Zhang, and Hu}{Wang et~al\mbox{.}}{2018a}]%
        {wang2018interpret}
\bibfield{author}{\bibinfo{person}{Yulong Wang}, \bibinfo{person}{Hang Su}, \bibinfo{person}{Bo Zhang}, {and} \bibinfo{person}{Xiaolin Hu}.} \bibinfo{year}{2018}\natexlab{a}.
\newblock \showarticletitle{Interpret neural networks by identifying critical data routing paths}. In \bibinfo{booktitle}{\emph{proceedings of the IEEE conference on computer vision and pattern recognition}}. \bibinfo{pages}{8906--8914}.
\newblock


\bibitem[\protect\citeauthoryear{Wang, Yao, Kwok, and Ni}{Wang et~al\mbox{.}}{2020c}]%
        {wang2020generalizing}
\bibfield{author}{\bibinfo{person}{Yaqing Wang}, \bibinfo{person}{Quanming Yao}, \bibinfo{person}{James~T Kwok}, {and} \bibinfo{person}{Lionel~M Ni}.} \bibinfo{year}{2020}\natexlab{c}.
\newblock \showarticletitle{Generalizing from a few examples: A survey on few-shot learning}.
\newblock \bibinfo{journal}{\emph{ACM computing surveys (csur)}} \bibinfo{volume}{53}, \bibinfo{number}{3} (\bibinfo{year}{2020}), \bibinfo{pages}{1--34}.
\newblock


\bibitem[\protect\citeauthoryear{Wang, Walsh, and Koirala}{Wang et~al\mbox{.}}{2019b}]%
        {wang2019mango}
\bibfield{author}{\bibinfo{person}{Zhenglin Wang}, \bibinfo{person}{Kerry Walsh}, {and} \bibinfo{person}{Anand Koirala}.} \bibinfo{year}{2019}\natexlab{b}.
\newblock \showarticletitle{Mango fruit load estimation using a video based MangoYOLO—Kalman filter—hungarian algorithm method}.
\newblock \bibinfo{journal}{\emph{Sensors}} \bibinfo{volume}{19}, \bibinfo{number}{12} (\bibinfo{year}{2019}), \bibinfo{pages}{2742}.
\newblock


\bibitem[\protect\citeauthoryear{Wei, Feng, Chen, and An}{Wei et~al\mbox{.}}{2020}]%
        {wei2020combating}
\bibfield{author}{\bibinfo{person}{Hongxin Wei}, \bibinfo{person}{Lei Feng}, \bibinfo{person}{Xiangyu Chen}, {and} \bibinfo{person}{Bo An}.} \bibinfo{year}{2020}\natexlab{}.
\newblock \showarticletitle{Combating noisy labels by agreement: A joint training method with co-regularization}. In \bibinfo{booktitle}{\emph{CVPR}}. \bibinfo{pages}{13726--13735}.
\newblock


\bibitem[\protect\citeauthoryear{Wei, Jia, Lan, Li, Zeng, and Wang}{Wei et~al\mbox{.}}{2014}]%
        {wei2014automatic}
\bibfield{author}{\bibinfo{person}{Xiangqin Wei}, \bibinfo{person}{Kun Jia}, \bibinfo{person}{Jinhui Lan}, \bibinfo{person}{Yuwei Li}, \bibinfo{person}{Yiliang Zeng}, {and} \bibinfo{person}{Chunmei Wang}.} \bibinfo{year}{2014}\natexlab{}.
\newblock \showarticletitle{Automatic method of fruit object extraction under complex agricultural background for vision system of fruit picking robot}.
\newblock \bibinfo{journal}{\emph{Optik}} \bibinfo{volume}{125}, \bibinfo{number}{19} (\bibinfo{year}{2014}), \bibinfo{pages}{5684--5689}.
\newblock


\bibitem[\protect\citeauthoryear{Wen, Tran, and Ba}{Wen et~al\mbox{.}}{2020}]%
        {wen2020batchensemble}
\bibfield{author}{\bibinfo{person}{Yeming Wen}, \bibinfo{person}{Dustin Tran}, {and} \bibinfo{person}{Jimmy Ba}.} \bibinfo{year}{2020}\natexlab{}.
\newblock \showarticletitle{BatchEnsemble: An Alternative Approach to Efficient Ensemble and Lifelong Learning}. In \bibinfo{booktitle}{\emph{ICLR}}.
\newblock


\bibitem[\protect\citeauthoryear{Weyler, Magistri, Seitz, Behley, and Stachniss}{Weyler et~al\mbox{.}}{2022}]%
        {weyler2022field}
\bibfield{author}{\bibinfo{person}{Jan Weyler}, \bibinfo{person}{Federico Magistri}, \bibinfo{person}{Peter Seitz}, \bibinfo{person}{Jens Behley}, {and} \bibinfo{person}{Cyrill Stachniss}.} \bibinfo{year}{2022}\natexlab{}.
\newblock \showarticletitle{In-Field Phenotyping Based on Crop Leaf and Plant Instance Segmentation}. In \bibinfo{booktitle}{\emph{Proceedings of the IEEE/CVF Winter Conference on Applications of Computer Vision}}. \bibinfo{pages}{2725--2734}.
\newblock


\bibitem[\protect\citeauthoryear{Whang, Roh, Song, and Lee}{Whang et~al\mbox{.}}{2021}]%
        {Whang2021:Data-Centric-AI}
\bibfield{author}{\bibinfo{person}{Steven~Euijong Whang}, \bibinfo{person}{Yuji Roh}, \bibinfo{person}{Hwanjun Song}, {and} \bibinfo{person}{Jae-Gil Lee}.} \bibinfo{year}{2021}\natexlab{}.
\newblock \showarticletitle{Data Collection and Quality Challenges in Deep Learning: A Data-Centric AI Perspective}.
\newblock  (\bibinfo{year}{2021}).
\newblock
\urldef\tempurl%
\url{https://doi.org/10.48550/ARXIV.2112.06409}
\showDOI{\tempurl}


\bibitem[\protect\citeauthoryear{Wolny, Yu, Pape, and Kreshuk}{Wolny et~al\mbox{.}}{2022a}]%
        {Wolny_2022_CVPR}
\bibfield{author}{\bibinfo{person}{Adrian Wolny}, \bibinfo{person}{Qin Yu}, \bibinfo{person}{Constantin Pape}, {and} \bibinfo{person}{Anna Kreshuk}.} \bibinfo{year}{2022}\natexlab{a}.
\newblock \showarticletitle{Sparse Object-Level Supervision for Instance Segmentation With Pixel Embeddings}. In \bibinfo{booktitle}{\emph{Proceedings of the IEEE/CVF Conference on Computer Vision and Pattern Recognition (CVPR)}}. \bibinfo{pages}{4402--4411}.
\newblock


\bibitem[\protect\citeauthoryear{Wolny, Yu, Pape, and Kreshuk}{Wolny et~al\mbox{.}}{2022b}]%
        {wolny2022sparse}
\bibfield{author}{\bibinfo{person}{Adrian Wolny}, \bibinfo{person}{Qin Yu}, \bibinfo{person}{Constantin Pape}, {and} \bibinfo{person}{Anna Kreshuk}.} \bibinfo{year}{2022}\natexlab{b}.
\newblock \showarticletitle{Sparse object-level supervision for instance segmentation with pixel embeddings}. In \bibinfo{booktitle}{\emph{Proceedings of the IEEE/CVF Conference on Computer Vision and Pattern Recognition}}. \bibinfo{pages}{4402--4411}.
\newblock


\bibitem[\protect\citeauthoryear{Wongpanich, Pham, Demmel, Tan, Le, You, and Kumar}{Wongpanich et~al\mbox{.}}{2021}]%
        {EfficientNet-Supercomputer-Scale-2021}
\bibfield{author}{\bibinfo{person}{Arissa Wongpanich}, \bibinfo{person}{Hieu Pham}, \bibinfo{person}{James Demmel}, \bibinfo{person}{Mingxing Tan}, \bibinfo{person}{Quoc Le}, \bibinfo{person}{Yang You}, {and} \bibinfo{person}{Sameer Kumar}.} \bibinfo{year}{2021}\natexlab{}.
\newblock \showarticletitle{Training EfficientNets at Supercomputer Scale: 83\% ImageNet Top-1 Accuracy in One Hour}. In \bibinfo{booktitle}{\emph{2021 IEEE International Parallel and Distributed Processing Symposium Workshops (IPDPSW)}}. \bibinfo{pages}{947--950}.
\newblock


\bibitem[\protect\citeauthoryear{Wu, Zhang, Zhou, Xiong, Gu, and Yang}{Wu et~al\mbox{.}}{2019c}]%
        {wu2019automatic}
\bibfield{author}{\bibinfo{person}{Jingui Wu}, \bibinfo{person}{Baohua Zhang}, \bibinfo{person}{Jun Zhou}, \bibinfo{person}{Yingjun Xiong}, \bibinfo{person}{Baoxing Gu}, {and} \bibinfo{person}{Xiaolong Yang}.} \bibinfo{year}{2019}\natexlab{c}.
\newblock \showarticletitle{Automatic recognition of ripening tomatoes by combining multi-feature fusion with a bi-layer classification strategy for harvesting robots}.
\newblock \bibinfo{journal}{\emph{Sensors}} \bibinfo{volume}{19}, \bibinfo{number}{3} (\bibinfo{year}{2019}), \bibinfo{pages}{612}.
\newblock


\bibitem[\protect\citeauthoryear{Wu, Zhan, Lai, Cheng, and Yang}{Wu et~al\mbox{.}}{2019b}]%
        {wu2019ip102}
\bibfield{author}{\bibinfo{person}{Xiaoping Wu}, \bibinfo{person}{Chi Zhan}, \bibinfo{person}{Yu-Kun Lai}, \bibinfo{person}{Ming-Ming Cheng}, {and} \bibinfo{person}{Jufeng Yang}.} \bibinfo{year}{2019}\natexlab{b}.
\newblock \showarticletitle{Ip102: A large-scale benchmark dataset for insect pest recognition}. In \bibinfo{booktitle}{\emph{Proceedings of the IEEE/CVF conference on computer vision and pattern recognition}}. \bibinfo{pages}{8787--8796}.
\newblock


\bibitem[\protect\citeauthoryear{Wu, Chen, and Merhof}{Wu et~al\mbox{.}}{2020}]%
        {wu2020improving}
\bibfield{author}{\bibinfo{person}{Yuli Wu}, \bibinfo{person}{Long Chen}, {and} \bibinfo{person}{Dorit Merhof}.} \bibinfo{year}{2020}\natexlab{}.
\newblock \showarticletitle{Improving Pixel Embedding Learning through Intermediate Distance Regression Supervision for Instance Segmentation}. In \bibinfo{booktitle}{\emph{European Conference on Computer Vision Workshop}}. Springer, \bibinfo{pages}{213--227}.
\newblock


\bibitem[\protect\citeauthoryear{Wu, Donahue, Balduzzi, Simonyan, and Lillicrap}{Wu et~al\mbox{.}}{2019a}]%
        {wu2019logan}
\bibfield{author}{\bibinfo{person}{Yan Wu}, \bibinfo{person}{Jeff Donahue}, \bibinfo{person}{David Balduzzi}, \bibinfo{person}{Karen Simonyan}, {and} \bibinfo{person}{Timothy Lillicrap}.} \bibinfo{year}{2019}\natexlab{a}.
\newblock \showarticletitle{Logan: Latent optimisation for generative adversarial networks}.
\newblock \bibinfo{journal}{\emph{arXiv preprint arXiv:1912.00953}} (\bibinfo{year}{2019}).
\newblock


\bibitem[\protect\citeauthoryear{Xavier, Lima, Gheyi, Silva, Soares, and Lacerda}{Xavier et~al\mbox{.}}{2022}]%
        {xavier2022gas}
\bibfield{author}{\bibinfo{person}{Adnelba Vit{\'o}ria~Oliveira Xavier}, \bibinfo{person}{Geovani Soares~de Lima}, \bibinfo{person}{Hans~Raj Gheyi}, \bibinfo{person}{Andr{\'e} Alisson Rodrigues~da Silva}, \bibinfo{person}{Lauriane Almeida dos~Anjos Soares}, {and} \bibinfo{person}{Cassiano Nogueira~de Lacerda}.} \bibinfo{year}{2022}\natexlab{}.
\newblock \showarticletitle{Gas exchange, growth and quality of guava seedlings under salt stress and salicylic acid}.
\newblock \bibinfo{journal}{\emph{Revista Ambiente \& {\'A}gua}}  \bibinfo{volume}{17} (\bibinfo{year}{2022}).
\newblock


\bibitem[\protect\citeauthoryear{Xian, Akata, Sharma, Nguyen, Hein, and Schiele}{Xian et~al\mbox{.}}{2016}]%
        {xian2016latent}
\bibfield{author}{\bibinfo{person}{Yongqin Xian}, \bibinfo{person}{Zeynep Akata}, \bibinfo{person}{Gaurav Sharma}, \bibinfo{person}{Quynh Nguyen}, \bibinfo{person}{Matthias Hein}, {and} \bibinfo{person}{Bernt Schiele}.} \bibinfo{year}{2016}\natexlab{}.
\newblock \showarticletitle{Latent embeddings for zero-shot classification}. In \bibinfo{booktitle}{\emph{Proceedings of the IEEE conference on computer vision and pattern recognition}}. \bibinfo{pages}{69--77}.
\newblock


\bibitem[\protect\citeauthoryear{Xie, Xu, and Chuang}{Xie et~al\mbox{.}}{2013}]%
        {xie2013horizontal}
\bibfield{author}{\bibinfo{person}{Jingjing Xie}, \bibinfo{person}{Bing Xu}, {and} \bibinfo{person}{Zhang Chuang}.} \bibinfo{year}{2013}\natexlab{}.
\newblock \showarticletitle{Horizontal and Vertical Ensemble with Deep Representation for Classification}.
\newblock \bibinfo{journal}{\emph{arXiv 1306.2759}} (\bibinfo{year}{2013}).
\newblock


\bibitem[\protect\citeauthoryear{Xie, Girshick, Doll{\'{a}}r, Tu, and He}{Xie et~al\mbox{.}}{2016}]%
        {ResNeXT2016}
\bibfield{author}{\bibinfo{person}{Saining Xie}, \bibinfo{person}{Ross~B. Girshick}, \bibinfo{person}{Piotr Doll{\'{a}}r}, \bibinfo{person}{Zhuowen Tu}, {and} \bibinfo{person}{Kaiming He}.} \bibinfo{year}{2016}\natexlab{}.
\newblock \showarticletitle{Aggregated Residual Transformations for Deep Neural Networks}.
\newblock \bibinfo{journal}{\emph{arXiv Preprint 1611.05431}} (\bibinfo{year}{2016}).
\newblock


\bibitem[\protect\citeauthoryear{Yan, Li, Li, Wang, Wu, and Zhang}{Yan et~al\mbox{.}}{2021}]%
        {yan2021contnet}
\bibfield{author}{\bibinfo{person}{Haotian Yan}, \bibinfo{person}{Zhe Li}, \bibinfo{person}{Weijian Li}, \bibinfo{person}{Changhu Wang}, \bibinfo{person}{Ming Wu}, {and} \bibinfo{person}{Chuang Zhang}.} \bibinfo{year}{2021}\natexlab{}.
\newblock \showarticletitle{ConTNet: Why not use convolution and transformer at the same time?}
\newblock \bibinfo{journal}{\emph{arXiv Preprint 2104.13497}} (\bibinfo{year}{2021}).
\newblock


\bibitem[\protect\citeauthoryear{Yang and Xu}{Yang and Xu}{2021}]%
        {yang2021applications}
\bibfield{author}{\bibinfo{person}{Biyun Yang} {and} \bibinfo{person}{Yong Xu}.} \bibinfo{year}{2021}\natexlab{}.
\newblock \showarticletitle{Applications of deep-learning approaches in horticultural research: a review}.
\newblock \bibinfo{journal}{\emph{Horticulture Research}}  \bibinfo{volume}{8} (\bibinfo{year}{2021}).
\newblock


\bibitem[\protect\citeauthoryear{Yang, Gan, Wang, Hu, Lu, Liu, and Wang}{Yang et~al\mbox{.}}{2022}]%
        {yang2022empirical}
\bibfield{author}{\bibinfo{person}{Zhengyuan Yang}, \bibinfo{person}{Zhe Gan}, \bibinfo{person}{Jianfeng Wang}, \bibinfo{person}{Xiaowei Hu}, \bibinfo{person}{Yumao Lu}, \bibinfo{person}{Zicheng Liu}, {and} \bibinfo{person}{Lijuan Wang}.} \bibinfo{year}{2022}\natexlab{}.
\newblock \showarticletitle{An empirical study of gpt-3 for few-shot knowledge-based vqa}. In \bibinfo{booktitle}{\emph{Proceedings of the AAAI Conference on Artificial Intelligence}}, Vol.~\bibinfo{volume}{36}. \bibinfo{pages}{3081--3089}.
\newblock


\bibitem[\protect\citeauthoryear{Yeh, Hsieh, Suggala, Inouye, and Ravikumar}{Yeh et~al\mbox{.}}{2019}]%
        {yeh2019fidelity}
\bibfield{author}{\bibinfo{person}{Chih-Kuan Yeh}, \bibinfo{person}{Cheng-Yu Hsieh}, \bibinfo{person}{Arun Suggala}, \bibinfo{person}{David~I Inouye}, {and} \bibinfo{person}{Pradeep~K Ravikumar}.} \bibinfo{year}{2019}\natexlab{}.
\newblock \showarticletitle{On the (in) fidelity and sensitivity of explanations}.
\newblock \bibinfo{journal}{\emph{Advances in Neural Information Processing Systems}}  \bibinfo{volume}{32} (\bibinfo{year}{2019}).
\newblock


\bibitem[\protect\citeauthoryear{Yeh, Kim, Yen, and Ravikumar}{Yeh et~al\mbox{.}}{2018}]%
        {yeh2018representer}
\bibfield{author}{\bibinfo{person}{Chih-Kuan Yeh}, \bibinfo{person}{Joon Kim}, \bibinfo{person}{Ian En-Hsu Yen}, {and} \bibinfo{person}{Pradeep~K Ravikumar}.} \bibinfo{year}{2018}\natexlab{}.
\newblock \showarticletitle{Representer point selection for explaining deep neural networks}.
\newblock \bibinfo{journal}{\emph{Advances in neural information processing systems}}  \bibinfo{volume}{31} (\bibinfo{year}{2018}).
\newblock


\bibitem[\protect\citeauthoryear{Yeshitela, Robbertse, and Stassen}{Yeshitela et~al\mbox{.}}{2005}]%
        {yeshitela2005effects}
\bibfield{author}{\bibinfo{person}{T Yeshitela}, \bibinfo{person}{PJ Robbertse}, {and} \bibinfo{person}{PJC Stassen}.} \bibinfo{year}{2005}\natexlab{}.
\newblock \showarticletitle{Effects of pruning on flowering, yield and fruit quality in mango (Mangifera indica)}.
\newblock \bibinfo{journal}{\emph{Australian Journal of Experimental Agriculture}} \bibinfo{volume}{45}, \bibinfo{number}{10} (\bibinfo{year}{2005}), \bibinfo{pages}{1325--1330}.
\newblock


\bibitem[\protect\citeauthoryear{Yeung, Rundo, Nan, Sala, Sch{\"o}nlieb, and Yang}{Yeung et~al\mbox{.}}{2021}]%
        {yeung2021calibrating}
\bibfield{author}{\bibinfo{person}{Michael Yeung}, \bibinfo{person}{Leonardo Rundo}, \bibinfo{person}{Yang Nan}, \bibinfo{person}{Evis Sala}, \bibinfo{person}{Carola-Bibiane Sch{\"o}nlieb}, {and} \bibinfo{person}{Guang Yang}.} \bibinfo{year}{2021}\natexlab{}.
\newblock \showarticletitle{Calibrating the Dice loss to handle neural network overconfidence for biomedical image segmentation}.
\newblock \bibinfo{journal}{\emph{arXiv preprint arXiv:2111.00528}} (\bibinfo{year}{2021}).
\newblock


\bibitem[\protect\citeauthoryear{Ying, Huang, Liu, Shao, and Zhou}{Ying et~al\mbox{.}}{2021}]%
        {ying2021embedmask}
\bibfield{author}{\bibinfo{person}{Hui Ying}, \bibinfo{person}{Zhaojin Huang}, \bibinfo{person}{Shu Liu}, \bibinfo{person}{Tianjia Shao}, {and} \bibinfo{person}{Kun Zhou}.} \bibinfo{year}{2021}\natexlab{}.
\newblock \showarticletitle{EmbedMask: Embedding Coupling for Instance Segmentation.}. In \bibinfo{booktitle}{\emph{IJCAI}}. \bibinfo{pages}{1266--1273}.
\newblock


\bibitem[\protect\citeauthoryear{Yu, Zhang, Chang, and Jaakkola}{Yu et~al\mbox{.}}{2021}]%
        {yu2021understanding}
\bibfield{author}{\bibinfo{person}{Mo Yu}, \bibinfo{person}{Yang Zhang}, \bibinfo{person}{Shiyu Chang}, {and} \bibinfo{person}{Tommi Jaakkola}.} \bibinfo{year}{2021}\natexlab{}.
\newblock \showarticletitle{Understanding interlocking dynamics of cooperative rationalization}.
\newblock \bibinfo{journal}{\emph{Advances in Neural Information Processing Systems}}  \bibinfo{volume}{34} (\bibinfo{year}{2021}), \bibinfo{pages}{12822--12835}.
\newblock


\bibitem[\protect\citeauthoryear{Yu, Zhang, Liu, Yang, and Zhang}{Yu et~al\mbox{.}}{2020}]%
        {9119372}
\bibfield{author}{\bibinfo{person}{Yang Yu}, \bibinfo{person}{Kailiang Zhang}, \bibinfo{person}{Hui Liu}, \bibinfo{person}{Li Yang}, {and} \bibinfo{person}{Dongxing Zhang}.} \bibinfo{year}{2020}\natexlab{}.
\newblock \showarticletitle{Real-Time Visual Localization of the Picking Points for a Ridge-Planting Strawberry Harvesting Robot}.
\newblock \bibinfo{journal}{\emph{IEEE Access}}  \bibinfo{volume}{8} (\bibinfo{year}{2020}), \bibinfo{pages}{116556--116568}.
\newblock
\urldef\tempurl%
\url{https://doi.org/10.1109/ACCESS.2020.3003034}
\showDOI{\tempurl}


\bibitem[\protect\citeauthoryear{Yu, Zhang, Yang, and Zhang}{Yu et~al\mbox{.}}{2019}]%
        {yu2019fruit}
\bibfield{author}{\bibinfo{person}{Yang Yu}, \bibinfo{person}{Kailiang Zhang}, \bibinfo{person}{Li Yang}, {and} \bibinfo{person}{Dongxing Zhang}.} \bibinfo{year}{2019}\natexlab{}.
\newblock \showarticletitle{Fruit detection for strawberry harvesting robot in non-structural environment based on Mask-RCNN}.
\newblock \bibinfo{journal}{\emph{Computers and Electronics in Agriculture}}  \bibinfo{volume}{163} (\bibinfo{year}{2019}), \bibinfo{pages}{104846}.
\newblock


\bibitem[\protect\citeauthoryear{Yu, Wong, and Wen}{Yu et~al\mbox{.}}{2011}]%
        {yu2011modified}
\bibfield{author}{\bibinfo{person}{Zhiwen Yu}, \bibinfo{person}{Hau-San Wong}, {and} \bibinfo{person}{Guihua Wen}.} \bibinfo{year}{2011}\natexlab{}.
\newblock \showarticletitle{A modified support vector machine and its application to image segmentation}.
\newblock \bibinfo{journal}{\emph{Image and Vision Computing}} \bibinfo{volume}{29}, \bibinfo{number}{1} (\bibinfo{year}{2011}), \bibinfo{pages}{29--40}.
\newblock


\bibitem[\protect\citeauthoryear{Yuan, Zhu, Wang, Cheng, and Cai}{Yuan et~al\mbox{.}}{2022}]%
        {yuan2022improved}
\bibfield{author}{\bibinfo{person}{Hongbo Yuan}, \bibinfo{person}{Jiajun Zhu}, \bibinfo{person}{Qifan Wang}, \bibinfo{person}{Man Cheng}, {and} \bibinfo{person}{Zhenjiang Cai}.} \bibinfo{year}{2022}\natexlab{}.
\newblock \showarticletitle{An Improved DeepLab v3+ Deep Learning Network Applied to the Segmentation of Grape Leaf Black Rot Spots}.
\newblock \bibinfo{journal}{\emph{Frontiers in Plant Science}}  \bibinfo{volume}{13} (\bibinfo{year}{2022}).
\newblock


\bibitem[\protect\citeauthoryear{Yuan, Guo, Liu, Zhou, Yu, and Wu}{Yuan et~al\mbox{.}}{2021}]%
        {Yuan_2021_ICCV}
\bibfield{author}{\bibinfo{person}{Kun Yuan}, \bibinfo{person}{Shaopeng Guo}, \bibinfo{person}{Ziwei Liu}, \bibinfo{person}{Aojun Zhou}, \bibinfo{person}{Fengwei Yu}, {and} \bibinfo{person}{Wei Wu}.} \bibinfo{year}{2021}\natexlab{}.
\newblock \showarticletitle{Incorporating Convolution Designs Into Visual Transformers}. In \bibinfo{booktitle}{\emph{Proceedings of the IEEE/CVF International Conference on Computer Vision (ICCV)}}. \bibinfo{pages}{579--588}.
\newblock


\bibitem[\protect\citeauthoryear{Yuan, Lv, Zhang, Fu, Gao, Zhang, Li, Zhang, and Zhang}{Yuan et~al\mbox{.}}{2020}]%
        {yuan2020robust}
\bibfield{author}{\bibinfo{person}{Ting Yuan}, \bibinfo{person}{Lin Lv}, \bibinfo{person}{Fan Zhang}, \bibinfo{person}{Jun Fu}, \bibinfo{person}{Jin Gao}, \bibinfo{person}{Junxiong Zhang}, \bibinfo{person}{Wei Li}, \bibinfo{person}{Chunlong Zhang}, {and} \bibinfo{person}{Wenqiang Zhang}.} \bibinfo{year}{2020}\natexlab{}.
\newblock \showarticletitle{Robust cherry tomatoes detection algorithm in greenhouse scene based on SSD}.
\newblock \bibinfo{journal}{\emph{Agriculture}} \bibinfo{volume}{10}, \bibinfo{number}{5} (\bibinfo{year}{2020}), \bibinfo{pages}{160}.
\newblock


\bibitem[\protect\citeauthoryear{Zambon, Cecchini, Egidi, Saporito, and Colantoni}{Zambon et~al\mbox{.}}{2019}]%
        {zambon2019revolution}
\bibfield{author}{\bibinfo{person}{Ilaria Zambon}, \bibinfo{person}{Massimo Cecchini}, \bibinfo{person}{Gianluca Egidi}, \bibinfo{person}{Maria~Grazia Saporito}, {and} \bibinfo{person}{Andrea Colantoni}.} \bibinfo{year}{2019}\natexlab{}.
\newblock \showarticletitle{Revolution 4.0: Industry vs. agriculture in a future development for SMEs}.
\newblock \bibinfo{journal}{\emph{Processes}} \bibinfo{volume}{7}, \bibinfo{number}{1} (\bibinfo{year}{2019}), \bibinfo{pages}{36}.
\newblock


\bibitem[\protect\citeauthoryear{Zhang, Xie, Zhou, Wang, and Zhang}{Zhang et~al\mbox{.}}{2020b}]%
        {zhang2020state}
\bibfield{author}{\bibinfo{person}{Baohua Zhang}, \bibinfo{person}{Yuanxin Xie}, \bibinfo{person}{Jun Zhou}, \bibinfo{person}{Kai Wang}, {and} \bibinfo{person}{Zhen Zhang}.} \bibinfo{year}{2020}\natexlab{b}.
\newblock \showarticletitle{State-of-the-art robotic grippers, grasping and control strategies, as well as their applications in agricultural robots: A review}.
\newblock \bibinfo{journal}{\emph{Computers and Electronics in Agriculture}}  \bibinfo{volume}{177} (\bibinfo{year}{2020}), \bibinfo{pages}{105694}.
\newblock


\bibitem[\protect\citeauthoryear{Zhang, Bengio, Hardt, Recht, and Vinyals}{Zhang et~al\mbox{.}}{2017a}]%
        {zhang2017understanding}
\bibfield{author}{\bibinfo{person}{Chiyuan Zhang}, \bibinfo{person}{Samy Bengio}, \bibinfo{person}{Moritz Hardt}, \bibinfo{person}{Benjamin Recht}, {and} \bibinfo{person}{Oriol Vinyals}.} \bibinfo{year}{2017}\natexlab{a}.
\newblock \showarticletitle{Understanding deep learning requires rethinking generalization}.
\newblock \bibinfo{journal}{\emph{arXiv 1611.03530}} (\bibinfo{year}{2017}).
\newblock


\bibitem[\protect\citeauthoryear{Zhang, Zhang, Li, and Qiao}{Zhang et~al\mbox{.}}{2016}]%
        {zhang2016joint}
\bibfield{author}{\bibinfo{person}{Kaipeng Zhang}, \bibinfo{person}{Zhanpeng Zhang}, \bibinfo{person}{Zhifeng Li}, {and} \bibinfo{person}{Yu Qiao}.} \bibinfo{year}{2016}\natexlab{}.
\newblock \showarticletitle{Joint face detection and alignment using multitask cascaded convolutional networks}.
\newblock \bibinfo{journal}{\emph{IEEE signal processing letters}} \bibinfo{volume}{23}, \bibinfo{number}{10} (\bibinfo{year}{2016}), \bibinfo{pages}{1499--1503}.
\newblock


\bibitem[\protect\citeauthoryear{Zhang, Gui, Khattak, Wang, Gao, and Jia}{Zhang et~al\mbox{.}}{2019a}]%
        {zhang2019multi}
\bibfield{author}{\bibinfo{person}{Li Zhang}, \bibinfo{person}{Guan Gui}, \bibinfo{person}{Abdul~Mateen Khattak}, \bibinfo{person}{Minjuan Wang}, \bibinfo{person}{Wanlin Gao}, {and} \bibinfo{person}{Jingdun Jia}.} \bibinfo{year}{2019}\natexlab{a}.
\newblock \showarticletitle{Multi-task cascaded convolutional networks based intelligent fruit detection for designing automated robot}.
\newblock \bibinfo{journal}{\emph{IEEE Access}}  \bibinfo{volume}{7} (\bibinfo{year}{2019}), \bibinfo{pages}{56028--56038}.
\newblock


\bibitem[\protect\citeauthoryear{Zhang, Jia, Gui, Hao, Gao, and Wang}{Zhang et~al\mbox{.}}{2018a}]%
        {8520836}
\bibfield{author}{\bibinfo{person}{Li Zhang}, \bibinfo{person}{Jingdun Jia}, \bibinfo{person}{Guan Gui}, \bibinfo{person}{Xia Hao}, \bibinfo{person}{Wanlin Gao}, {and} \bibinfo{person}{Minjuan Wang}.} \bibinfo{year}{2018}\natexlab{a}.
\newblock \showarticletitle{Deep Learning Based Improved Classification System for Designing Tomato Harvesting Robot}.
\newblock \bibinfo{journal}{\emph{IEEE Access}}  \bibinfo{volume}{6} (\bibinfo{year}{2018}), \bibinfo{pages}{67940--67950}.
\newblock
\showISSN{2169-3536}
\urldef\tempurl%
\url{https://doi.org/10.1109/ACCESS.2018.2879324}
\showDOI{\tempurl}


\bibitem[\protect\citeauthoryear{Zhang, Jia, Gui, Hao, Gao, and Wang}{Zhang et~al\mbox{.}}{2018b}]%
        {zhang2018deep}
\bibfield{author}{\bibinfo{person}{Li Zhang}, \bibinfo{person}{Jingdun Jia}, \bibinfo{person}{Guan Gui}, \bibinfo{person}{Xia Hao}, \bibinfo{person}{Wanlin Gao}, {and} \bibinfo{person}{Minjuan Wang}.} \bibinfo{year}{2018}\natexlab{b}.
\newblock \showarticletitle{Deep learning based improved classification system for designing tomato harvesting robot}.
\newblock \bibinfo{journal}{\emph{IEEE Access}}  \bibinfo{volume}{6} (\bibinfo{year}{2018}), \bibinfo{pages}{67940--67950}.
\newblock


\bibitem[\protect\citeauthoryear{Zhang, Xu, Xu, Ma, Chen, and Fu}{Zhang et~al\mbox{.}}{2020c}]%
        {zhang2020growth}
\bibfield{author}{\bibinfo{person}{Lingxian Zhang}, \bibinfo{person}{Zanyu Xu}, \bibinfo{person}{Dan Xu}, \bibinfo{person}{Juncheng Ma}, \bibinfo{person}{Yingyi Chen}, {and} \bibinfo{person}{Zetian Fu}.} \bibinfo{year}{2020}\natexlab{c}.
\newblock \showarticletitle{Growth monitoring of greenhouse lettuce based on a convolutional neural network}.
\newblock \bibinfo{journal}{\emph{Horticulture research}}  \bibinfo{volume}{7} (\bibinfo{year}{2020}).
\newblock


\bibitem[\protect\citeauthoryear{Zhang, Chi, Yao, Lei, and Li}{Zhang et~al\mbox{.}}{2020a}]%
        {Zhang_2020_CVPR}
\bibfield{author}{\bibinfo{person}{Shifeng Zhang}, \bibinfo{person}{Cheng Chi}, \bibinfo{person}{Yongqiang Yao}, \bibinfo{person}{Zhen Lei}, {and} \bibinfo{person}{Stan~Z. Li}.} \bibinfo{year}{2020}\natexlab{a}.
\newblock \showarticletitle{Bridging the Gap Between Anchor-Based and Anchor-Free Detection via Adaptive Training Sample Selection}. In \bibinfo{booktitle}{\emph{Proceedings of the IEEE/CVF Conference on Computer Vision and Pattern Recognition (CVPR)}}.
\newblock


\bibitem[\protect\citeauthoryear{Zhang, Zhang, Zhang, Wang, and Shi}{Zhang et~al\mbox{.}}{2019b}]%
        {zhang2019cucumber}
\bibfield{author}{\bibinfo{person}{Shanwen Zhang}, \bibinfo{person}{Subing Zhang}, \bibinfo{person}{Chuanlei Zhang}, \bibinfo{person}{Xianfeng Wang}, {and} \bibinfo{person}{Yun Shi}.} \bibinfo{year}{2019}\natexlab{b}.
\newblock \showarticletitle{Cucumber leaf disease identification with global pooling dilated convolutional neural network}.
\newblock \bibinfo{journal}{\emph{Computers and Electronics in Agriculture}}  \bibinfo{volume}{162} (\bibinfo{year}{2019}), \bibinfo{pages}{422--430}.
\newblock


\bibitem[\protect\citeauthoryear{Zhang}{Zhang}{2021}]%
        {zhang2021case}
\bibfield{author}{\bibinfo{person}{Wendong Zhang}.} \bibinfo{year}{2021}\natexlab{}.
\newblock \showarticletitle{The Case for Healthy US-China Agricultural Trade Relations despite Deglobalization Pressures}.
\newblock \bibinfo{journal}{\emph{Applied Economic Perspectives and Policy}} \bibinfo{volume}{43}, \bibinfo{number}{1} (\bibinfo{year}{2021}), \bibinfo{pages}{225--247}.
\newblock


\bibitem[\protect\citeauthoryear{Zhang, Pang, Chen, and Loy}{Zhang et~al\mbox{.}}{2021a}]%
        {zhang2021knet}
\bibfield{author}{\bibinfo{person}{Wenwei Zhang}, \bibinfo{person}{Jiangmiao Pang}, \bibinfo{person}{Kai Chen}, {and} \bibinfo{person}{Chen~Change Loy}.} \bibinfo{year}{2021}\natexlab{a}.
\newblock \showarticletitle{{K-Net: Towards} Unified Image Segmentation}. In \bibinfo{booktitle}{\emph{NeurIPS}}.
\newblock


\bibitem[\protect\citeauthoryear{Zhang and Cai}{Zhang and Cai}{2011}]%
        {zhang2011climate}
\bibfield{author}{\bibinfo{person}{Xiao Zhang} {and} \bibinfo{person}{Ximing Cai}.} \bibinfo{year}{2011}\natexlab{}.
\newblock \showarticletitle{Climate change impacts on global agricultural land availability}.
\newblock \bibinfo{journal}{\emph{Environmental Research Letters}} \bibinfo{volume}{6}, \bibinfo{number}{1} (\bibinfo{year}{2011}), \bibinfo{pages}{014014}.
\newblock


\bibitem[\protect\citeauthoryear{Zhang, Zhou, Lin, and Sun}{Zhang et~al\mbox{.}}{2017b}]%
        {zhang2017shufflenet}
\bibfield{author}{\bibinfo{person}{Xiangyu Zhang}, \bibinfo{person}{Xinyu Zhou}, \bibinfo{person}{Mengxiao Lin}, {and} \bibinfo{person}{Jian Sun}.} \bibinfo{year}{2017}\natexlab{b}.
\newblock \showarticletitle{ShuffleNet: An Extremely Efficient Convolutional Neural Network for Mobile Devices}.
\newblock \bibinfo{journal}{\emph{arXiv Preprint 1707.01083}} (\bibinfo{year}{2017}).
\newblock


\bibitem[\protect\citeauthoryear{Zhang, Ti{\v{n}}o, Leonardis, and Tang}{Zhang et~al\mbox{.}}{2021b}]%
        {zhang2021survey}
\bibfield{author}{\bibinfo{person}{Yu Zhang}, \bibinfo{person}{Peter Ti{\v{n}}o}, \bibinfo{person}{Ale{\v{s}} Leonardis}, {and} \bibinfo{person}{Ke Tang}.} \bibinfo{year}{2021}\natexlab{b}.
\newblock \showarticletitle{A survey on neural network interpretability}.
\newblock \bibinfo{journal}{\emph{IEEE Transactions on Emerging Topics in Computational Intelligence}} (\bibinfo{year}{2021}).
\newblock


\bibitem[\protect\citeauthoryear{Zhao, Sheng, Wang, Tang, Chen, Cai, and Ling}{Zhao et~al\mbox{.}}{2018}]%
        {zhao2019m2det}
\bibfield{author}{\bibinfo{person}{Qijie Zhao}, \bibinfo{person}{Tao Sheng}, \bibinfo{person}{Yongtao Wang}, \bibinfo{person}{Zhi Tang}, \bibinfo{person}{Ying Chen}, \bibinfo{person}{Ling Cai}, {and} \bibinfo{person}{Haibin Ling}.} \bibinfo{year}{2018}\natexlab{}.
\newblock \showarticletitle{M2Det: A Single-Shot Object Detector based on Multi-Level Feature Pyramid Network}.
\newblock \bibinfo{journal}{\emph{arXiv Preprint 1811.04533}} (\bibinfo{year}{2018}).
\newblock


\bibitem[\protect\citeauthoryear{Zheng, Awadallah, and Dumais}{Zheng et~al\mbox{.}}{2021}]%
        {zheng2021mlc}
\bibfield{author}{\bibinfo{person}{Guoqing Zheng}, \bibinfo{person}{Ahmed~Hassan Awadallah}, {and} \bibinfo{person}{Susan Dumais}.} \bibinfo{year}{2021}\natexlab{}.
\newblock \showarticletitle{Meta Label Correction for Noisy Label Learning}. In \bibinfo{booktitle}{\emph{AAAI}}, Vol.~\bibinfo{volume}{35}.
\newblock


\bibitem[\protect\citeauthoryear{Zhong, Deng, Wang, Hu, Peng, Tao, and Huang}{Zhong et~al\mbox{.}}{2019}]%
        {zhong2019unequal}
\bibfield{author}{\bibinfo{person}{Yaoyao Zhong}, \bibinfo{person}{Weihong Deng}, \bibinfo{person}{Mei Wang}, \bibinfo{person}{Jiani Hu}, \bibinfo{person}{Jianteng Peng}, \bibinfo{person}{Xunqiang Tao}, {and} \bibinfo{person}{Yaohai Huang}.} \bibinfo{year}{2019}\natexlab{}.
\newblock \showarticletitle{Unequal-training for deep face recognition with long-tailed noisy data}. In \bibinfo{booktitle}{\emph{CVPR}}. \bibinfo{pages}{7812--7821}.
\newblock


\bibitem[\protect\citeauthoryear{Zhou, Cui, Wei, and Chen}{Zhou et~al\mbox{.}}{2020}]%
        {boyan2020bbn}
\bibfield{author}{\bibinfo{person}{Boyan Zhou}, \bibinfo{person}{Quan Cui}, \bibinfo{person}{Xiu-Shen Wei}, {and} \bibinfo{person}{Zhao-Min Chen}.} \bibinfo{year}{2020}\natexlab{}.
\newblock \showarticletitle{{BBN}: Bilateral-Branch Network with Cumulative Learning for Long-Tailed Visual Recognition}.
\newblock  (\bibinfo{year}{2020}), \bibinfo{pages}{1--8}.
\newblock


\bibitem[\protect\citeauthoryear{Zhou, Kang, Jin, Yang, Lian, Hou, and Feng}{Zhou et~al\mbox{.}}{2021}]%
        {zhou2021deepvit}
\bibfield{author}{\bibinfo{person}{Daquan Zhou}, \bibinfo{person}{Bingyi Kang}, \bibinfo{person}{Xiaojie Jin}, \bibinfo{person}{Linjie Yang}, \bibinfo{person}{Xiaochen Lian}, \bibinfo{person}{Qibin Hou}, {and} \bibinfo{person}{Jiashi Feng}.} \bibinfo{year}{2021}\natexlab{}.
\newblock \showarticletitle{DeepViT: Towards Deeper Vision Transformer}.
\newblock \bibinfo{journal}{\emph{arXiv preprint arXiv:2103.11886}} (\bibinfo{year}{2021}).
\newblock


\bibitem[\protect\citeauthoryear{Zhou, Zhuo, and Krähenbühl}{Zhou et~al\mbox{.}}{2019}]%
        {Bottom_Up_Zhou_2019}
\bibfield{author}{\bibinfo{person}{Xingyi Zhou}, \bibinfo{person}{Jiacheng Zhuo}, {and} \bibinfo{person}{Philipp Krähenbühl}.} \bibinfo{year}{2019}\natexlab{}.
\newblock \showarticletitle{Bottom-Up Object Detection by Grouping Extreme and Center Points}. In \bibinfo{booktitle}{\emph{2019 IEEE/CVF Conference on Computer Vision and Pattern Recognition (CVPR)}}.
\newblock


\bibitem[\protect\citeauthoryear{Zhou}{Zhou}{2018}]%
        {zhou2018brief}
\bibfield{author}{\bibinfo{person}{Zhi-Hua Zhou}.} \bibinfo{year}{2018}\natexlab{}.
\newblock \showarticletitle{A brief introduction to weakly supervised learning}.
\newblock \bibinfo{journal}{\emph{National science review}} \bibinfo{volume}{5}, \bibinfo{number}{1} (\bibinfo{year}{2018}), \bibinfo{pages}{44--53}.
\newblock


\bibitem[\protect\citeauthoryear{Zhu, Zhu, Li, Wu, Li, Wang, and Dai}{Zhu et~al\mbox{.}}{2022}]%
        {zhu2022uni}
\bibfield{author}{\bibinfo{person}{Xizhou Zhu}, \bibinfo{person}{Jinguo Zhu}, \bibinfo{person}{Hao Li}, \bibinfo{person}{Xiaoshi Wu}, \bibinfo{person}{Hongsheng Li}, \bibinfo{person}{Xiaohua Wang}, {and} \bibinfo{person}{Jifeng Dai}.} \bibinfo{year}{2022}\natexlab{}.
\newblock \showarticletitle{Uni-perceiver: Pre-training unified architecture for generic perception for zero-shot and few-shot tasks}. In \bibinfo{booktitle}{\emph{Proceedings of the IEEE/CVF Conference on Computer Vision and Pattern Recognition}}. \bibinfo{pages}{16804--16815}.
\newblock


\end{thebibliography}
\end{spacing}
\appendix

\end{document}